\UseRawInputEncoding
\RequirePackage{fix-cm}
\documentclass[twocolumn]{svjour3}          
\smartqed  
\usepackage{graphicx}
\usepackage[colorlinks,bookmarksopen,bookmarksnumbered,citecolor=green, linkcolor=red, urlcolor=blue]{hyperref}
\usepackage{makecell} 
\usepackage{siunitx}
\usepackage{afterpage}
\usepackage[figuresright]{rotating}
\usepackage{adjustbox}
\usepackage[numbers]{natbib}
\usepackage[doipre={DOI:}]{uri}
\usepackage{subfigure}
\usepackage{multirow}
\usepackage{booktabs}
\usepackage{xcolor}
\usepackage{amsmath,amssymb,amsfonts}
\usepackage{verbatim}
\usepackage{url}
\usepackage{tabularx}
\usepackage{lscape}
\usepackage{colortbl}
\usepackage{float}
\newcolumntype{C}[1]{>{\centering\let\newline\\\arraybackslash\hspace{0pt}}m{#1}} 
\newcolumntype{L}[1]{>{\let\newline\\\arraybackslash\hspace{0pt}}m{#1}} 

\usepackage{booktabs,lscape}

\journalname{IJCV}

\begin{document}

\title{Multi-Modal 3D Object Detection in Autonomous Driving: A Survey
}

\author{Yingjie Wang$^{*}$         \and
        Qiuyu Mao$^{*}$ \and
        Hanqi Zhu \and
        Jiajun Deng \and
        Yu Zhang \and
        Jianmin Ji \and
        Houqiang Li \and
        Yanyong Zhang
}

\institute{
            *equal contribution\\
            \\
            Yingjie Wang \at
            University of Science and Technology of China\\
              \email{yingjiewang@mail.ustc.edu.cn}
              \and
            Qiuyu Mao \at
            University of Science and Technology of China\\
              \email{qymao@mail.ustc.edu.cn}
              \and
            Hanqi Zhu \at
            University of Science and Technology of China\\
              \email{zhuhanqi@mail.ustc.edu.cn}
              \and
            Jiajun Deng \at
            University of Science and Technology of China\\
              \email{dengjj@ustc.edu.cn}
              \and
            Yu Zhang \at
            University of Science and Technology of China\\
              \email{yuzhang@ustc.edu.cn}
              \and
            Jianmin Ji \at
            University of Science and Technology of China\\
             \email{jianmin@ustc.edu.cn}
             \and
            Houqiang Li \at
            University of Science and Technology of China\\
             \email{lihq@ustc.edu.cn}
             \and
            Yanyong Zhang, corresponding author \at
            University of Science and Technology of China\\
              \email{yanyongz@ustc.edu.cn}
}

\date{Received: date / Accepted: date}

\maketitle

\begin{abstract}
The past decade has witnessed the rapid development of autonomous driving systems. However, it remains a daunting task to achieve full autonomy, especially when it comes to understanding the ever-changing, complex driving scenes.
To alleviate the difficulty of perception, self-driving vehicles are usually equipped with a suite of sensors (\emph{e.g.}, cameras, LiDARs), hoping to capture the scenes with overlapping perspectives to minimize blind spots. Fusing these data streams and exploiting their complementary properties is thus rapidly becoming the current trend. 

Nonetheless, combining data that are captured by different sensors with drastically different ranging/ima-ging mechanisms is not a trivial task; instead, many factors need to be considered and optimized. If not careful, data from one sensor may act as noises to data from another sensor, with even poorer results by fusing them. Thus far, there has been no in-depth guidelines to designing the multi-modal fusion based 3D perception algorithms. 
To fill in the void and motivate further investigation, this survey conducts a thorough study of tens of recent deep learning based multi-modal 3D detection networks (with a special emphasis on LiDAR-camera fusion), focusing on their fusion stage (\emph{i.e.}, when to fuse), fusion inputs (\emph{i.e.}, what to fuse), and fusion granularity (\emph{i.e.}, how to fuse). These important design choices play a critical role in determining the performance of the fusion algorithm.   

In this survey, we first introduce the background of popular sensors used for self-driving, their data properties, and the corresponding object detection algorithms.  
Next, we discuss existing datasets that can be used for evaluating multi-modal 3D object detection algorithms.
Then we present a review of multi-modal fusion based 3D detection networks, taking a close look at their fusion stage, fusion input and fusion granularity, and how these design choices evolve with time and technology. 
After the review, we discuss open challenges as well as possible solutions.
We hope that this survey can help researchers to get familiar with the field and embark on investigations in the area of multi-modal 3D object detection.
\keywords{3D Object Detection \and Multi-modal Fusion \and Sensor Fusion \and Autonomous Driving}
\end{abstract}

\section{Introduction}
\label{sec:intro}
Recent breakthroughs in deep learning and computer vision~\citep{DBLP:journals/tits/ChenLLCWGLW21,DBLP:journals/csur/IoannidouCNK17,DBLP:journals/nn/Schmidhuber15} have fostered the rapid development of autonomous driving, which promises to free the drivers, to decrease  traffic congestion, and to improve road safety. The potential of autonomous driving is, however, not yet fully unleashed, largely due to the unsatisfactory perception performance in real-world driving scenarios.
As a result, even if autonomous vehicles (AVs) have seen applications in many confined and controlled environments, deploying them in urban environments still poses dire technological challenges~\citep{Guo2019Is, Urmson2008Autonomous}.

\begin{figure}[t]
  \centering
  \includegraphics[width=1\linewidth]{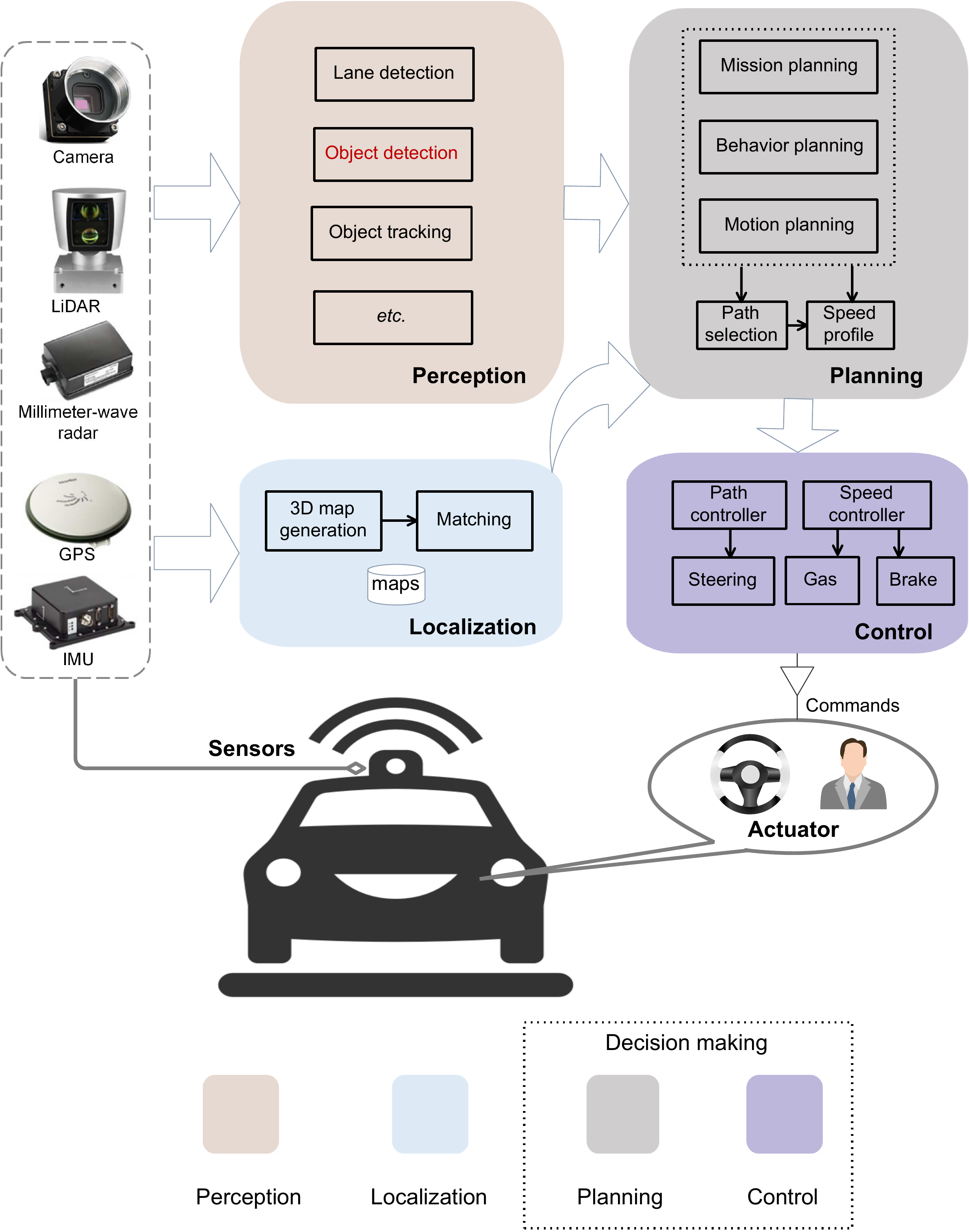}
  \caption{The typical architecture for an autonomous driving system, consisting of three subsystems: perception, localization and decision making } 
  \label{fig:perception}
\end{figure}

Fig.~\ref{fig:perception} illustrates a typical AV system that consists of three subsystems: perception, localization and decision making. 
The AV system capitalizes multiple sensors (\emph{e.g.}, LiDAR, camera) to collect raw sensor data. Taking the raw sensor data as input, the perception and localization subsystems execute several important tasks to identify and localize objects of interest, namely, object detection, tracking, 3D map generation and mapping, \emph{etc.} Given the objects and their locations, the decision making subsystem can navigate and make self-driving decisions. Among all the tasks, object detection, aiming to localize and categorize objects of interest, is of great significance. 

With the breakthrough of deep learning techniques, 2D object detection has drawn a great deal of attention, resulting in a plethora of algorithms~\citep{girshick2015fast, rcnn, Lin2016Feature, Lin2017Focal,ssd,YOLO,Ren2017Faster}. However, localizing the objects in the 2D image plane is far from the demand of AVs to perceive the 3D real world. To this end, the task of 3D object detection is proposed with the requirement of predicting the object's three-dimensional location, shape, and rotational angles.
Compared to the well-studied 2D object detection, 3D object detection is not only more important to autonomous vehicles but also more challenging. The challenges mainly stem from the fact that 3D driving scenes are also much more complex for perception~\citep{zhang2021autonomous}.
For example, we need additional depth and rotation parameters to locate an object in 3D space.

\begin{figure}[t]
\centering
\subfigure[person (10m): image pixels]{
\begin{minipage}[t]{0.5\linewidth}
\centering
\includegraphics[width=1.6in]{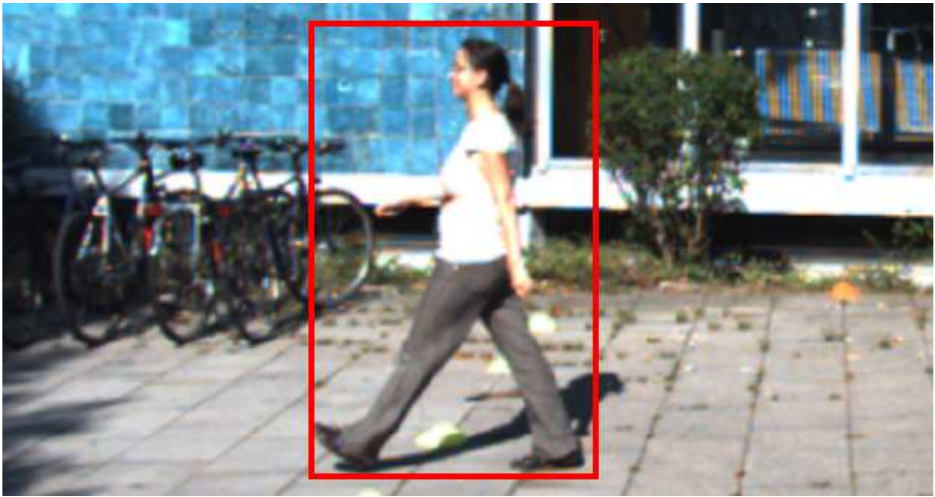}
\end{minipage}%
}%
\subfigure[person (10m): LiDAR points]{
\begin{minipage}[t]{0.5\linewidth}
\centering
\includegraphics[width=1.6in]{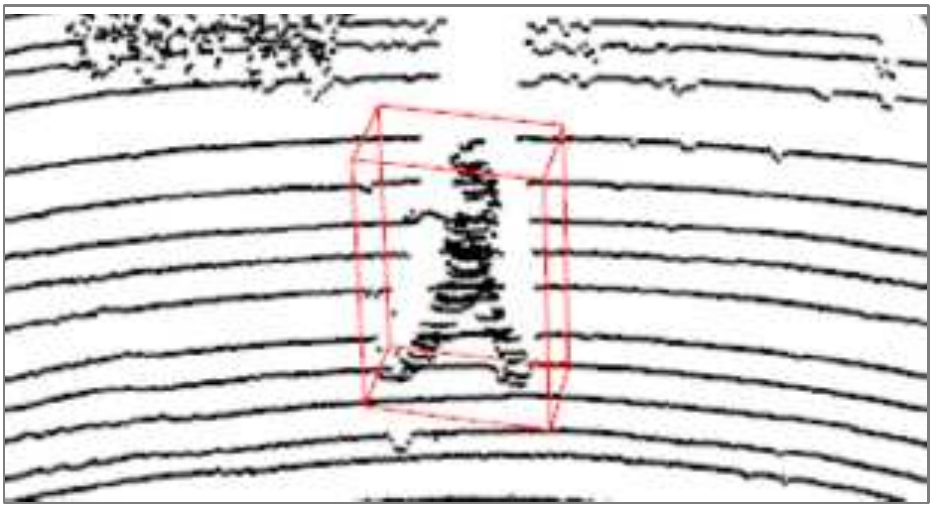}
\end{minipage}%
}%
\\
\subfigure[person (60m): image pixels]{
\begin{minipage}[t]{0.5\linewidth}
\centering
\includegraphics[width=1.6in]{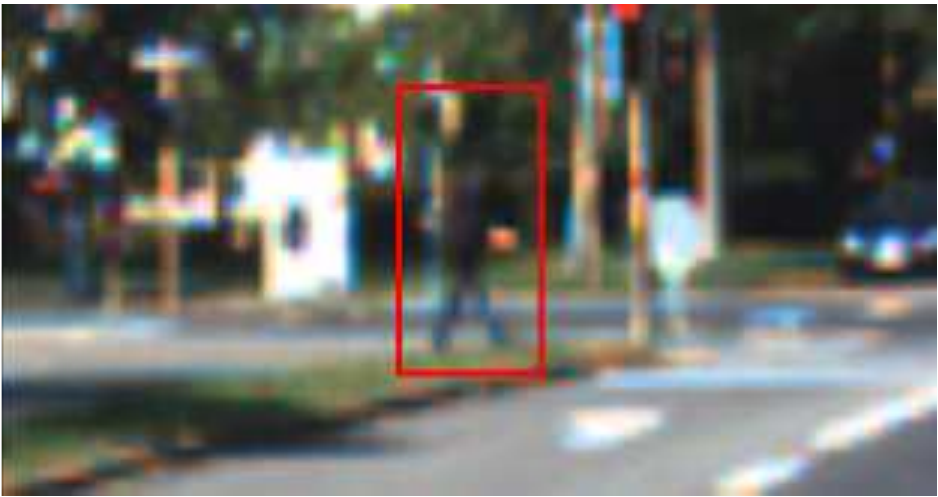}
\end{minipage}%
}%
\subfigure[person (60m): LiDAR points]{
\begin{minipage}[t]{0.5\linewidth}
\centering
\includegraphics[width=1.6in]{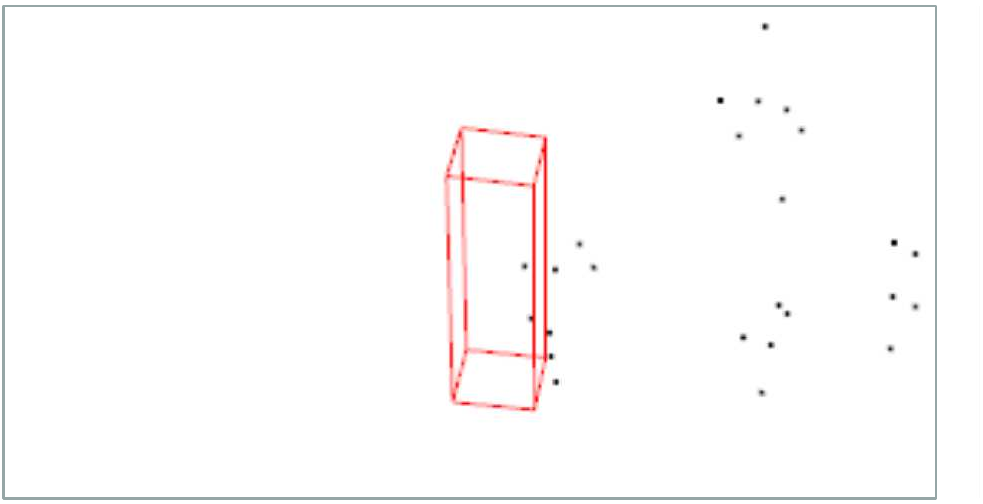}
\end{minipage}%
}%
\centering
\caption{ The image for a person 10 meters away is shown in (a), and the corresponding point cloud data is shown in (b). The image for another person 60 meters away is shown in (c), and the corresponding point cloud data is shown in (d). It is clear that point clouds get very sparse at long distances.  We modify the picture from Fig. 1 in~\citep{zhang2021farawayfrustum}. 
}
\label{fig:problem for lidar}
\end{figure}

In the real world, performing 3D object detection through a single type of sensor data is far from being sufficient. Firstly, each type of sensor data has its inherent limitation and shortcomings. For example, a camera-based system suffers from the lack of accurate depth information,
while a LiDAR-only system is hampered by lower input data resolution, especially at long distances. 
As shown in Fig.~\ref{fig:problem for lidar} and Tab.~\ref{Point Cloud and Image}, in average, for objects which are far from the ego-sensor (\textgreater{} 60m in KITTI), there are usually less than 10 LiDAR points but are still with more than 400 image pixels.
Secondly, the perception system must be robust against sensor malfunctioning, failure, or simply under-performing, hence mandating the necessity of having more than one type of sensor. Thirdly, data from different sensors complement each other naturally. Their combination could lead to a more comprehensive depicting of the environment and thus better detection results. 

\begin{figure}[t]
  \centering
  \includegraphics[width=0.65\linewidth]{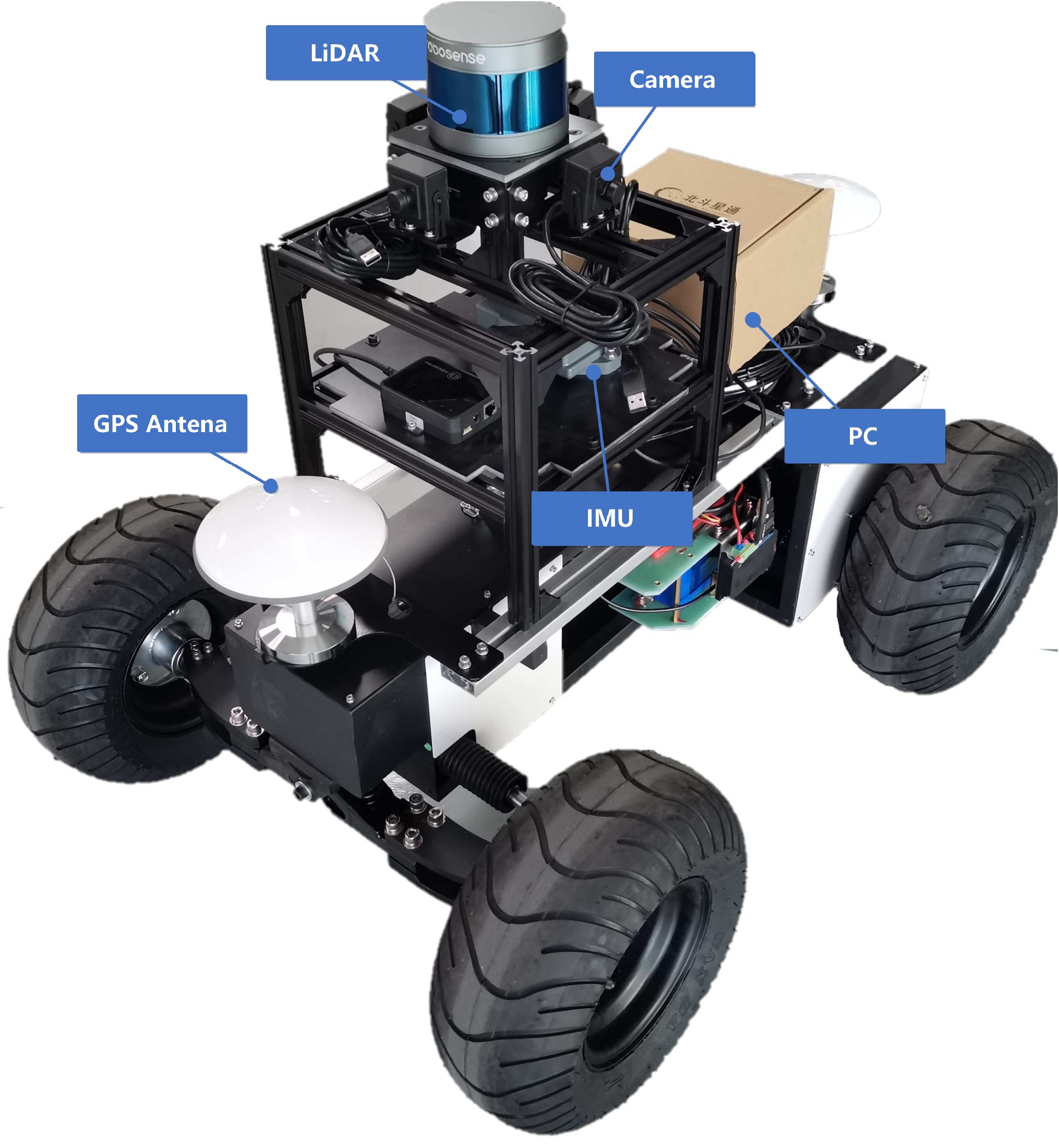}
  \caption{\label{fig:sonic}The autonomous car \emph{Sonic} is equipped with one LiDAR sensor (Velodyne VLP-16), four cameras, and one GPS. Note that the image is modified from~\citep{yanyong}.}
\end{figure}
Therefore, a recent trend in 3D object detection is to \emph{combine data streams from different sensors and develop multi-modal detection methods}. 
Fig.~\ref{fig:sonic} shows multiple sensors in the AV system. AVs are typically equipped with cameras, LiDARs (\emph{i.e.}, \underline{Li}ght \underline{D}etection \underline{A}nd \underline{R}anging sensors), Radars (\underline{Ra}dio \underline{d}etection \underline{a}nd \underline{r}anging sensors), GPS (Global Positioning System) and IMUs (Inertial Measurement Units)~\citep{Urmson2009Autonomous, yanyong}.
In the multi-modal methods, data from multiple types of sensors that have complementary characteristics are fused to capture the scenes with overlapping perspectives, aiming at minimizing blind spots. 

Though recent studies have demonstrated the benefits of fusion in various settings, conducting efficient and effective multi-modal detection in the real world still largely remains a myth and faces many challenges. Below we list some of these open challenges: 
\begin{itemize} 
    \item \textbf{Multi-Sensor Calibration:} Sensors of different types are not synchronized either temporally or spatially. In the temporal domain, it is hard to collect data at the same time due to independent acquisition cycles for each sensor. In the spatial domain, sensors have different angles of view when they are deployed. Thus, multi-sensor calibration is the first step before data fusion, which has not received much attention so far. 
    \item \textbf{Information Loss During Fusion:} Due to the large gap between different types of sensor data (illustrated in Tab.~\ref{Point Cloud and Image}), it is difficult to precisely align these data streams either in the input stage or in the feature space. To convert the sensor data into a representation format in which they can be aligned and fused correctly, a certain amount of information loss becomes inevitable. 
    \item \textbf{Consistent Data Augmentation Across Multiple Modalities:} Data augmentation plays a vital role in 3D object detection to enhance the size of training samples, and to ameliorate the problem of model over-fitting~\citep{augmentation}.
    Augmentation strategies such as global rotation~\cite{8578570} and random flip~\cite{pointrcnn} are widely adopted by LiDAR-based and camera-based methods but are absent in many multi-modal methods due to the concerns of leading to inconsistencies across modalities.
\end{itemize}

At present, how to address the above challenges and conduct efficient data fusion  still remains an open problem. 
If not carefully done, data fusion may cause different data streams to act as noises to each another~\citep{asvadi2018multimodal,caltagirone2019lidar}, leading to even poorer results.
In this paper, we set out \emph{to conduct a comprehensive review of recent fusion-based 3D object detection methods}. 
Such a review can help pinpoint technical challenges in sensor fusion, and help us compare and contrast various models proposed to address these challenges. 
In particular, since cameras and LiDARs are the most common sensors for autonomous driving, our review mainly focuses on the fusion of these two types of sensor data.
Specifically, when we discuss a multi-modal fusion based 3D detection algorithm, we focus on how the algorithm deals with the following three crucial design considerations:
\begin{itemize}
    \item \textbf{Fusion Stage:}
    The first design consideration is concerned with at what pipeline stage the multi-modal fusion module takes place, \emph{i.e.}, ``where to fuse''. It has three options here: early fusion~\citep{9156790,8578131}, late fusion~\citep{CLOCs}, and cascade fusion~\citep{8578200}. Early fusion usually occurs in the input stage or feature extraction stage before each branch reaches its prediction. Late fusion takes place in the prediction stage. Cascade fusion employs the hybrid mode by fusing one branch's prediction with the other's input. 
    
    The fusion stage is the most influential design consideration as it determines the overall network architecture of the fusion based detection algorithm, and early fusion is the predominant choice.

    \item \textbf{Fusion Input:} The second design consideration is concerned with how the multi-modal data are input into the fusion module, \emph{i.e.}, ``what to fuse''. The fusion module can be designed to take the raw data as input, or some type of intermediate features. For example, the fusion module can take in the LiDAR data as raw point clouds~\citep{8578131, 2020EPNet}, voxel grids~\citep{3D-CVF, chen2022autoalign,chen2022autoalignv2}, and projection on the bird's eye view (BEV) or the range view (RV)~\citep{2017Multi, 2020FuseSeg}. Meanwhile, the fusion module can take in the camera data as the feature maps, segmentation masks, and even pseudo-LiDAR point clouds~\citep{Pseudo-LiDAR}. 
    
    The fusion input is a crucial design consideration because data representation plays a significant role in the overall detection performance. Among the three considerations, it has the most options. We will carefully review these options in Sec.~\ref{sec:multi-moda}, and discuss how they evolve with time and technology. 
    
    \item \textbf{Fusion Granularity:} The third design consideration is concerned with at what granularity the two data streams are combined, \emph{i.e.}, ``how to fuse''. It usually has the three options: region of interest (RoI)-level, voxel-level, and point-level (with the last one at the finest granularity). 
    
    The fusion granularity plays an important role in determining the complexity and effectiveness of fusion; usually, finer fusion granularity requires more computing and leads to superior performance.
\end{itemize}

\begin{table*}[]
\caption{Using the sensors employed in the KITTI dataset as example, we compare the LiDAR and camera sensors~\citep{geiger2013vision}. In addition to important sensor parameters, we also qualitatively compare how external factors may affect the data quality of different sensors, with $\triangle$ indicating minor influences, $\triangle\triangle$ moderate influences and $\triangle\triangle\triangle$ significant influences~\cite{DBLP:journals/corr/abs-2112-08936}. 
}
\label{Point Cloud and Image}
\resizebox{\textwidth}{!}{
\begin{tabular}{
>{\columncolor[HTML]{FFFFFF}}l |
>{\columncolor[HTML]{FFFFFF}}l 
>{\columncolor[HTML]{FFFFFF}}l 
>{\columncolor[HTML]{FFFFFF}}l 
>{\columncolor[HTML]{FFFFFF}}l 
>{\columncolor[HTML]{FFFFFF}}l 
>{\columncolor[HTML]{FFFFFF}}l 
>{\columncolor[HTML]{FFFFFF}}l |
>{\columncolor[HTML]{EFEFEF}}l 
>{\columncolor[HTML]{EFEFEF}}l 
>{\columncolor[HTML]{EFEFEF}}l }
\hline
\cellcolor[HTML]{FFFFFF}                                      & \multicolumn{7}{c|}{\cellcolor[HTML]{FFFFFF}sensor parameters and external factors in KITTI~\citep{geiger2013vision,zhang2021farawayfrustum}}                                                                                                                                                                                                                                                                                                                                                                                                                                                                                                                                                                                                                         & \multicolumn{3}{c}{\cellcolor[HTML]{EFEFEF} external factors from~\citep{DBLP:journals/corr/abs-2112-08936}}                                                                                                                                                                                                         \\ \cline{2-11} 
\cellcolor[HTML]{FFFFFF}                                      & \cellcolor[HTML]{FFFFFF}                                 & \cellcolor[HTML]{FFFFFF}                                                           & \cellcolor[HTML]{FFFFFF}                                                                                 & \cellcolor[HTML]{FFFFFF}                                                                            & \cellcolor[HTML]{FFFFFF}                                                                               & \cellcolor[HTML]{FFFFFF}                      & \cellcolor[HTML]{FFFFFF}                                                                                                                                       & \cellcolor[HTML]{EFEFEF}                                                                          & \cellcolor[HTML]{EFEFEF}                                                                                      & \cellcolor[HTML]{EFEFEF}                               \\
\multirow{-3}{*}{\cellcolor[HTML]{FFFFFF}\textbf{}} & \multirow{-2}{*}{\cellcolor[HTML]{FFFFFF}number}         & \multirow{-2}{*}{\cellcolor[HTML]{FFFFFF}cost}                                     & \multirow{-2}{*}{\cellcolor[HTML]{FFFFFF}\begin{tabular}[c]{@{}l@{}}data \\ format\end{tabular}}         & \multirow{-2}{*}{\cellcolor[HTML]{FFFFFF}\begin{tabular}[c]{@{}l@{}}data \\ dimension\end{tabular}} & \multirow{-2}{*}{\cellcolor[HTML]{FFFFFF}\begin{tabular}[c]{@{}l@{}}density/\\ per frame\end{tabular}} & \multirow{-2}{*}{\cellcolor[HTML]{FFFFFF}FPS} & \multirow{-2}{*}{\cellcolor[HTML]{FFFFFF}\begin{tabular}[c]{@{}l@{}}object \\dis.\textgreater{}60m\end{tabular}} & \multirow{-2}{*}{\cellcolor[HTML]{EFEFEF}\begin{tabular}[c]{@{}l@{}}strong \\ light\end{tabular}} & \multirow{-2}{*}{\cellcolor[HTML]{EFEFEF}\begin{tabular}[c]{@{}l@{}}smog\\ vis.\textgreater{}2km\end{tabular}} & \multirow{-2}{*}{\cellcolor[HTML]{EFEFEF}installation} \\ \hline
\cellcolor[HTML]{FFFFFF}                                      & \cellcolor[HTML]{FFFFFF}                                 & \cellcolor[HTML]{FFFFFF}                                                           & \cellcolor[HTML]{FFFFFF}                                                                                 & \cellcolor[HTML]{FFFFFF}                                                                            & \cellcolor[HTML]{FFFFFF}                                                                               & \cellcolor[HTML]{FFFFFF}                      & \cellcolor[HTML]{FFFFFF}                                                                                                                                       & \cellcolor[HTML]{EFEFEF}                                                                          & \cellcolor[HTML]{EFEFEF}                                                                                      & \cellcolor[HTML]{EFEFEF}                               \\
\multirow{-2}{*}{\cellcolor[HTML]{FFFFFF}\textbf{LiDAR}}      & \multirow{-2}{*}{\cellcolor[HTML]{FFFFFF}1}              & \multirow{-2}{*}{\cellcolor[HTML]{FFFFFF}\$8,0000}                                 & \multirow{-2}{*}{\cellcolor[HTML]{FFFFFF}\begin{tabular}[c]{@{}l@{}}binary float \\ matrix\end{tabular}} & \multirow{-2}{*}{\cellcolor[HTML]{FFFFFF}3D}                                                        & \multirow{-2}{*}{\cellcolor[HTML]{FFFFFF}$\approx$10,000 points}                                               & \multirow{-2}{*}{\cellcolor[HTML]{FFFFFF}10}  & \multirow{-2}{*}{\cellcolor[HTML]{FFFFFF}$\approx$10 points}                                                                                                 & \multirow{-2}{*}{\cellcolor[HTML]{EFEFEF}$\triangle$}                                             & \multirow{-2}{*}{\cellcolor[HTML]{EFEFEF}$\triangle\triangle$}                                                & \multirow{-2}{*}{\cellcolor[HTML]{EFEFEF}easy}     \\
\textbf{Camera}                                               & \begin{tabular}[c]{@{}l@{}}2 grey\\ 2 color\end{tabular} & {\color[HTML]{4D5156} \begin{tabular}[c]{@{}l@{}}\$268 \\ \$511.25\end{tabular}} & png format                                                                                               & 2D                                                                                                  & 466,240 pixels                                                                                         & 15                                            & $\approx$400 pixels                                                                                                                                                 & $\triangle\triangle\triangle$                                                                              & $\triangle\triangle\triangle$                                                                                 & easier                                          \\ \hline
\end{tabular}
}
\end{table*}

Previous surveys on deep learning based multi-modal fusion methods~\citep{DBLP:journals/tits/ArnoldADFOM19, 9380166,DBLP:journals/tits/FengHRHGTWD21} cover a broad range of sensors, including radars, cameras, LiDARs, ultrasonic sensors, \emph{etc.}, and provide a relatively brief review on a list of topics including multi-object detection, tracking, environment reconstruction, \emph{etc.} 
While they are considered as a useful guide for readers to browse through the general area, our survey serves a distinctly different purpose: it targets at researchers who would like to carefully investigate the field of multi-modal 3D detection. As such, our survey intends to provide a deep and detailed review of recent research on this topic. 
Our contributions are summarized as below:
\begin{itemize}
\item[$\bullet$] We conduct an in-depth review of sensor fusion based 3D detection networks, with a special focus on LiDAR-camera fusion. We organize our discussions around the three core design considerations: fusion stage, fusion input, and fusion granularity, which answer the questions of “where to fuse”, “what to fuse”, and “how to fuse”, respectively. 

\item[$\bullet$] Most of the previous surveys have largely overlooked the fusion inputs of 3D multi-modal networks. In fact, compared to the other two design considerations, a fusion module's input exhibits the most diversity and represents the unique idea of each design. In our survey, we discuss this design consideration thoroughly. According to their fusion input choices, we categorize the fusion based 3D detection networks into a total of five categories. We review the schemes in each category in detail, and discuss how the input combination evolve with time and technology. 

\item[$\bullet$] We also summarize the popular multi-modal datasets that can be employed for 3D object detection evaluation. In addition, we carefully discuss a list of open challenges in the field as well as possible solutions, which can hopefully inspire some future research in the area of multi-modal 3D object detection.

\end{itemize}

In this paper, we first provide a brief background of typical sensors used in autonomous driving, their data properties, and 3D object detectors through single modality respectively in Sec.~\ref{sec:background}. 
In Sec.~\ref{sec:datasets}, We present a summary of popular datasets that can be employed for evaluating multi-modal 3D object detection networks. 
In Sec.~\ref{sec:multi-moda}, we present a review of multi-modal fusion methods based on three crucial design choices: fusion stage, fusion input, and fusion granularity.
Finally, we discuss open challenges and possible solutions in Sec.~\ref{sec:open_challenge}.

\section{Background}
\label{sec:background}

In this section, we provide a background overview of typical sensors employed in autonomous driving, especially on 3D object detection methods that rely on each type of sensor. 
We mainly focus our discussions on cameras and LiDARs. 
Besides, we also introduce other sensors that can be employed for 3D object detection.

\begin{figure}[t]
\centering
\subfigure[2D and 3D bounding boxes]{
\begin{minipage}[t]{0.48\linewidth}
\centering
\includegraphics[width=1.6in]{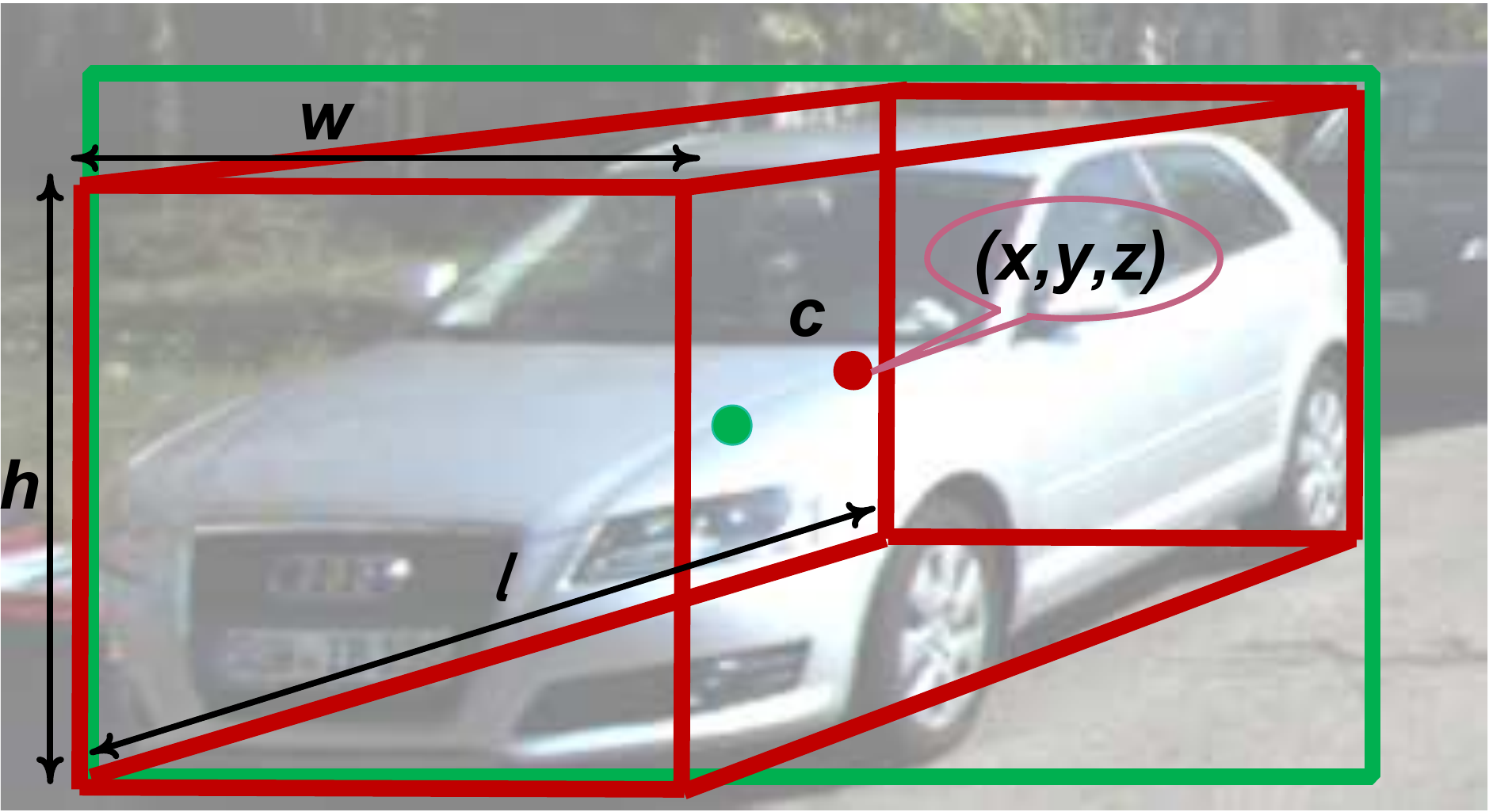}
\end{minipage}%
}
\subfigure[Bird's eye view (BEV) of the 3D box]{
\begin{minipage}[t]{0.48\linewidth}
\centering
\includegraphics[width=1.6in]{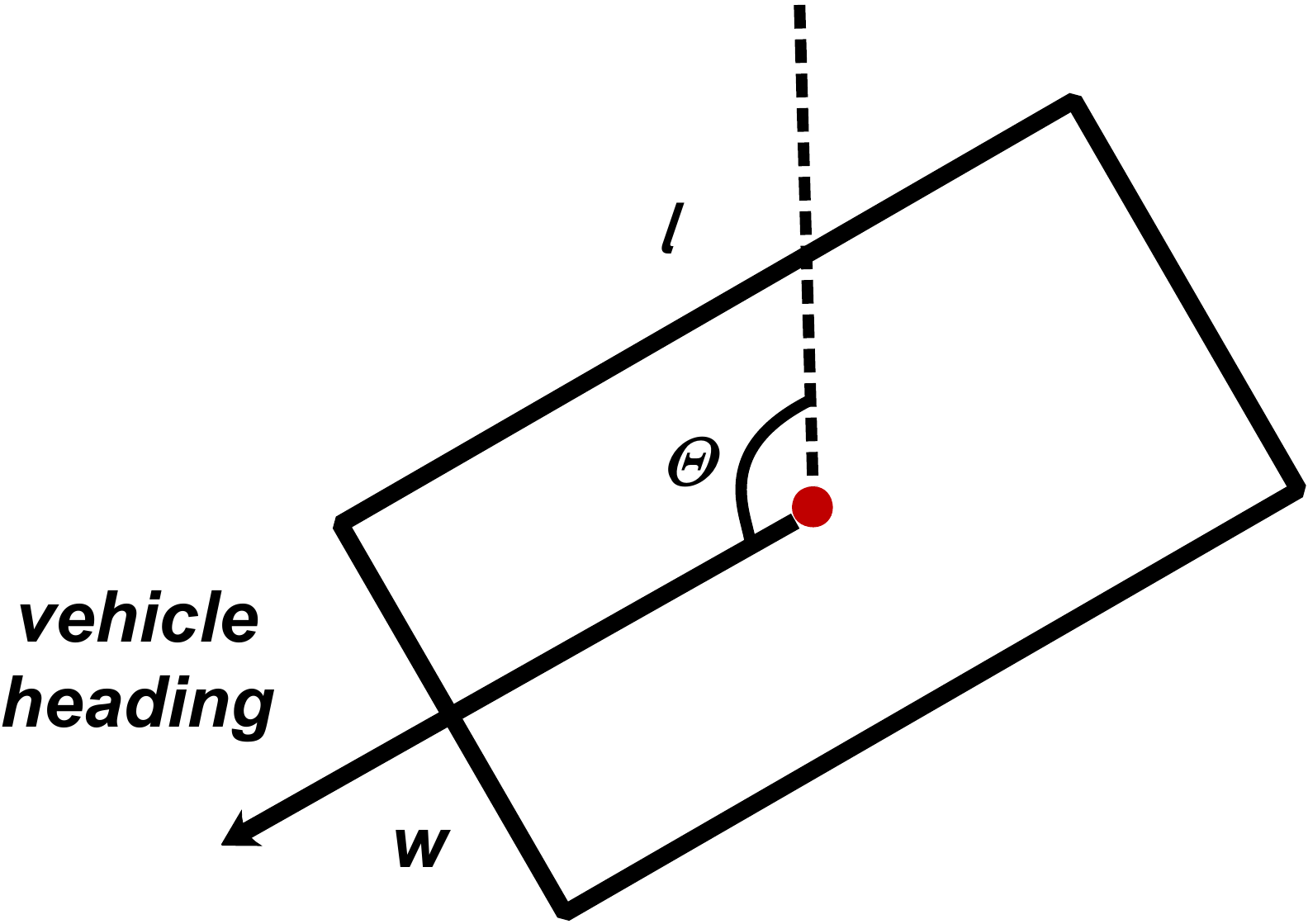}
\end{minipage}%
}%
\centering
\caption{(a) Example of 2D (green) and 3D (red) object detection bounding boxes, (b) Parameters of a 3D box in the BEV including its width \emph{w}, length \emph{l}, and vehicle heading angle \emph{$\theta$}}
\label{fig:detection_result}
\end{figure}

\subsection{3D Object Detection Task}

Before introducing 3D object detection methods through different camera settings, we first give an overview of 3D object detection. 
In the 3D object detection task, we need to provide the 3D bounding boxes of objects in the scene. As depicted in Fig.~\ref{fig:detection_result}, we are required to predict the object center's 3D coordinates \emph{c}, length \emph{l}, width \emph{w}, height \emph{h} as well as its deflection angle \emph{$\theta$} to obtain the red 3D bounding box.

\subsection{3D Object Detection through Cameras}
Cameras are the most common sensors for self-driving cars. 
A series of mature methods in 2D object detection have been developed in recent years, which can be reused in 3D object detection~\citep{girshick2015fast, Ren2017Faster}. 
Accordingly, image-based 3D object detection methods can achieve satisfactory performance at low expenses, often outperforming human experts~\citep{Volodymyr2019Human,SilverMastering}. 
Several types of cameras have been widely deployed in AV, each with pros and cons.  Below we talk about the 3D object detection algorithms via different camera settings.

\noindent\textbf{Monocular 3D Object Detection.} 
Monocular cameras provide dense information in the form of pixel intensity, which reveals shape and texture properties~\citep{Andriluka2010Monocular,Enzweiler2009Monocular}. 
The shape and texture information can also be utilized to detect lane geometry, traffic signs, and type of objects. 
The main disadvantage of using monocular cameras for 3D detection stems from the lack of depth information, which is necessary for accurate object size and position estimation for AVs~\citep{Matthews2012Monocular}. 
To compensate for this, many studies have been devoted to enhancing detection accuracy through monocular cameras~\citep{Monocular,DBLP:conf/mm/ChuDLYZJZ21,Mono3D,2020SMOKE,2020Accurate,8100080, park2021pseudo,MonoGRNet,Pseudo-LiDAR}. For example, \citet{8100080} employ a designed CNN to estimate the missing depth information, which is used later to upgrade the 2D bounding box to the 3D space. \citet{DBLP:conf/mm/ChuDLYZJZ21} perform monocular depth estimation first and lifts the 2D pixels to pseudo 3D points. They design a novel neighbor-voting method that incorporates neighbor predictions to improve object detection from severely deformed pseudo-LiDAR point clouds.
\citet{park2021pseudo} propose an end-to-end single-stage monocular-based detector. With the large-scale unlabeled data pre-training, it achieves promising detection results.

\noindent\textbf{Stereo 3D Object Detection.} 
Compared to monocular cameras, stereo cameras estimate a more accurate depth map~\citep{Engelberg2009Method,Lee2011Stereo}. Specifically, multi-view cameras can cover different ranges of scenes through different cameras and capture depth maps more accurately~\citep{Kim2005A,Kim2004A,Park2009MULTI}. 
Meanwhile, the complexity and cost involved in processing stereo images will also increase considerably. 
Some works exploit stereo images to generate dense point clouds to conduct 3D object detection tasks~\citep{3DOP,2019Stereo,Object-Centric,Triangulation,Pseudo-LiDAR++}. 
For example, \citet{3DOP} focus on generating 3D proposals by encoding object size prior, ground-plane prior, and depth information into an energy function. \citet{2019Stereo} add extra branches after the stereo Region Proposal Network (RPN) to predict sparse keypoints, viewpoints, and object dimensions, which are used to calculate coarse 3D object bounding boxes. Next, the accurate 3D bounding boxes are recovered by a region-based photometric alignment. \citet{guo2021liga} encode the depth information in the stereo cost volume, taking LiDAR features as the guidance to \textit{distill} high-level geometry-aware representations for the stereo detection network.

\noindent\textbf{Shortcomings of Camera-based 3D Object Detection.} 
To summarize, camera-based 3D object detection has several shortcomings. 
Firstly, it is difficult for monocular cameras to estimate depth, which severely limits the detection accuracy~\citep{Chen2019Surrounding,Huang2017Traffic}. Secondly, camera-based detection further suffers from adverse conditions such as poor lighting, dense smoke, or heavy fog~\cite{zhang2021autonomous,DA4AD} 
So far, camera-only 3D object detection has not been able to obtain reliable performance. As far as KITTI~\citep{geiger2013vision} dataset is concerned, the state-of-the-art stereo-based method LIGA-Stereo~\citep{guo2021liga} achieves 64.66\% mAP while monocular-based DD3D~\citep{park2021pseudo} achieves only 16.87\% mAP. To accomplish a more reliable AV system, We need to explore more powerful sensors for AVs.

\begin{table*}[]
\caption{Technical specifications for two typical Velodyne LiDARs. Note that point cloud images are from ~\citep{velodyne}. 
}
\label{64and128}
\centering
\begin{tabular}{ccllcccccc}
\hline
\multirow{2}{*}{type}                                                        & \multicolumn{3}{c}{\multirow{2}{*}{\begin{tabular}[c]{@{}c@{}}point cloud \\ image\end{tabular}}} & \multirow{2}{*}{channel} & \multirow{2}{*}{range} & \multirow{2}{*}{\begin{tabular}[c]{@{}c@{}}points scanned \\ per second\end{tabular}} & \multicolumn{1}{l}{\multirow{2}{*}{\begin{tabular}[c]{@{}l@{}}Horizontal\\ Field of View\end{tabular}}} & \multicolumn{1}{l}{\multirow{2}{*}{\begin{tabular}[c]{@{}l@{}}Vertical   \\ Field of View\end{tabular}}} & \multicolumn{1}{l}{\multirow{2}{*}{price}} \\
                                                                             & \multicolumn{3}{c}{}                                                                              &                          &                        &                                                                                       & \multicolumn{1}{l}{}                                                                                    & \multicolumn{1}{l}{}                                                                                     & \multicolumn{1}{l}{}                       \\ \hline
\multirow{4}{*}{\begin{tabular}[c]{@{}c@{}}Velodyne\\ HDL-64E\end{tabular}}  & \multicolumn{3}{c}{\multirow{4}{*}{\includegraphics[width=3.45cm,height=1.3cm]{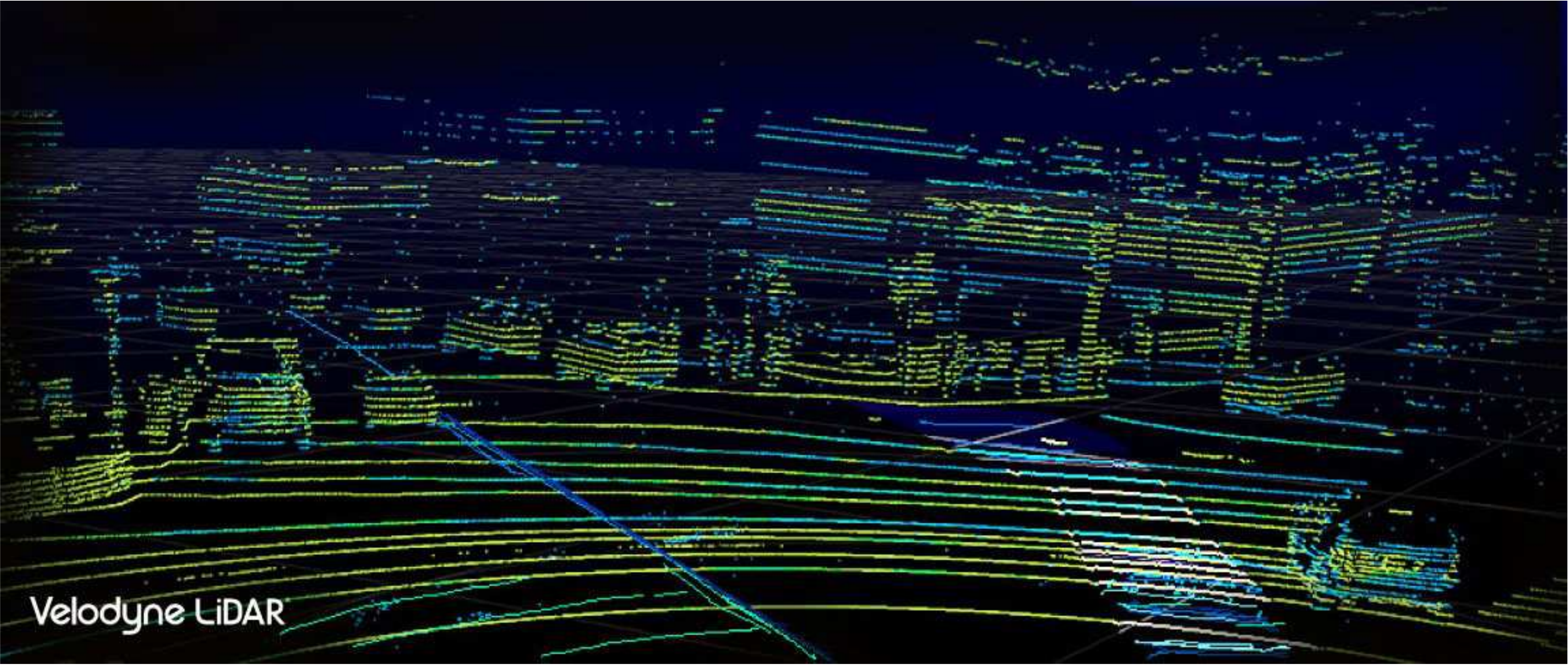}}}                                                        & \multirow{4}{*}{64}      & \multirow{4}{*}{120m}   & \multirow{4}{*}{2.2 million}                                                        & \multirow{4}{*}{\ang{360}}                                                                                    & \multirow{4}{*}{\ang{26.9}}                                                                                    & \multirow{4}{*}{\$80,000}                    \\
                                                                             & \multicolumn{3}{c}{}                                                                              &                          &                        &                                                                                       &                                                                                                         &                                                                                                          &                                            \\
                                                                             & \multicolumn{3}{c}{}                                                                              &                          &                        &                                                                                       &                                                                                                         &                                                                                                          &                                            \\
                                                                             & \multicolumn{3}{c}{}                                                                              &                          &                        &                                                                                       &                                                                                                         &                                                                                                          &                                            \\
\multirow{4}{*}{\begin{tabular}[c]{@{}c@{}}Velodyne \\ VLS-128\end{tabular}} & \multicolumn{3}{c}{\multirow{4}{*}{\includegraphics[width=3.5cm,height=1.3cm]{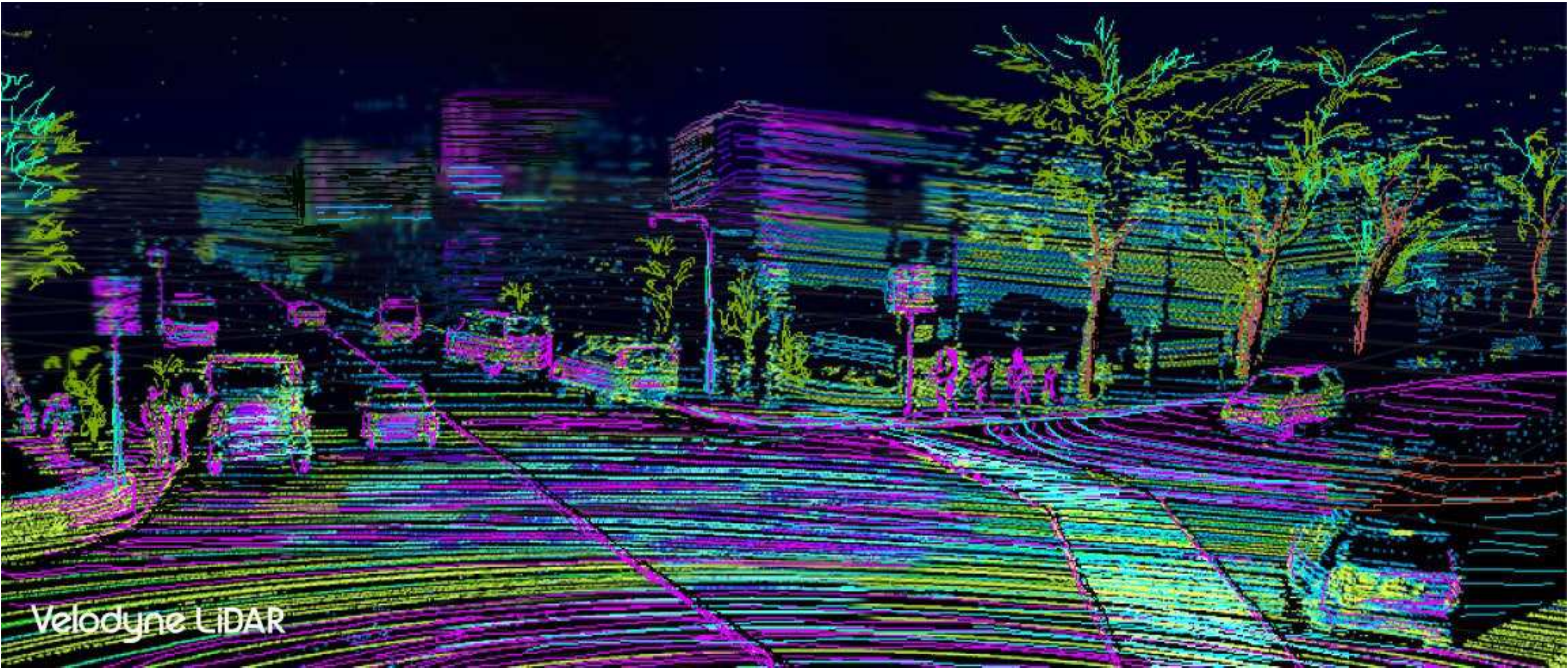}}}                                                        & \multirow{4}{*}{128}     & \multirow{4}{*}{220m}   & \multirow{4}{*}{4.8 million}                                                          & \multirow{4}{*}{\ang{360}}                                                                                    & \multirow{4}{*}{\ang{40}}                                                                                      & \multirow{4}{*}{\$100,000}                    \\
                                                                             & \multicolumn{3}{c}{}                                                                              &                          &                        &                                                                                       &                                                                                                         &                                                                                                          &                                            \\
                                                                             & \multicolumn{3}{c}{}                                                                              &                          &                        &                                                                                       &                                                                                                         &                                                                                                          &                                            \\
                                                                             & \multicolumn{3}{c}{}                                                                              &                          &                        &                                                                                       &                                                                                                         &                                                                                                          &                                            \\ \hline
\end{tabular}
\end{table*}

\subsection{3D Object Detection through LiDARs}
LiDAR sensors use lasers as the light source to complete remote sensing measurements. LiDARs detect the lightwave signal between the LiDAR sensor and the detected object~\citep{Wandinger2005Introduction}. 
It continuously emits lasers and collects the information of the reflection points to obtain a full range of environmental information. 
When the LiDAR sensor rotates one circle, all the reflected point coordinates form a \emph{point cloud}. 
As an active sensor, external illumination is not required and thus we can achieve more reliable detection under extreme lighting conditions. 
The typical resolution of LiDAR points ranges from 16 channels to 128 channels. As shown in Tab.~\ref{64and128}, we conduct a detailed comparison to help readers form a clear understanding of the two popular LiDAR sensors: Velodyne HDL-64L and VLS-128. 
From specific figures, it can be seen that all parameters of the 128-channel LiDAR are better than those of the 64-channel LiDAR. Obviously, LiDARs are quite costly compared to cameras. We can see the price of a Velodyne HDL-64 sensor is officially at \$80,000. The latest VLS-128 sensor has better performance but is also more expensive. 
Below, we briefly review existing works on 3D object detection based on the LiDAR data.

\noindent\textbf{View-Based Detection.} Many LiDAR-based methods project the LiDAR point clouds into the BEV, or RV, to leverage the off-the-shelf 2D Convolutional Neural Networks (CNNs). Early on, \citet{PIXOR} propose an efficient, proposal-free single-stage detector. It transforms the point cloud to BEV and performs 2D CNNs to extract the point cloud features. Compact and dense RV-based methods are also proposed for 3D object detection. Recently, \citet{RangeRCNN} employ a 2D backbone on the RV to learn the spatial features directly, and then adopt an R-CNN to get the 3D bounding boxes. H$^{2}$3D R-CNN~\citep{DBLP:journals/tcsv/DengZZL21} first learns RV and BEV features in a sequential pattern, then fuses the two 3D representations in a multi-view fusion block.

\noindent\textbf{Voxel-Based Detection.}
Voxel-based methods first divide points into regular 3D voxels, and then leverage the sparse convolutional neural networks~\cite{SECOND} and transformers~\cite{mao2021voxel,fan2022embracing} for feature extraction and bounding box prediction.
VoxelNet~\citep{8578570} extracts discriminative voxel features to speed up the model execution. SECOND~\citep{SECOND} reduces the computational overhead of dense 3D CNNs by applying sparse convolution. PointPillars~\citep{Lang2019PointPillars} introduces a \textit{pillar} representation (a particular form of the voxel) for the point cloud. Pillars are fast because all key operations can be formulated as 2D convolutions. Voxel R-CNN~\citep{deng2020voxel} further improves the accuracy and speed of voxel-based detectors by introducing a voxel RoI pooling operation. In addition, \citet{mao2021voxel} introduce a Transformer-based architecture that enables long-range relationships between voxels by self-attention.

\noindent\textbf{Point-Based Detection.} Recent point cloud encoders such as PointNet~\citep{8099499}, PointNet++~\citep{ISI:000452649405018}, Pointformer~\citep{pan20213d}, and other point cloud backbones~\citep{PointSIFT, OctNet} could learn representations from raw point clouds. Point-based detectors employ them to extract the spatial geometry information for downstream tasks~\citep{pointrcnn,shi2020point,9156597}. For example, \citet{pointrcnn} employ PointNet++~\citep{ISI:000452649405018} as point clouds encoder and generate 3D proposals based on the extracted semantic and geometric features. \citet{shi2020point} propose a graph neural network to detect objects from a LiDAR point cloud. 
To this end, they encode the point cloud efficiently in a fixed radius near-neighbors graph.

\noindent\textbf{Point-Voxel hybrid Detection.} In addition to the point and voxel representation introduced above, there are some works~\citep{he2020sassd,9157234,yang2019std} that adopt a hybrid pattern, utilizing both point and voxel features for 3D object detection. For example, STD~\citep{yang2019std} first generates proposals based on the point features, then employs the voxel representation in the box refinement stage. PV-RCNN~\citep{9157234} integrates the multi-scale voxel features and point cloud features for accurate 3D object detection. 
M3DETR~\citep{guan2022m3detr} encodes  point and voxel features with multi-level scale via transformers.
In general, point-voxel hybrid detectors benefit from both representations, which is superior to point or voxel-only detectors~\citep{9157234}.

Compared with camera images, LiDAR points provide strong 3D geometric information, which is essential for 3D object detection. 
Furthermore, LiDAR sensors can better adapt to external factors such as strong light, which is depicted in Tab.~\ref{Point Cloud and Image}. 
At present, LiDAR-based methods achieve better detection accuracy and higher recall than camera-based methods~\citep{2017Multi}. As far as the KITTI 3D object detection benchmark is concerned, the top monocular-images-based method DD3D~\citep{park2021pseudo} achieves 16.87\% mAP while quite a few LiDAR-based methods~\citep{deng2020voxel, DBLP:journals/tcsv/DengZZL21, 9157234,xu2021behind} achieve over 80\% mAP. However, LiDAR-only algorithms are not yet ready to be widely deployed on AVs for the following reasons: 1) LiDARs are expensive and bulky, especially compared with cameras~\citep{de2020evaluating}. 2) The working distance of the LiDAR is rather limited, point clouds far away from the LiDAR are extremely sparse~\citep{zhang2021farawayfrustum}. 3) LiDARs can not work properly under extremely severe weather such as heavy rain~\citep{DBLP:journals/tvt/WallaceHB20}.

\subsection{3D Object Detection through Other Sensors.}
In addition to cameras and LiDARs, AVs are often equipped with sensors such as millimeter wave (mmWave in short) radar sensors, infrared cameras, \emph{etc.}
In particular, mmWave radar has long been used on self-driving cars because it is more robust to severe weather conditions than cameras and LiDARs ~\citep{DBLP:journals/corr/abs-2112-08936}. More importantly, radar points provide the velocity information of the corresponding object, which is crucial for avoiding dynamic objects~\citep{patole2017automotive}. Next, we give a brief background of mmWave radars below.
\par

\noindent\textbf{MmWave Radar Sensor.}
MmWave radars are active sensors that operate in the millimeter-wave bands. They can measure the reflected waves to determine the location and velocity of objects~\citep{DBLP:conf/gcaiot/AhmadWNK20}. They are considerably cheaper than LiDARs, resistant to adverse weather conditions (fog, smoke, and dust), and insensitive to lighting variations~\citep{DBLP:journals/corr/abs-2112-08936}. However, compared with the camera and LiDAR, there are limited large-scale and public mmWave radar datasets~\citep{radar}. Moreover, due to the low resolution of the mmWave radar, it is hard to directly detect the shape of an object through sparse 2D radar points. Compared with the 3D point cloud, radar points are much noisier because of multi-path reflection, rendering it hard to perform 3D detection alone~\citep{radar2}.
\par
\begin{figure}
\centering
\subfigure[the RGB image]{
\includegraphics[width=0.9\linewidth,height=3.8cm]{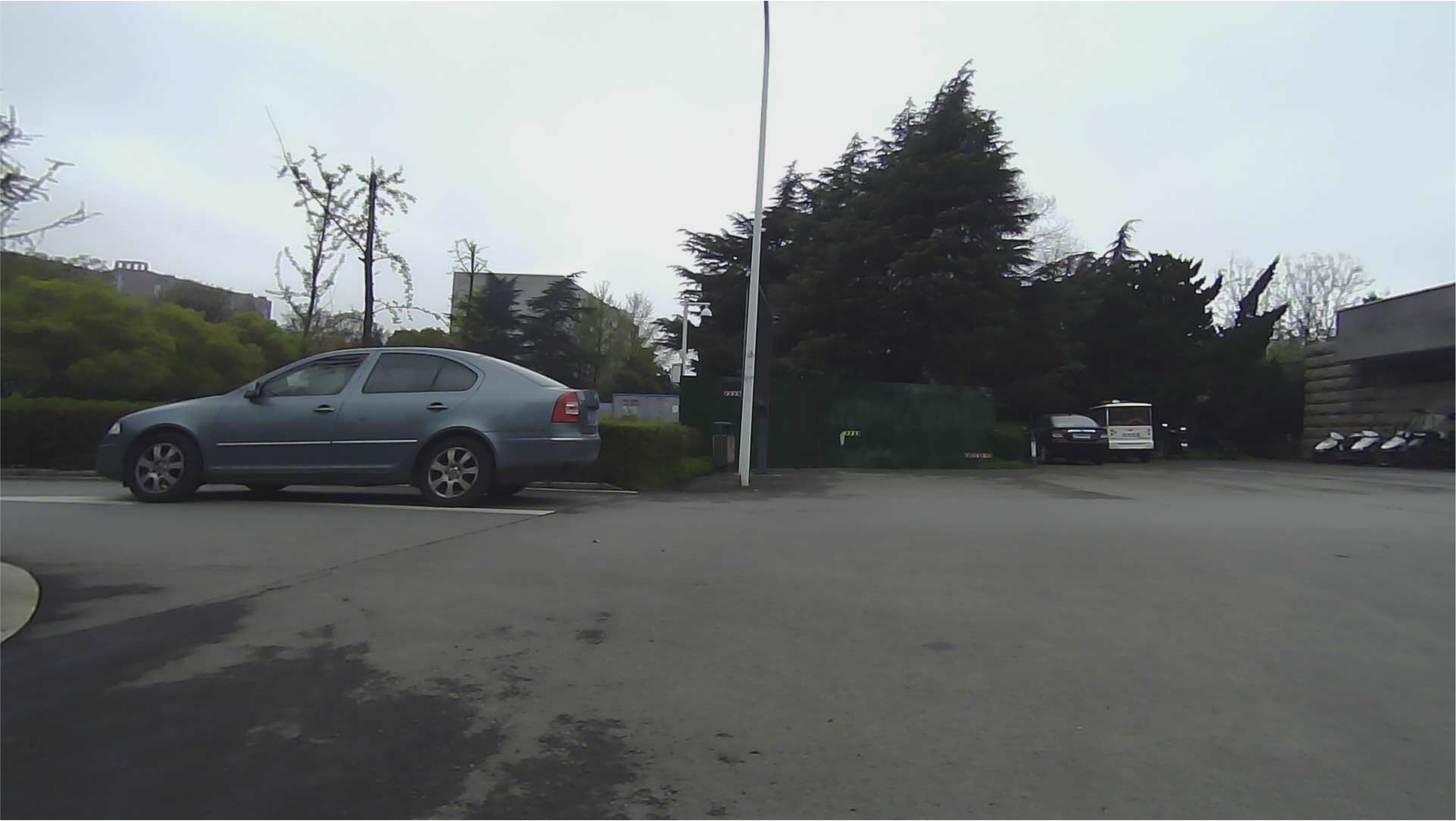}
}%
\\
\subfigure[mmWave radar: the range-azimuth heatmap]{
\includegraphics[width=0.9\linewidth,height=3.8cm]{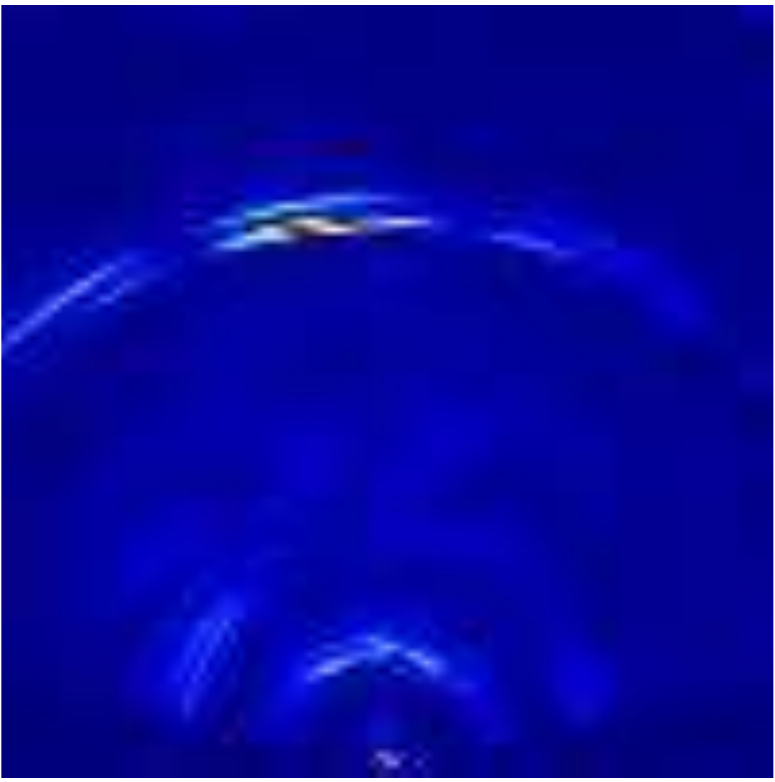}
}%
\caption{An RGB image (a) and a mmWave radar heatmap (b) on the same scene. The brightness in (b) indicates high confidence of objects. The data was collected by the authors at the north entrance of the West Campus of the University of Science and Technology of China.}
\label{radar}
\end{figure}

The mmWave radar outputs can be organized at three levels: 1) \textit{raw data} in the form of time-frequency spectrograms; 2) \textit{clusters} from applying clustering algorithms~\citep{DBLP:conf/ivs/KellnerKD12} on raw data; and 3) \textit{tracks} from performing object tracking on the clusters. 
Here, we process raw data which is collected on campus for visualization. 
As shown in Fig.~\ref{radar}, we perform two fast Fourier transforms on raw data to get the range-azimuth heatmap. The brightness in (b) represents the signal strength at that location and indicates high confidence of objects. 
Normally, datasets containing radar points generally utilize the representation of clusters, which are radar reflections with information containing position, velocity, and signal strength. The clusters are newly evaluated every cycle~\citep{nuScenes}.

\par
\noindent\textbf{MmWave Radar based 3D Object Detection.}
The mmWave radar has been widely exploited in AV systems~\citep{DBLP:journals/ijcv/MarchandC99}. 
Radars usually report the detected objects as 2D points in BEV and provide the azimuth angle and radial distance to the object. 
For each detection, the radar also reports the instantaneous velocity of the object in the radial direction. 
To the best of our knowledge, \citet{DBLP:conf/iccvw/MajorFASGHLGS19} propose the first radar-based deep neural network object detection with reliable results. 
However, radar-based 3D detectors face many challenges. Compared with LiDAR points, radar points are much noisier and less accurate, which brings difficulties in adapting LiDAR pipeline to the radar. 
Another bottleneck in radar-based detectors is the lack of publicly usable data annotated with ground-truth information~\citep{DBLP:journals/corr/abs-2010-09076}. 
In practice, mmWave radars are more often used for fusion with other sensors: \emph{e.g.}, radar-camera, radar-LiDAR~\citep{DBLP:journals/corr/abs-2011-04841,DBLP:conf/eccv/YangGLCU20}.
\subsection{Discussion}
As discussed above, different sensors have different advantages, sometimes are complementary. 
For example, cameras are high-resolution and low-cost sensors, but lack depth information and are sensitive to light conditions. 
On the contrary, LiDAR points can provide 3D spatial information of the surrounding environment, but capture only sparse points at high price.  
\par
In general, camera-based methods generate less accurate 3D bounding boxes than LiDAR-based methods.
Currently LiDAR-based methods lead in popularity in 3D object detection, while with some shortcomings. For example, 
the density of point clouds tends to decrease quickly as the distance increases while image-based detectors could still detect faraway objects.
To make good use of the complementary features and improve the overall performance, more methods try to design fusion networks to combine images with point clouds. These methods have achieved superior performance in 3D object detection tasks compared to methods relying on a single sensor. We will discuss this in Sec.~\ref{sec:multi-moda} afterwards.

\section{Datasets and Metrics}
\label{sec:datasets}
\begin{table*}[ht]
\caption{Popular multi-modal dataset comparison, including year, number of LiDARs, 
number of LiDAR channels (we report the number of channels of the top LiDAR for Waymo dataset and the maximum number of channels among 4 LiDARs for AIODrive dataset), 
number of cameras,
whether with radar,
number of 2D boxes (we don't distinguish between 2D boxes and 2D instance segmentation annotation),
number of 3D boxes, 
number of annotated classes,
and location (KA: Karlsruhe; SF: San Francisco; SG: Singapore; PT: Pittsburgh).
Note that ApolloScape's LiDARs scan with 1 beam at a high frequency to get dense point clouds.}
\label{tab:datasetcomparison}
\renewcommand\arraystretch{1.2}
\resizebox{\textwidth}{!}{
\begin{tabular}{l|lllllllll}
\hline
\textbf{dataset}    & \textbf{year}&\textbf{n-LiDAR} & \textbf{n-chn} & \textbf{n-Cam}&\textbf{radar}&\textbf{n-2D}&\textbf{n-3D}&\textbf{n-cls} & \textbf{loc} \\ \hline
KITTI~\citep{2012Are}               & 2012          & 1                    & 64              & 4             & No                   & 80K               & 80K               & 8                & KA \\
ApolloScape~\citep{apolloscape}         & 2018          & 2                    & 1              & 6             & No                    & 2.5M              & 70K               & 35               & 4x China \\
H3D~\citep{h3d}                 & 2019          & 1                    & 64              & 3             & No                    & -                 & 1.1M              & 8                & SF \\
nuScenes~\citep{nuScenes}            & 2019          & 1                    & 32              & 6             & Yes                   & -                 & 1.4M              & 23               & Boston, SG \\ 
Argoverse~\citep{argoverse}           & 2019          & 2                    & 32              & 9             & No                    & -                 & 993K              & 15               & PT, Miami \\
Waymo~\citep{waymo}               & 2019          & 5                    &64  & 5             & No                    & 9.9M              & 12M               & 4                & 3x USA \\
AIODrive~\citep{aiodrive}         & 2021          & 4                    &1280 &10             & Yes                   & 26M               & 26M               & 23               & synthetic \\\hline
\end{tabular}
}
\end{table*}
Datasets are an integral part of the field of deep learning. 
The availability of large-scale image datasets such as ImageNet, PASCAL, and COCO motivate outstanding evolution of image classification task~\citep{2009ImageNet, everingham2010pascal, COCO}. 
Benefiting from the vigorous development of 2D images, 3D object detection eagerly requires plentiful labeled data to adapt to a changeable environment.
Consequently, we discuss some widely used datasets for 3D object detection in autonomous driving.
\subsection{KITTI}
One of the earliest datasets for autonomous driving, KITTI~\citep{2012Are}, provides stereo color images, LiDAR point clouds, GPS coordinates, \emph{etc.} The dataset supports multiple tasks: stereo matching, visual odometry, 3D tracking, 3D object detection, \emph{etc.}\footnote{\url{http://www.cvlibs.net/datasets/kitti/index.php}} 
It collects data with a car equipped with a 64-channel LiDAR, 4 cameras, and a combined GPS/IMU system. 
There are over 20 scenes covering cities, residential and roads in the dataset. 
In particular, the object detection dataset contains 7,481 training and 7,518 testing frames with calibration information and annotated 2D/3D bounding boxes. KITTI annotates 8 different classes. Each class is categorized as ``easy'', ``moderate'' and ``hard'' cases. 

mAP (mean Average Precision) is a commonly used metrics in object detection. Some datasets containing multiple classes usually average the AP (Average Precision) score of each class, denoted as mAP.
KITTI requires detection results of ``car'', ``pedestrian'' and ``cyclist'' and calculates mAP of each class.
It considers a predicted box as true positive (TP) if the IoU with the ground-truth box is greater than the threshold, otherwise as false positive (FP). 
The not detected ground-truth boxes are denoted as false negative (FN). We define \textit{precision} and \textit{recall} as:
\begin{equation}
precision = \frac{TP}{TP+FP},
\label{eq:precision}
\end{equation}
\begin{equation}
recall = \frac{TP}{TP+FN}.
\label{eq:recall}
\end{equation}
Based on the predicted and ground-truth boxes, we get a function of $p(r)$ with respect to recall $r$, the calculation of Average Precision (AP) is as below:
\begin{equation}
AP =  \int_{0}^{1} p(r) dr.
\label{eq:ap}
\end{equation}

Remarkably, in order to facilitate the development of multi-modal detection methods in autonomous driving, the KITTI development team proposes a dataset KITTI360~\citep{kitti360} with richer sensor information and 360$^{\circ}$ annotations. 
Specifically, they annotate 3D scene elements with rough bounding primitives and then transfer this information into the image domain. As such, KITTI360 has dense semantic and instance annotations for both 3D point clouds and 2D images.
\subsection{NuScenes}
Developed by Motional, the nuScenes dataset is one of the largest datasets with ground-truth labels for autonomous driving~\citep{nuScenes}. 
It consists of 700 scenes for training, 150 scenes for validation, and 150 scenes for testing. 
The dataset is collected using six cameras and a 32-beam LiDAR to provide 3D annotations for 23 classes in a 360-degree field of view. NuScenes also provides 5 radar sensors for the  measurement of the object velocity.
The full dataset includes approximately 1.4M camera images, 390k LiDAR sweeps, 1.4M radar sweeps, and 1.4M object bounding boxes in 40k key frames. The driving scenes are collected in Boston and Singapore, which are known for their dense traffic and highly challenging driving situations. Additionally, nuScenes annotates object-level attributes such as visibility, activity, pose, \emph{etc}.

As far as the object detection task\footnote{\url{https://www.nuscenes.org/object-detection?externalData=all&mapData=all&modalities=Any}} is concerned, the nuScenes requires the detection of 10 classes, including traffic cone, bicycle, pedestrian, car, bus, \emph{etc.} When calculating AP for a class, instead of adopting the traditional bounding box overlap, nuScenes employs center-distance-based metrics. When matching the prediction and ground-truth, nuScenes computes their center distance and obtains AP based on a list of distance thresholds. mAP is calculated by averaging AP.

Unlike KITTI, nuScenes also considers TP's average translation, scale, orientation, velocity, and attribute error with ground-truth, marked as ATE, ASE, AOE, AVE, and AAE, respectively. The final metric, nuScenes detection score (NDS), is derived from a weighted sum of mAP and errors, leading to a more comprehensive description of detection performance. 
We give the official formula below:
\begin{align}
    \mathrm{NDS}=\frac{1}{10}\left[5 \mathrm{mAP}+\sum_{\mathrm{mTP} \in \mathrm{TP}}(1-\min (1, \mathrm{mTP}))\right].
\end{align}
\subsection{Waymo Open Dataset}
The Waymo Open Dataset\footnote{\url{https://waymo.com/intl/en_us/open}} is a high-quality annotated multi-modality dataset for autonomous driving~\citep{waymo}. It consists of annotated data collected by Waymo's self-driving vehicles. The dataset covers a wide variety of scenes from urban to suburban areas. 
There are a total of 798 scenes for training and 202 scenes for validation with 2D and 3D annotated labels, which are collected by five LiDAR sensors and five pinhole cameras. Each scene captures 20 seconds of continuous driving. The annotations provide four object categories including ``car'', ``pedestrian'', ``cyclist'' and ``sign''.

Same as the KITTI dataset, Waymo Open Dataset adopts AP as the metric. Waymo Open Dataset also proposes a new metric APH which incorporates the heading accuracy of the predicted objects into the traditional AP metric.
Waymo also supports the task of domain adaptation. Domain adaptation is a popular technology that learns knowledge from the source domain with sufficient annotations and transfers it to the target domain with limited or no annotations, which mitigates the lack of huge amount of labeled data~\citep{wang2018deep}. 
\subsection{Other Datasets}
In addition to the three widely used datasets introduced above, there are a few recent datasets that are gaining rapid popularity~\cite{apolloscape,h3d,argoverse,Argoverse2,aiodrive,a_star_dataset,cityscapes_3d,lyft_dataset}. We select some of them for detailed introduction as follows.
\begin{itemize}
	\item  ApolloScape~\citep{apolloscape} consists of data from 4 regions in China under various weather conditions. ApolloScape dataset is collected with 2 LiDAR sensors, 6 video cameras, and a combined IMU/GNSS system. 
	It supports a variety of autonomous driving tasks such as scene parsing, lane segmentation, trajectory prediction, object detection, tracking. The dataset contains 140K+ annotated images with annotation of lane lines. For 3D object detection, ApolloScape provides 6K+ point cloud frames with annotated 3D bounding boxes. ApolloScape's evaluation metrics are the same as KITTI. It requires the detection of vehicles, pedestrians, and bicyclists. 

	\item \vspace{4pt} H3D~\citep{h3d}is a large-scale full-surround 3D object detection and tracking dataset, with a special focus on crowded traffic scenes in the urban areas. The dataset is collected with 3 cameras with 260-degree field of view (FoV), and a 64-beam Velodyne LiDAR sensor. It contains over 27K frames in 160 scenes with over 1 million objects.
	For evaluation, H3D employs a similar protocol as KITTI with a 0.5 IoU threshold for car and a 0.25 IoU threshold for pedestrian. 

	\item \vspace{4pt}  Argoverse~\citep{argoverse,Argoverse2} supports advancements in 3D tracking, motion forecasting, and other perception tasks for self-driving vehicles.
	It provides rich semantic annotation for maps. For sensor setup, it is equipped with two 32-channel LiDARs, seven surround-view cameras, and two stereo cameras. It provides rich semantic information about road infrastructure and traffic rules. Argoverse dataset also provides HD maps for automatic map creation.
	
	\item  \vspace{4pt} Cityscapes 3D~\citep{cityscapes_3d}  extends the original Cityscapes dataset~\citep{cordts2016cityscapes} with 3D bounding box annotations to support the task of 3D vehicle detection. It also provides benchmarks of pixel-level or instance-level semantic labeling and panoptic semantic labeling tasks. It annotates 3D bounding boxes and corresponding 2D instance segmentation masks for each vehicle. The 3D bounding boxes are annotated with stereo RGB images and with nine degrees of freedom.
	It also proposes a new metric for monocular 3D objection detection.

	\item  \vspace{4pt} AIODrive~\citep{aiodrive} is a large-scale synthetic dataset generated by the urban driving simulator, namely CARLA~\citep{carla}. It synthesizes data from multiple sensors including 3D LiDARs, RGB cameras, depth cameras, radars, and IMU/GPS. All sensors collect data at a frequency of 10Hz. With the help of the simulator, it provides petty detailed annotation with the object's 2D/3D bounding boxes, trajectories, velocities, and accelerations. The dataset also synthesizes some adverse scenes such as terrible weather and car accidents.

\end{itemize}

\subsection{Discussion}
\begin{figure}[t]
	\centering
	\includegraphics[width=0.8\linewidth]{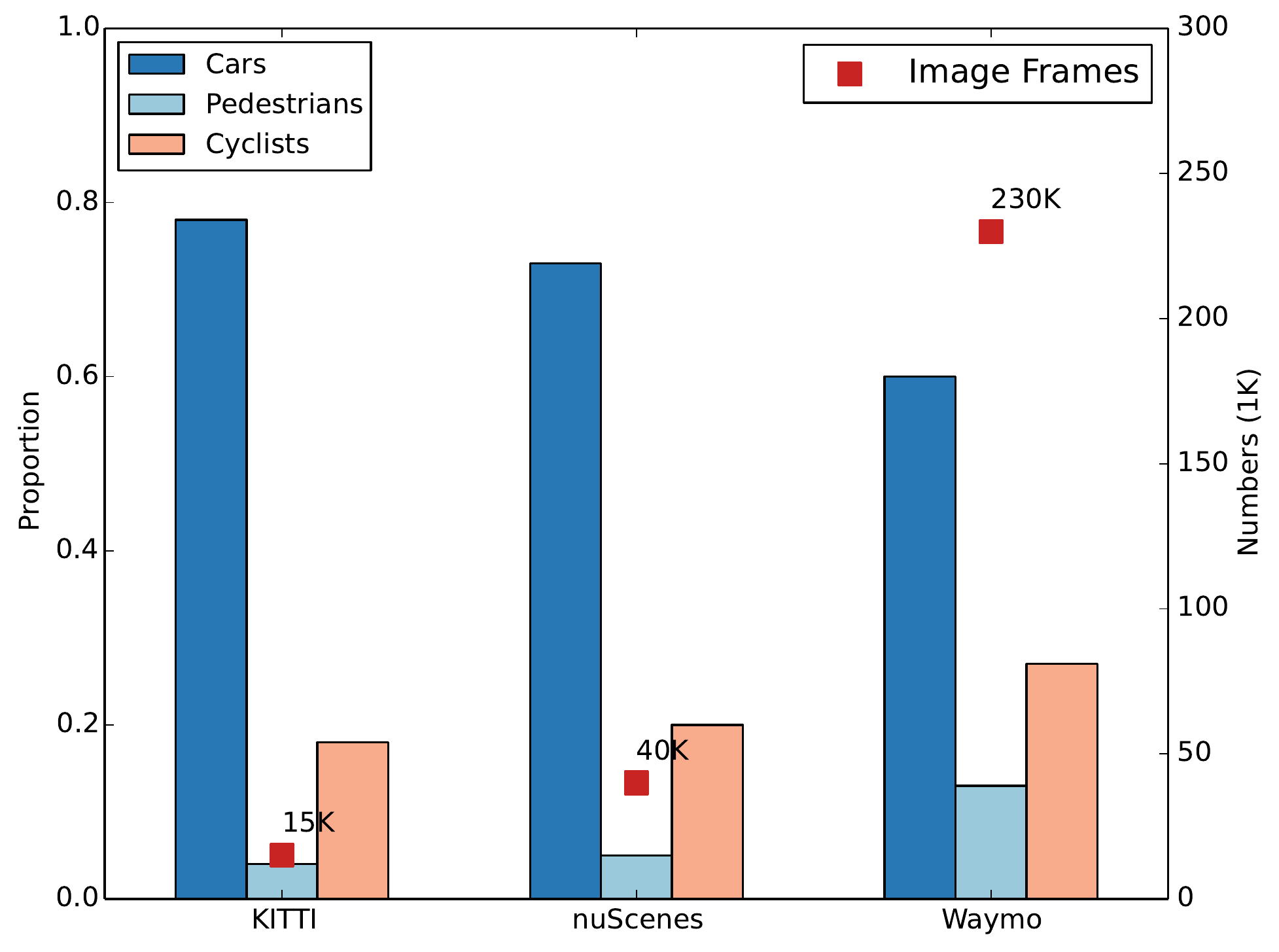}
	\vspace{-0.1cm}
	\caption{Comparison of KITTI, nuScenes, and Waymo Open Dataset. From left Y-axis, we find the proportions of objects belonging to ``car'', ``person'', and ``cyclist'' classes are imbalanced clearly. From right Y-axis, we mark the total image frame number of the three datasets, ranging from 15K to 230K.}
	\label{datasets}
\end{figure}

Datasets for autonomous driving are developing rapidly.
From Fig.~\ref{datasets}, we observe that the size of the three popular datasets ranges from only 15,000 frames to over 230,000 frames. 
However, compared to the image datasets in 2D computer vision, 3D datasets are still relatively small. For example, ImageNet~\citep{2009ImageNet} provides image frames of over 1.4 million.
Besides, the object classes are limited and unbalanced. 
Fig.~\ref{datasets} compares the percentages of car, person, and cyclist classes.
We also make a comprehensive comparison for all discussed datasets in Tab.~\ref{tab:datasetcomparison}.

\begin{figure}[t]
  \centering
  \includegraphics[width=\linewidth]{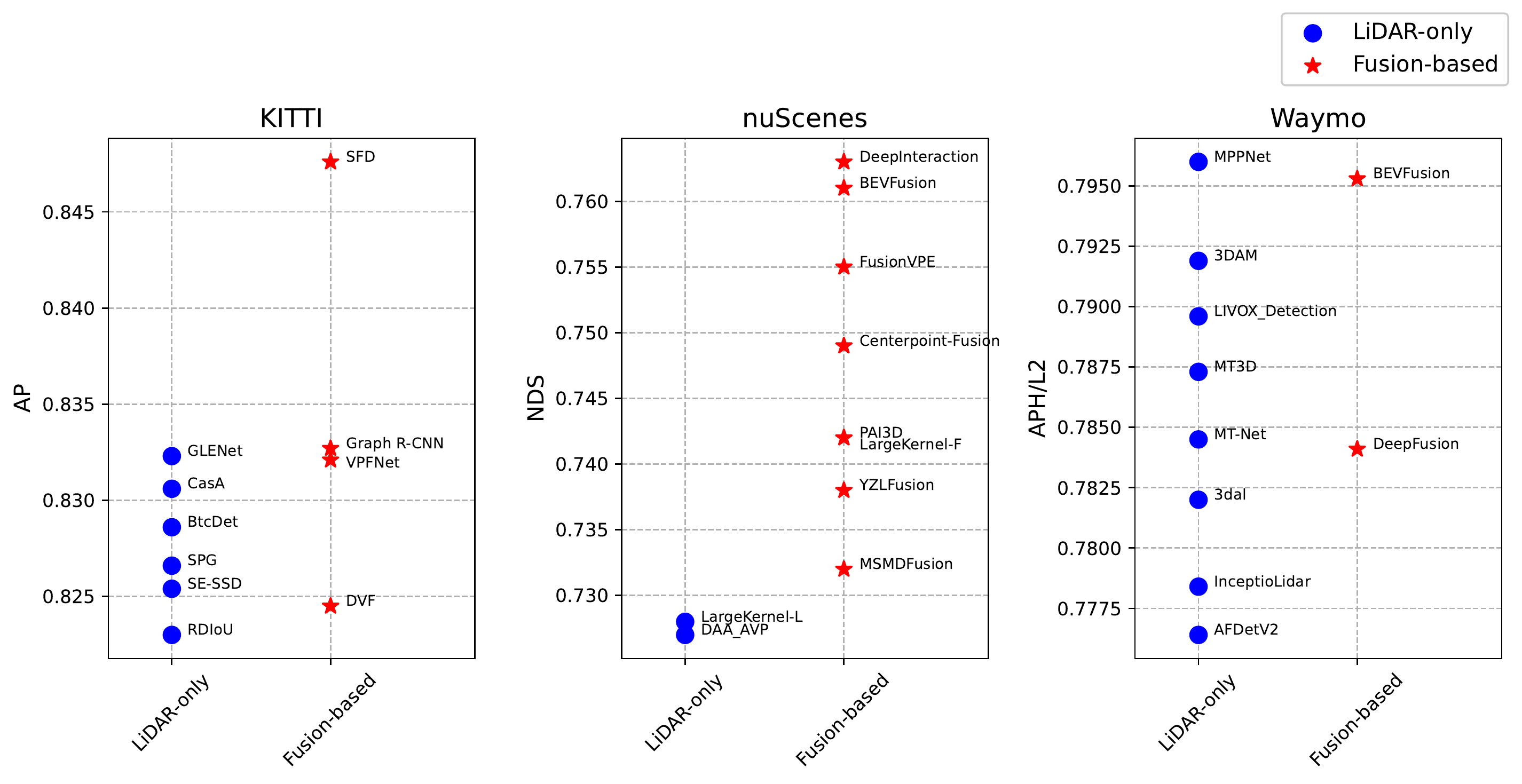}
  \vspace{-0.4cm}
  \caption{Top-10 methods of the popular datasets. Among the top-10 methods, 8 methods are fusion-based on the nuScenes leaderboard, 4 methods are fusion-based on KITTI, and 2 methods are fusion-based on Waymo, respectively. Note that we only report methods with paper link on the KITTI leaderboard, while for the nuScenes and Waymo leaderboards, we report all the listed methods except repeated entries. 
  }
  \label{fig:dataset_sota}
\end{figure}

Fig.~\ref{fig:dataset_sota} shows top-ranked methods on the three datasets.
Interestingly, we observe that top-ranked methods on the nuScenes leaderboard are mainly fusion-based methods~\citep{yang2022deepinteraction, liu2022bevfusion, chen2022scaling, jiao2022msmdfusion}. For example, the top 8 methods on the nuScenes leaderboard are all fusion-based methods.
In contrast, out of the top 10 methods on the KITTI / Waymo leaderboard, only 4 / 2 of them are multi-modal based~\citep{wu2022sparse, yang2022graph, zhu2021vpfnet, mahmoud2022dense, liu2022bevfusion, li2022deepfusion}.
The main reason is that the LiDAR sensors employed in these datasets have different resolutions. 
KITTI and Waymo employ a spinning LiDAR sensor of 64 beams, while nuScenes uses a rotating 32-beam LiDAR. We may infer that multi-modal methods are much more necessary for high-performance perception when point clouds are relatively sparse.

\section{Deep Learning Based Multi-Modal 3D Detection Networks}
\label{sec:multi-moda}
\begin{figure}[t]
  \centering
  \vspace{0.3cm}
  \includegraphics[width=0.9\linewidth]{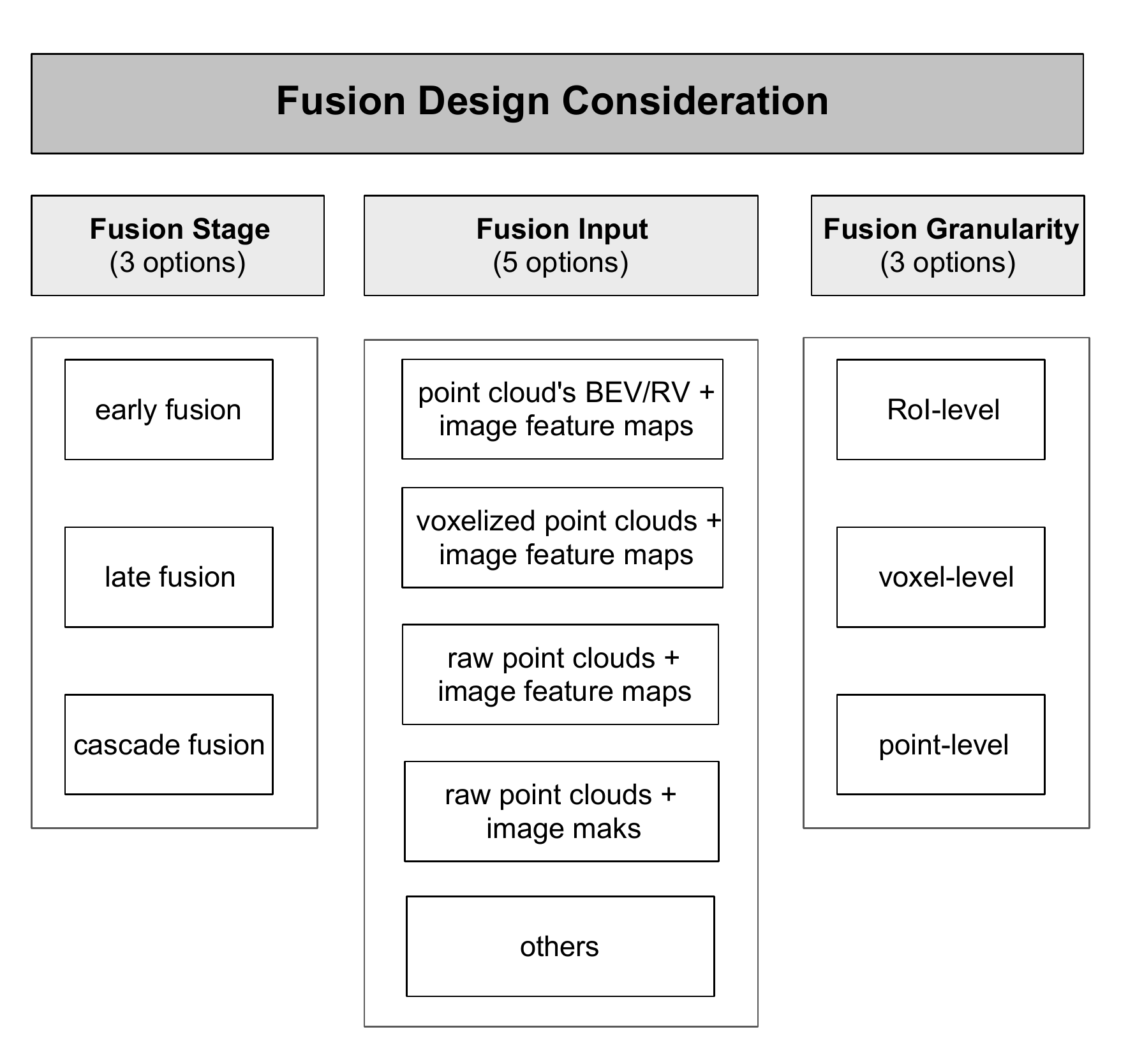}

  \caption{Overview of the three crucial design considerations for a fusion network. We briefly summarize all the options for these considerations here.}
  \label{choices}
\end{figure}

In this section, we present our review of deep learning based multi-modal 3D detection networks, with a special focus on LiDAR and camera data. We organize our review by the following three important design considerations for fusion: 
\begin{itemize}
\item fusion stage, \emph{i.e.}, at what pipeline stage the fusion occurs, answering the question ``where to fuse''; 
\item fusion input, \emph{i.e.}, what representations are used for fusion data, answering the question ``what to fuse''; 
\item fusion granularity, \emph{i.e.}, at what granularity the fusion data are combined, answering the question ``how to fuse''.
\end{itemize}
Fig.~\ref{choices} lists the possible options for each design consideration.
In the rest of this section, we discuss how the recent deep multi-modal 3D detection networks address these three design questions. 
\begin{figure*}[t]
  \centering
  \includegraphics[width=0.9\linewidth]{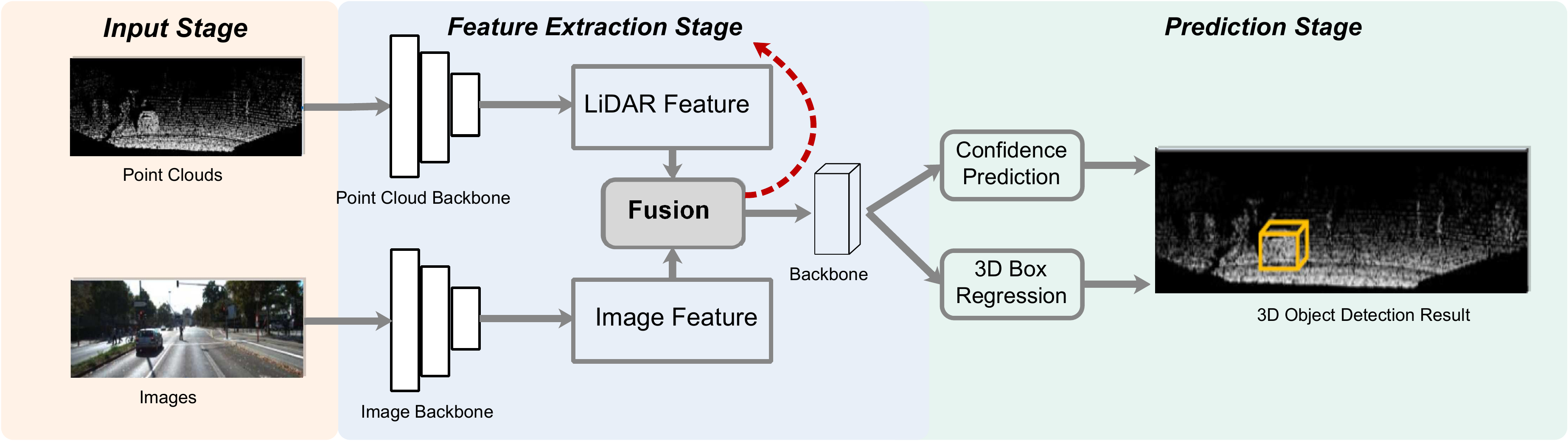}
  \caption{An early fusion pipeline. We first extract the image and point cloud features respectively, and then conduct fusion on these features in the feature extraction stage.} 
  \label{early fusion}
\end{figure*}

\begin{figure*}[t]
  \centering
  \includegraphics[width=0.9\linewidth]{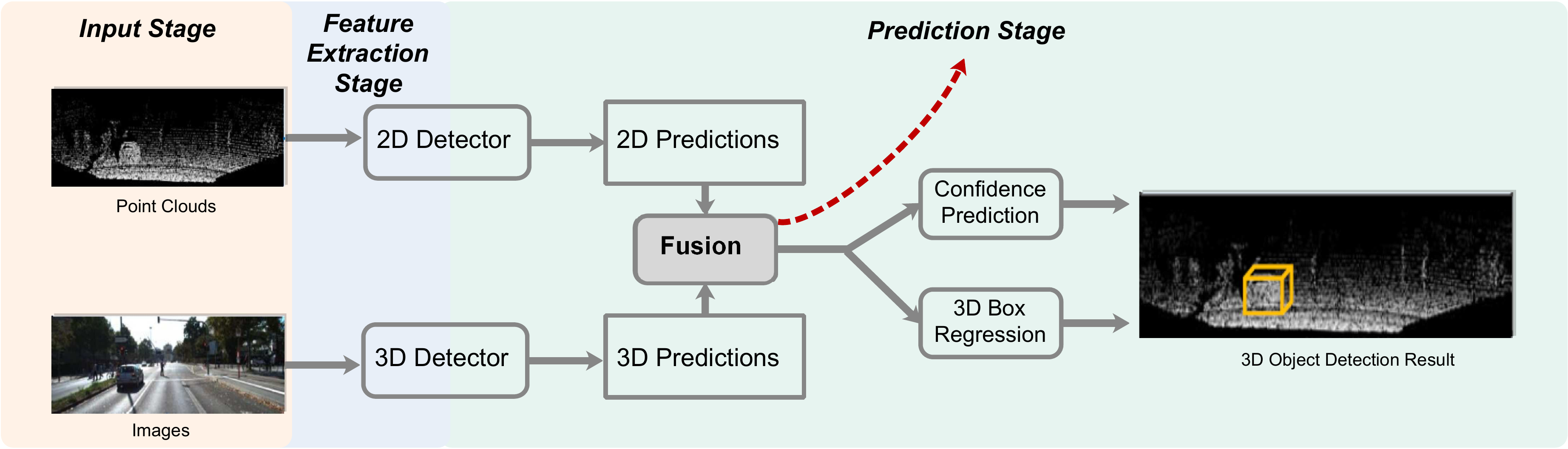}
  \caption{A late fusion pipeline. We first get the predictions from each modality, and then take these predictions as fusion inputs. As such, the fusion network is in the prediction stage. }
  \label{late fusion}
\end{figure*}
More importantly, compared to the other two design considerations, a fusion module's input exhibits the most diversity and represents the characteristic network design. Hence, we categorize the fusion based 3D detection networks into a total of five categories. In each category, we review the relative fusion schemes in detail.
\subsection{Fusion Stage: where to fuse?}
This design issue is concerned with which pipeline stage performs the fusion operation. Here, we broadly partition a detection network pipeline into the following three stages: the input stage, the feature extraction stage, and the prediction stage (illustrated in Fig.~\ref{early fusion}). Depending upon which of the three stages perform fusion, we have three options: early fusion, late fusion, and cascade fusion. 
\subsubsection{Early Fusion}
Early fusion usually occurs in the input stage or feature extraction stage before each branch reaches its prediction~\citep{2017Multi,9156790,8578131}.
It enables more direct interactions among multi-modal features of the intermediate layers, as shown in Fig.~\ref{early fusion}. 
The fused feature is then utilized to perform classification and regression tasks in the prediction stage. 
Early fusion can better leverage rich intermediate information from modalities, and is currently the most widely used fusion stage.

\begin{figure*}[t]
  \centering
  \includegraphics[width=0.9\linewidth]{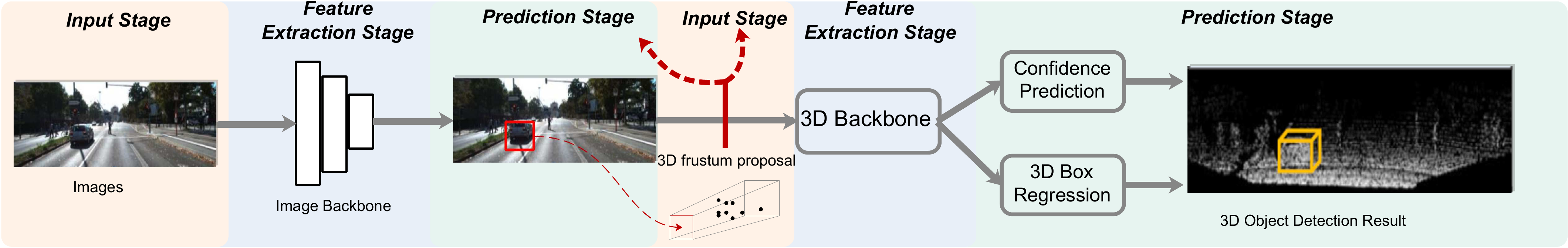}
  \caption{A cascade fusion network. We first obtain the predictions from the image branch, and then fuse the image predictions with point cloud data for further 3D object detection. Therefore, cascade fusion usually occupies the hybrid mode by fusing one branch’s prediction with the other’s input. }
  \label{cascade fusion}
\end{figure*}

\subsubsection{Late Fusion}
In contrast with early fusion, late fusion employs separate branches for each modality, and then combine individual decision-level outputs through a fusion network in the prediction stage~\citep{Review1}. 
Fig.~\ref{late fusion} outlines such a framework.
Late fusion can better leverage existing networks for each modality. 
It also does not require to deal with issues such as how to align the data from different modalities.

Notably, \citet{CLOCs} employ late fusion and outperforms single modality detectors on the KITTI leaderboard.
It exploits the geometric and semantic consistencies between 2D and 3D predictions and learns the probabilistic dependencies between the two from the training data. Specifically, it obtains 2D and 3D proposals and then encodes all proposals into a sparse tensor. 
As summarized in Tab.~\ref{Advantages and Disadvantages}, its shortcoming lies in the inability to exploit rich intermediate features~\citep{latefusion,7487370}.

\subsubsection{Cascade Fusion}
Cascade fusion employs the hybrid mode by fusing one branch's prediction with the other's input, which builds a cascade relationship between multiple modalities.
As illustrated in Fig.~\ref{cascade fusion}, we first obtain 2D proposals from the prediction stage of the image stream. Next, with a known camera projection matrix, a 2D proposal can be lifted to a frustum which defines a 3D search space~\citep{8578200}. We collect all points within the frustum to form a 3D frustum proposal that is used to classify and locate the object. Consequently, one modality provides prior information that can greatly reduce the other's search space in cascade fusion. 

The first fusion approach using the cascading structure was F-PointNet~\citep{8578200}. 
Nevertheless, its performance is greatly limited by the accuracy of the 2D detector. 
Subsequently, several following methods have been proposed~\citep{RoarNet,FConvNet,DBLP:journals/corr/abs-1812-05276} to further improve the accuracy.

\begin{table*}[h]
  \caption{Advantages and disadvantages for different fusion stages}
  \label{Advantages and Disadvantages}
  \renewcommand\arraystretch{1.3}
  \begin{tabularx}{\textwidth}{lXXX}
    \hline
    \multicolumn{1}{c}{\textbf{Categories}} &\multicolumn{1}{c}{\textbf{Advantages}} & \multicolumn{1}{c}{\textbf{Disadvantages}}\\
    \hline
    \multirow{4}{*}{Early Fusion} & 
    + Can leverage rich intermediate features from multiple modalities. \newline + Large feature vectors can lead to  better detection results with suitable learning methods.
    & - Sensitive to inherent data misalignment between modalities.  \newline - Large feature vectors lead to longer training/inference time.\\
    \hline
    \multirow{1}{*}{Cascade Fusion} & 
    + Can reduce the search space with prior information. & - Rely heavily on initial proposal generation.\\
    \hline
    \multirow{1}{*}{Late Fusion} & 
    + Can utilize off-the-shelf networks for each modality.  & - Unable to take advantage of useful intermediate features.\\
    \hline
  \end{tabularx}
\end{table*}

\subsubsection{Discussion}
Tab.~\ref{Advantages and Disadvantages} summarizes the advantages and disadvantages of the three fusion stages, and Tab.~\ref{methods summary} gives typical multi-modal methods for each fusion stage. From Tab.~\ref{methods summary}, we observe that most fusion-based algorithms employ early fusion, which is also the focus of our survey.

\begin{table*}[htbp]
    \centering
    \caption{Summary of typical multi-modal 3D detection methods: stage (fusion stage), PC-Input (point cloud Input), RGB-Input (image input), gran (fusion granularity), HW (hardware), lat (latency), DS (dataset used for evaluation), and mAP (mean average precision)}
    \label{methods summary}
    \renewcommand\arraystretch{1.2}
\resizebox{\textwidth}{!}{
    \begin{tabular}{lllllllll}
    \toprule
          & \textbf{stage} & \textbf{PC-Input} & \textbf{RGB-Input} & \textbf{gran} & \textbf{HW} & \textbf{lat} & \textbf{DS} & \textbf{mAP} \\
    \midrule
    MV3D~\citep{2017Multi}  & \multirow{17}[2]{*}{\cellcolor[rgb]{ .949,  .949,  .949}early} & \multirow{5}[2]{*}{view} & \multirow{5}[2]{*}{feature map} & RoI   & Titan X  & 0.36s & KITTI & 63.63\% \\
    AVOD~\citep{8594049}  & \cellcolor[rgb]{ .949,  .949,  .949} &       &       & RoI   & Titan XP & 0.08s & KITTI & 71.76\% \\
    Contfuse~\citep{Contfuse} & \cellcolor[rgb]{ .949,  .949,  .949} &       &       & voxel & GTX1080 & 0.06s & KITTI & 68.78\% \\
    SCANet~\citep{DBLP:conf/icassp/LuCZZMZ19} & \cellcolor[rgb]{ .949,  .949,  .949} &       &       & RoI   & GTX1080 & 0.09s & KITTI & 66.30\% \\
    FuseSeg~\citep{2020FuseSeg} & \cellcolor[rgb]{ .949,  .949,  .949} &       &       & point & -     & -     & KITTI & - \\
    \cline{1-1} \cline{3-9} 
    MVX-Net~\citep{MVXNet} & \cellcolor[rgb]{ .949,  .949,  .949} & \multirow{4}[2]{*}{voxel} & \multirow{4}[2]{*}{feature map} & \multicolumn{1}{p{4.19em}}{voxel} & GTX1080 & -     & KITTI & 72.70\% \\
    3D-CVF~\citep{3D-CVF} & \cellcolor[rgb]{ .949,  .949,  .949} &       &       & voxel \& RoI & - & 0.06s & KITTI & 80.45\% \\
    VPF-Net~\citep{zhu2021vpfnet} & \cellcolor[rgb]{ .949,  .949,  .949} &       &       & \multicolumn{1}{p{4.19em}}{point} & \multicolumn{1}{p{4.19em}}{2080Ti} & 0.06s & KITTI & 83.21\% \\
    PointAugmenting~\citep{PointAugmenting} & \cellcolor[rgb]{ .949,  .949,  .949} &       &       & point & -     & -     & nuScenes & 66.80\% \\
    \cline{1-1} \cline{3-9}
    PointFusion~\citep{8578131} & \cellcolor[rgb]{ .949,  .949,  .949} & \multirow{3}[2]{*}{point} & \multirow{3}[2]{*}{feature map} & RoI   & GTX1080 & 1.3s  & KITTI & 63.00\% \\
    EPNet~\citep{2020EPNet} & \cellcolor[rgb]{ .949,  .949,  .949} &       &       & point & Titan XP & 0.1s  & KITTI & 81.23\% \\
    \multirow{1}[1]{*}{PI-RCNN~\citep{PI-RCNN}} & \cellcolor[rgb]{ .949,  .949,  .949} &       &       & \multicolumn{1}{l}{\multirow{1}[1]{*}{point \& RoI}} & \multirow{1}[1]{*}{-} & \multirow{1}[1]{*}{0.06s} & \multirow{1}[1]{*}{KITTI} & \multirow{1}[1]{*}{78.53\%} \\
    \cline{1-1} \cline{3-9}
    PointPainting~\citep{9156790} & \cellcolor[rgb]{ .949,  .949,  .949} & \multirow{3}[2]{*}{point} & \multirow{3}[2]{*}{mask} & point & GTX1080 & 0.4s  & KITTI & 75.80\% \\
    CenterPointV2~\citep{yin2021center} & \cellcolor[rgb]{ .949,  .949,  .949} &       &       & point & -     & -     & nuScenes & 67.10\% \\
    HorizonLiDAR3D~\citep{HorizonLiDAR3D} & \cellcolor[rgb]{ .949,  .949,  .949} &       &       & point & -     & -     & Waymo & 78.49\% \\
    \cline{1-1} \cline{3-9}
    \multirow{2}[1]{*}{MMF~\citep{8954034}} & \cellcolor[rgb]{ .949,  .949,  .949} & \multirow{2}[1]{*}{view} & \multirow{2}[1]{*}{\begin{tabular}[c]{@{}l@{}}feature map \& \\  pseudo LiDAR\end{tabular}} & \multirow{2}[1]{*}{point} & \multirow{2}[1]{*}{GTX1080} & \multirow{2}[1]{*}{0.08s} & \multirow{2}[1]{*}{KITTI} & \multirow{2}[1]{*}{77.43\%} \\
          & \cellcolor[rgb]{ .949,  .949,  .949}      &       &       &       &       &       &       &  \\
    MVAF~\citep{MVAF} & \multirow{-17}[2]{*}{\cellcolor[rgb]{ .949,  .949,  .949}early} & \multicolumn{1}{l}{\multirow{1}[1]{*}{voxel \&\newline{}view}} & \multicolumn{1}{l}{\multirow{1}[1]{*}{feature map}} & \multirow{1}[1]{*}{voxel} & \multirow{1}[1]{*}{Titan X} & \multirow{1}[1]{*}{0.06s} & \multirow{1}[1]{*}{KITTI} & \multirow{1}[1]{*}{78.71\%} \\
    \midrule
    \midrule
    F-PointNet~\citep{8578200} & \cellcolor[rgb]{ .851,  .851,  .851} & \multirow{5}[2]{*}{-} & \multirow{5}[2]{*}{-} & \multirow{5}[2]{*}{RoI} & GTX1080 & 0.17s & KITTI & 69.79\% \\
    IPOD~\citep{DBLP:journals/corr/abs-1812-05276}  & \cellcolor[rgb]{ .851,  .851,  .851}      &       &       &       & -     & 0.1s  & KITTI & 72.57\% \\
    F-ConvNet~\citep{FConvNet} &\cellcolor[rgb]{ .851,  .851,  .851}       &       &       &       & -     & 0.1s  & KITTI & 75.50\% \\
    RoarNet~\citep{RoarNet} &\cellcolor[rgb]{ .851,  .851,  .851}       &       &       &       & Titan X & -     & KITTI & 73.04\% \\
    SIFRNet~\citep{DBLP:conf/aaai/ZhaoLHH19} &\multirow{-5}[2]{*}{\cellcolor[rgb]{ .851,  .851,  .851}cascade} &       &       &       & -     & -     & KITTI & - \\
    \midrule
    \midrule
    CLOCs~\citep{CLOCs} & \cellcolor[rgb]{ .749,  .749,  .749}late & -     & -     & -     & -     & -     & KITTI & 82.25\% \\
    \bottomrule
    \end{tabular}%
    }
  \label{tab:addlabel}%
\end{table*}%

\subsection{Fusion Input: what to fuse?}

The second design choice is concerned with what form or representation the multi-modal data are input into the fusion module. 
A fusion module's input exhibits the most diversity and represents the unique idea of each design.
For LiDAR-camera fusion, we are allowed to take raw sensor data, various intermediate features, or even result-level output from the image/point cloud branch as fusion input. 
Specifically, the fusion module can take in the LiDAR data in the form of voxel grids, raw point clouds, or point cloud's projection on BEV or RV, camera data in the form of image feature maps, segmentation masks, or even the corresponding pseudo-LiDAR point clouds.

In this section, we first present typical inputs that can be utilized by the image and point cloud branches respectively, and then we categorize the fusion based 3D detection networks into a total of five categories according to input combinations. Here, we focus our review on early fusion methods.

\subsubsection{Typical Fusion Inputs for LiDAR-Camera Fusion}
\label{Fusion Input Representation}
\begin{figure}[t]
  \centering
  \includegraphics[width=0.9\linewidth]{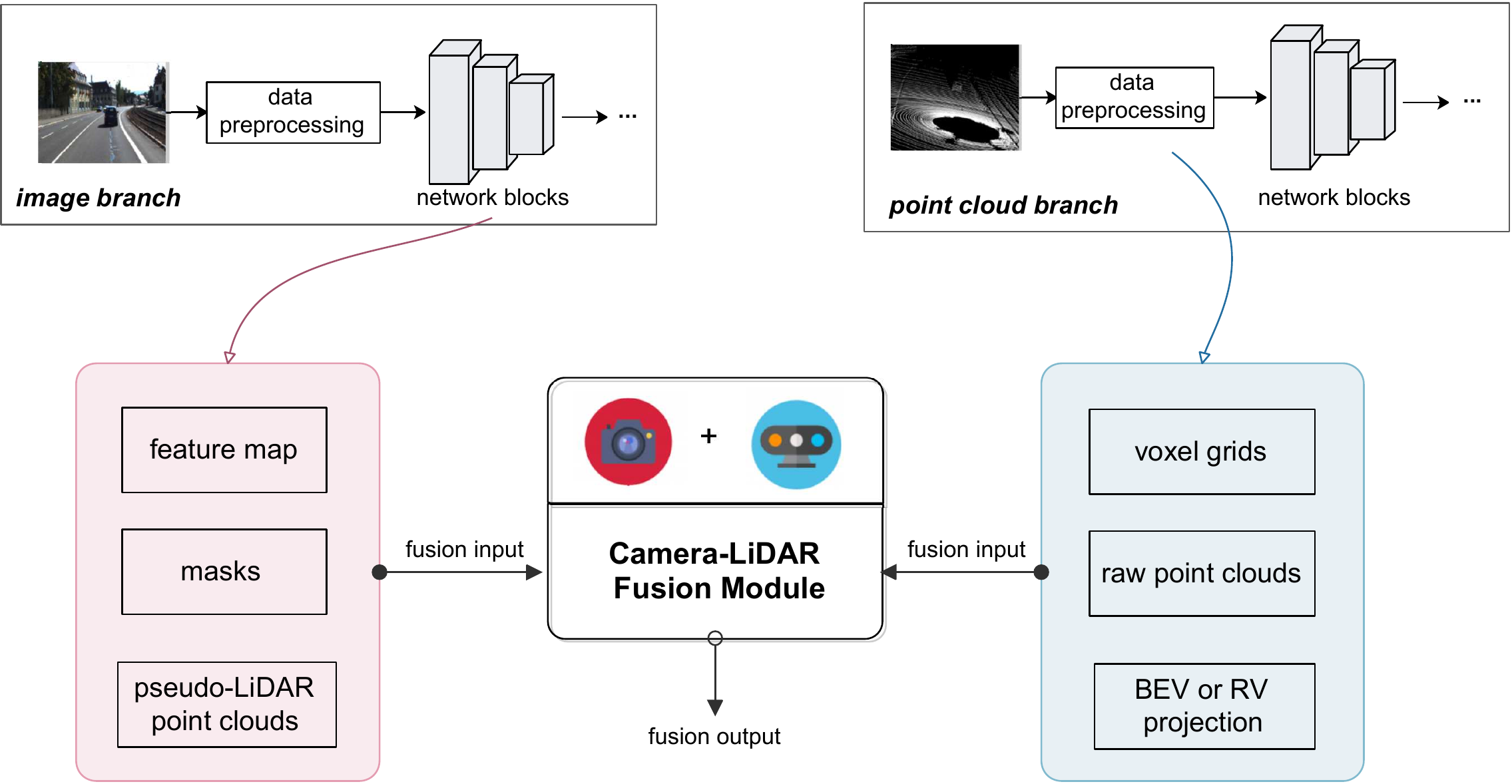}
  \caption{Illustration of typical fusion inputs from the image branch and the point cloud branch, respectively. For the image branch, its fusion input is usually the output of a neural network block; for the point cloud branch, its fusion input is usually from simple data preprocessing such as voxelization.}
  \label{input}
\end{figure}

We first introduce the typical fusion inputs employed for the image branch and the point cloud branch, respectively, in fusion-based detection pipelines, as illustrated in Fig.~\ref{input}. To be more precise, a modality's input to the fusion module is the output of a certain middle layer, be it a simple data preprocessing function or a neural network block.  

\par
\begin{figure*}
\centering
\subfigure[the original image]{
\includegraphics[width=8cm,height=3.8cm]{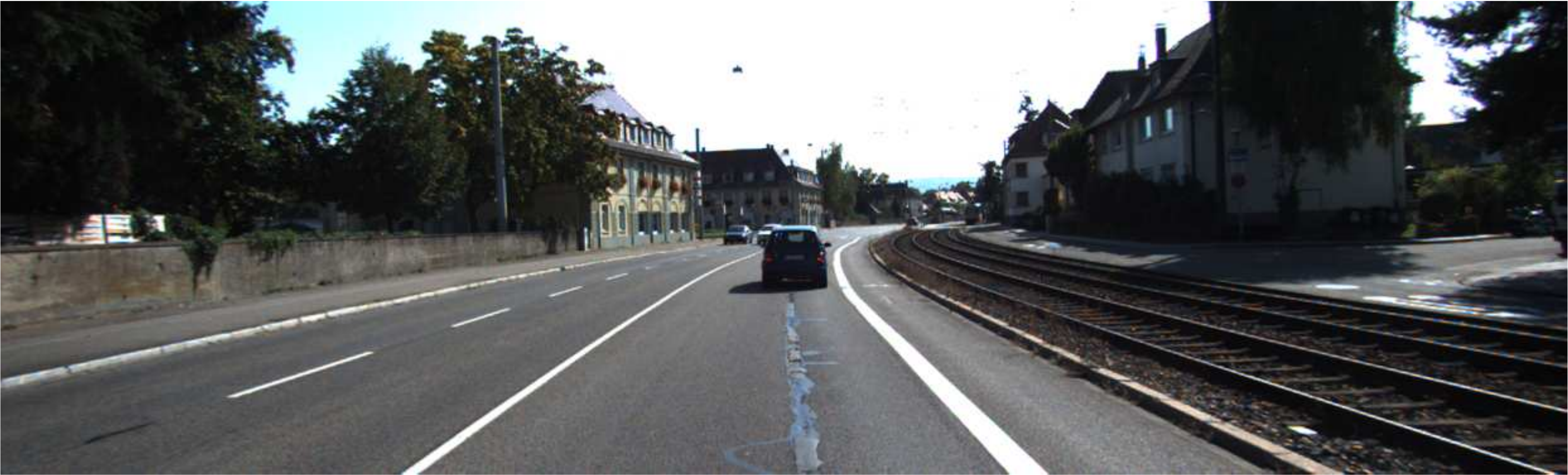}
}
\quad
\quad
\subfigure[the feature map (output of the 5-th Conv2d block)]{
\includegraphics[width=8cm,height=3.8cm]{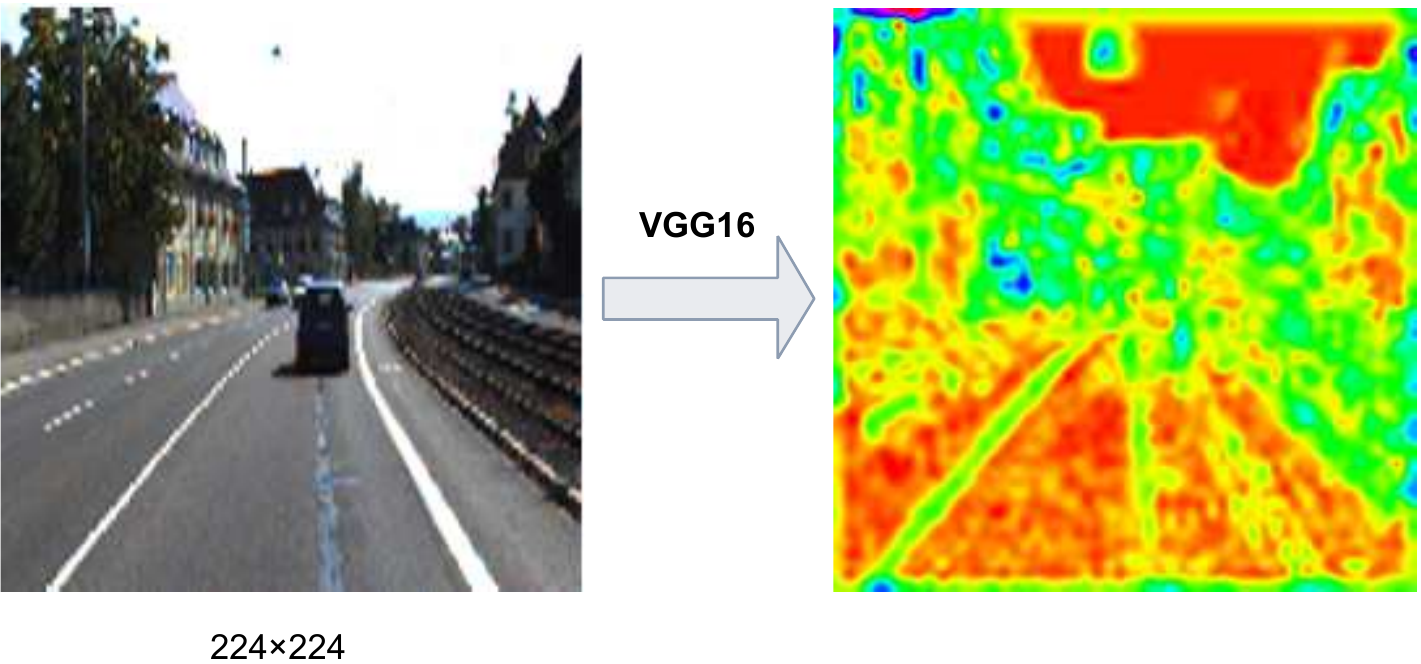}
}
\quad
\quad
\subfigure[the mask (output of mask R-CNN~\citep{DBLP:conf/iccv/HeGDG17})]{
\includegraphics[width=8cm,height=3.8cm]{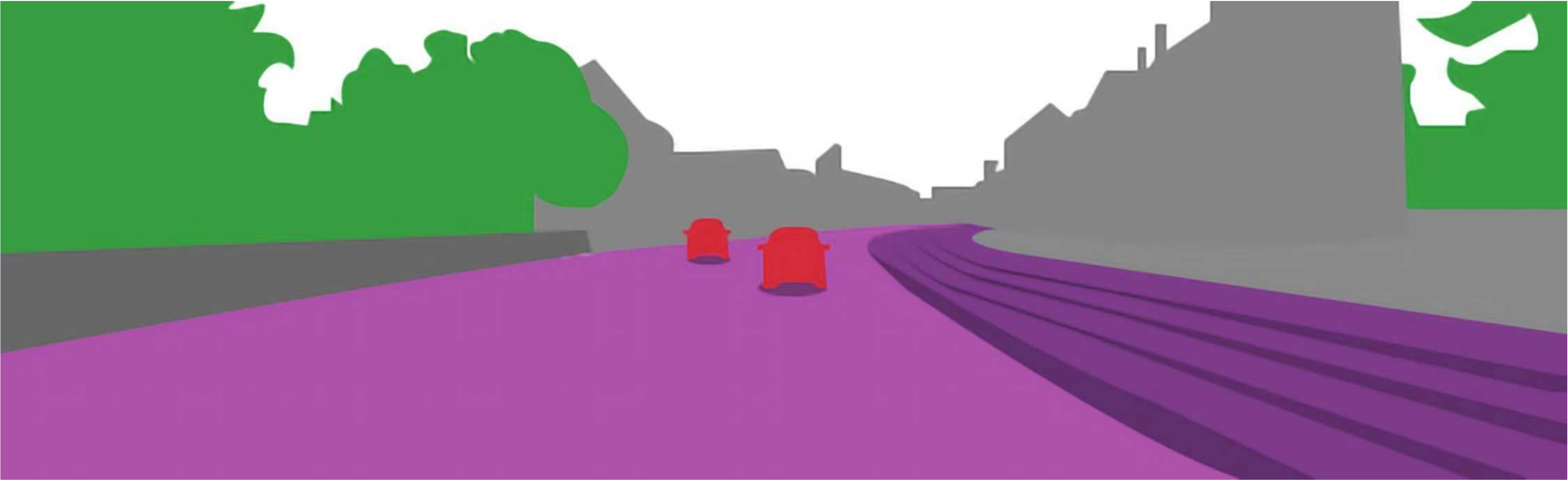}
}
\quad
\quad
\subfigure[the pseudo point cloud (shown in BEV)]{
\includegraphics[width=8cm,height=3.8cm]{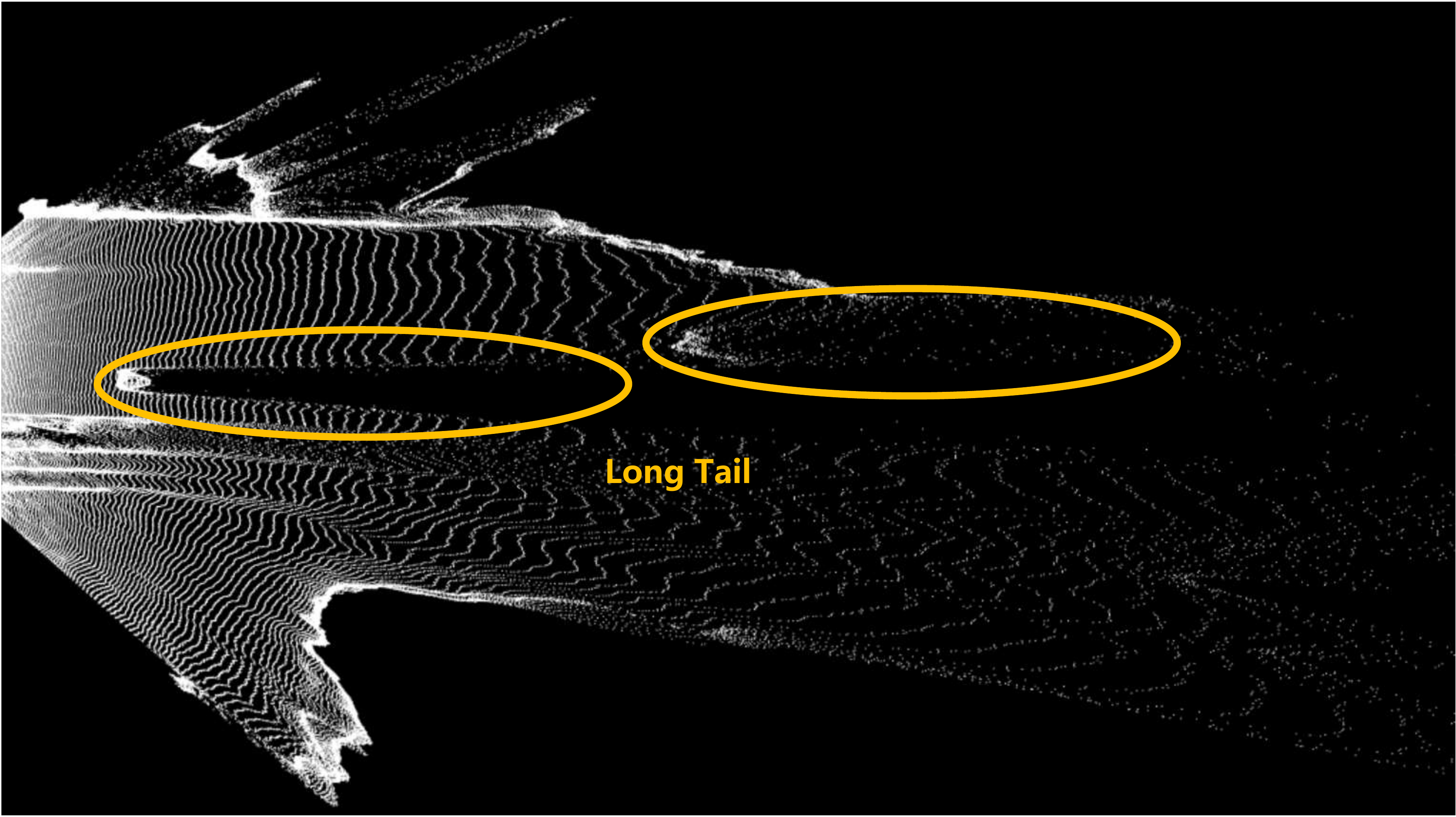}
}
\caption{Different inputs for the image branch in fusion-based 3D object detection. An RGB image (a), one of its feature maps (b), its segmentation mask (c), and its pseudo-LiDAR point cloud's projection on BEV (d). The raw image is taken from the KITTI training set. We use a pretrained VGG16~\citep{Simonyan2014Very} to obtain the feature map on the resized image (224$\times$224).}
\label{image}
\centering
\end{figure*}

\vspace{4pt}\noindent\textbf{Typical Fusion Input for the Image Branch.} 
Most of the LiDAR-camera fusion methods take one of the following three fusion inputs from the image branch, namely feature maps, segmentation masks and pseudo-LiDAR point clouds (depicted in Fig.~\ref{image}). 
\begin{description}
\vspace{4pt}\item[\textbf{feature maps:}] 
Deep neural networks are capable of extracting appearance and geometry feature maps from raw images ~\citep{2014Convolutional,DBLP:conf/nips/KrizhevskySH12,Yosinski2014How}, which are the most commonly used input for fusion between cameras and other sensors~\cite{2017Multi,8594049,Contfuse}.
Compared with raw images (Fig.~\ref{image} (a)), the utilization of feature maps explores richer appearance cues and larger receptive fields, which enables more in-depth and thorough interactions between modalities. 
For example, as illustrated in Fig.~\ref{image} (b), we observe that the edges and textures of feature maps are more distinct than other areas. We refer the readers to Sec.~\ref{dual} for more in-depth review of fusion algorithms that use image feature maps as input.
Here, we list some popular backbones that can be used to obtain feature maps, which can be fed to a fusion module: \emph{e.g.}, VGG-16~\citep{Simonyan2014Very}, ResNet~\citep{7780459}, DenseNet~\citep{densenet}. 

\vspace{4pt}\item[\textbf{masks:}] Images are passed through a semantic segmentation network to obtain pixel-wise segmentation masks~\citep{DBLP:journals/ijcv/GeigerY91,Long2015Fully}. Image masks are often utilized for fusion with other sensor data, as a stand-alone product from the image processing branch. Compared with feature maps, using masks as camera data fusion input has the following advantages. Firstly, image masks can serve as more compact summary features of the image. Secondly, pixel-level image masks can easily be used to ``paint'' or  ``decorate'' the LiDAR points by conducting a point-to-pixel mapping using a known calibration matrix~\citep{9156790}. We refer the readers to Sec.~\ref{dual} for more in-depth review of fusion algorithms that use image masks as input.
Here, we list some popular image segmentation networks for the fusion-based algorithms: \emph{e.g.} DeepLabV3~\citep{Chen2018DeepLab}, Mask-RCNN ~\citep{DBLP:conf/iccv/HeGDG17}, and lightweight network Unet~\citep{Ronneberger2015U}.
\vspace{4pt}\item[\textbf{pseudo-LiDAR point clouds:}] The camera data can also be converted to pseudo point clouds as fusion input~\citep{8954034}. As pointed out in~\citep{Pseudo-LiDAR}, the pseudo point cloud representation raises image pixels to the 3D space, whose signal is much denser than actual LiDAR point cloud. On the downside, it often has a \emph{long tail} problem since the estimated depth may not be accurate around the boundaries of the object~\citep{Pseudo-LiDAR++}, as depicted in Fig.~\ref{image} (d) with yellow circles. According to~\citep{Pseudo-LiDAR}, the pseudo-LiDAR points are obtained by back-projecting image pixels into pseudo 3D points according to the estimated depth map. 
In the context of 3D multi-modal detection, this representation contributes to multi-task learning~\citep{8954034}. Using the pseudo point clouds as fusion input, we can readily facilitate dense feature map fusion between images and point clouds.
\end{description}

\vspace{4pt}\noindent\textbf{Typical Fusion Input for the Point Cloud Branch.}
A LiDAR point cloud is often synthesized from depth measurements collected from different viewpoints. It is basically a set of points in a 3D coordinate system, commonly defined by x, y, z, and the reflection intensity. 
Below, we discuss the typical LiDAR inputs that are commonly used for LiDAR-camera fusion, \emph{i.e.}, voxel grids, raw point clouds, and point cloud's projection on BEV or RV, whose visualization results are shown in Fig.~\ref{point cloud} respectively.
\begin{description}
\item {\textbf{voxelized point clouds or voxel grids}}: A voxelized point cloud is widely utilized as the fusion input due to the efficient parallel processing potentials on a regular voxel grid~\citep{BirdNet,AFDet,Lang2019PointPillars,9018080,PIXOR,8578896,8578570}. 
We first discretize the 3D space into 3D voxel grids, and then obtain the voxel features through the voxel feature encoding (VFE) layer as shown in Fig.~\ref{voxel-based}. We can thus utilize 3D CNNs to extract deeper point cloud features. 
We refer the readers to Sec.~\ref{dual} for more in-depth review of fusion algorithms that employ voxel grids as input for point cloud branch.
However, fusion with voxelized points also has several disadvantages. Firstly, it suffers from information loss. The voxel size is highly correlated with how much spatial information is lost. To illustrate this point, let us look at Fig.~\ref{point cloud} (b) where voxels (in blue color) are much more sparse than the original points in (a). Secondly, during voxelization, a large number of empty voxels will be produced as the LiDAR points are only on the surface of the objects~\citep{DBLP:journals/tog/NiessnerZIS13}, which may adversely affect the fusion performance. Thirdly, processing 3D voxels require time-consuming 3D convolution operations. Accordingly, the training time of the fusion network will inevitably increase. In practice, the point cloud data is usually voxelized into an evenly spaced grid only in the x-y plane (which we call \emph{pillars}) to meet the computation and effectiveness demands~\citep{Lang2019PointPillars}.
\begin{figure}[t]
  \centering
  \includegraphics[width=\linewidth]{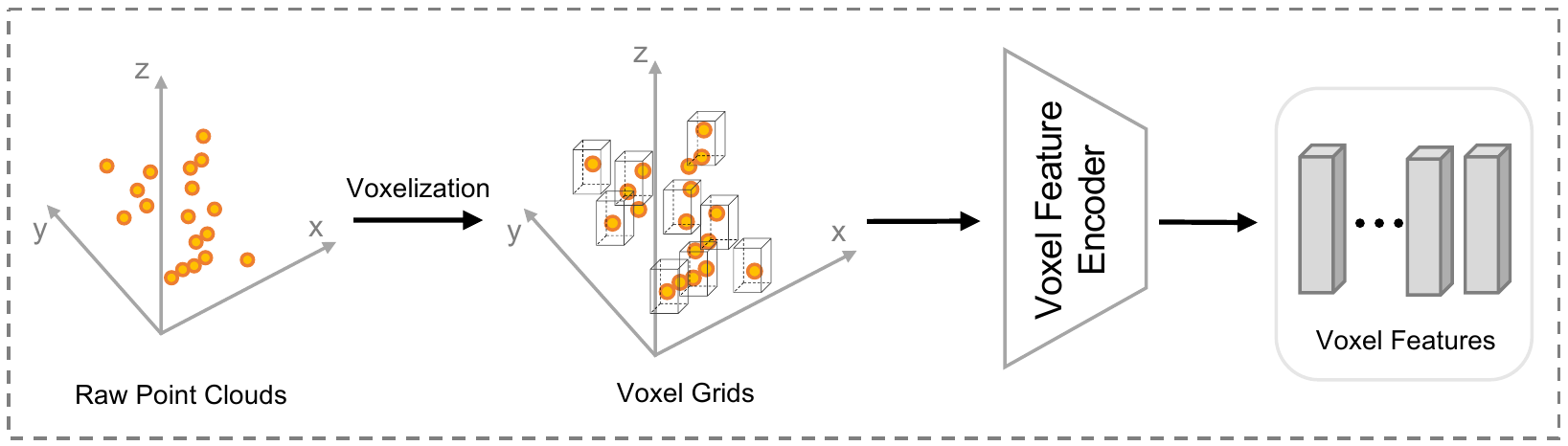}
  \caption{A typical voxelized point cloud processing network~\cite{8578570}}
  \label{voxel-based}
\end{figure}
\vspace{4pt}\item{\textbf{raw point clouds}}: Thanks to efficient 3D point cloud processing networks, the raw 3D point cloud can be directly processed to obtain suitable point features without voxelization loss~\citep{8578131}. 
Specifically, in Fig.~\ref{point-based}, we employ the point cloud encoder~\citep{8099499,ISI:000452649405018} to process raw points and obtain point feature vectors.
Sec.~\ref{dual} provides more detailed review of the fusion algorithms that use raw points as input from the point cloud branch.
Directly taking raw points as input can retain more information, compared with voxel-based methods~\citep{9008567,9157234,9156597}. 
However, point-based methods are generally computationally expensive, especially when dealing with large scenes. For example, for a widely used Velodyne LiDAR HDL-64E, it collects more than 100,000 points per frame (in Tab.~\ref{Point Cloud and Image}). Therefore, considering the efficiency and performance, down-sampling point cloud data appropriately is necessary for data preprocessing.
\begin{figure}[t]
  \centering
  \includegraphics[width=\linewidth]{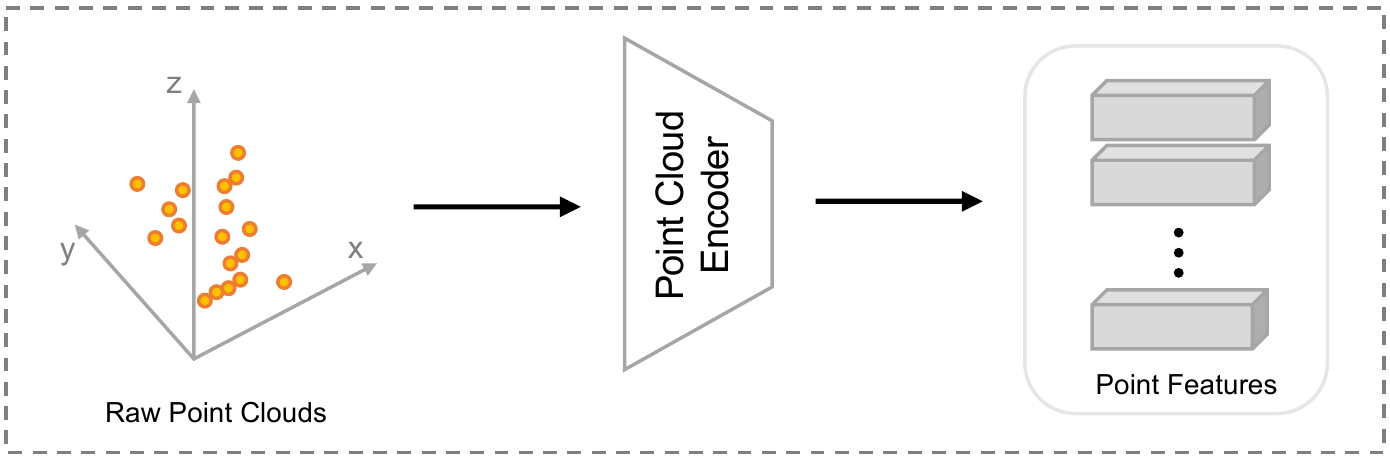}
  \caption{A typical raw point cloud processing network~\citep{8099499}}
  \label{point-based}
\end{figure}

\begin{figure*}
\centering
\subfigure[the raw point cloud]{
\includegraphics[width=8cm,height=3.8cm]{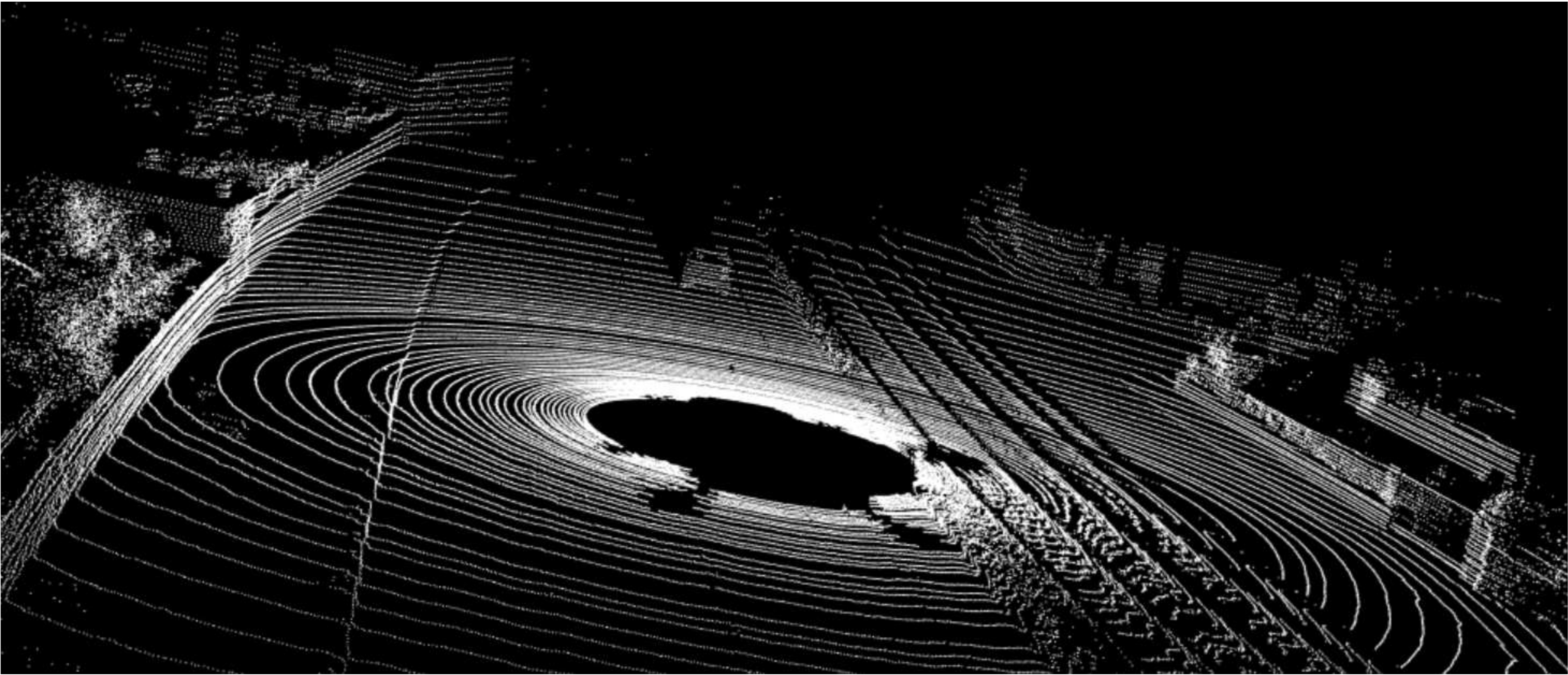}
}
\quad
\quad
\subfigure[the voxel grid with a voxel size of (0.2m, 0.2m, 0.2m)]{
\includegraphics[width=8cm,height=3.8cm]{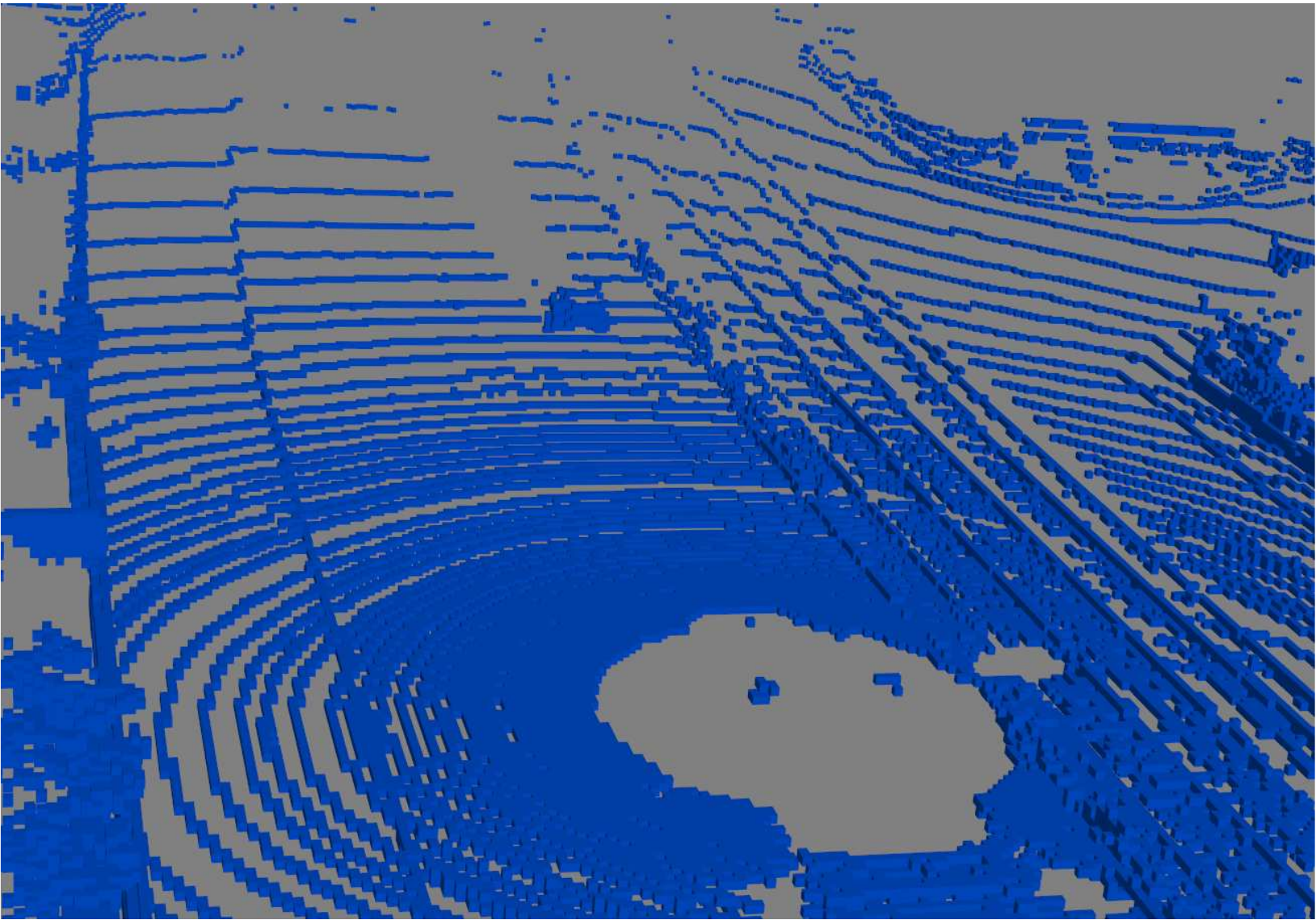}
}
\quad
\quad
\subfigure[the BEV projection]{
\includegraphics[width=8cm,height=3.8cm]{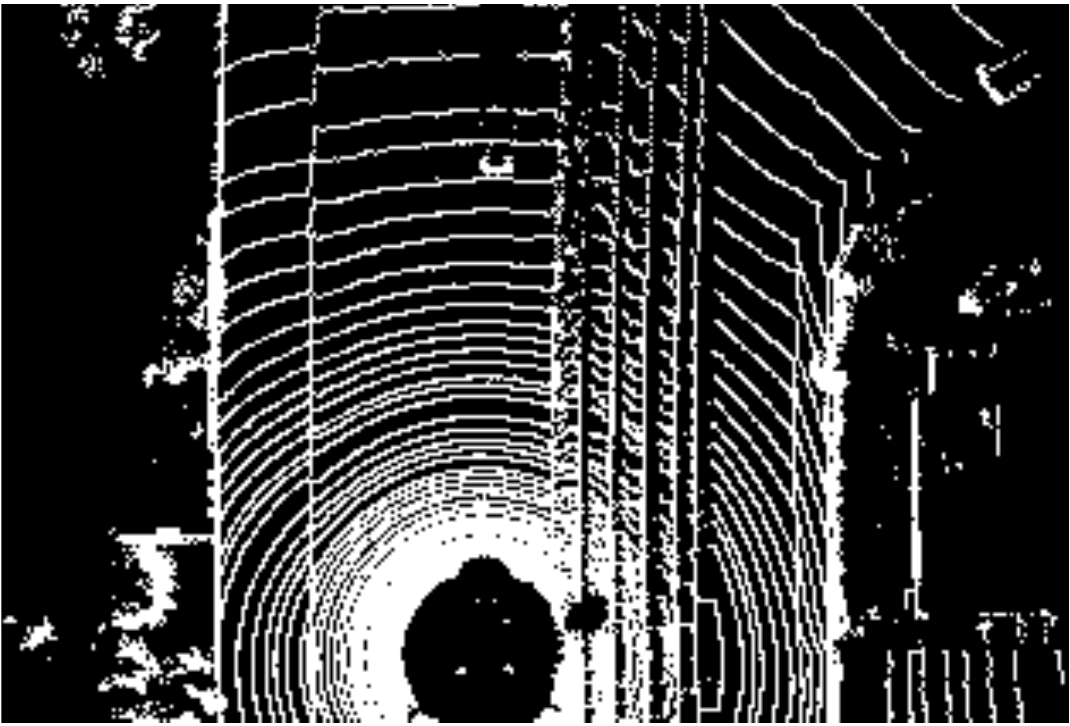}}
\quad
\quad
\subfigure[the RV projection]{
\includegraphics[width=8cm,height=3.8cm]{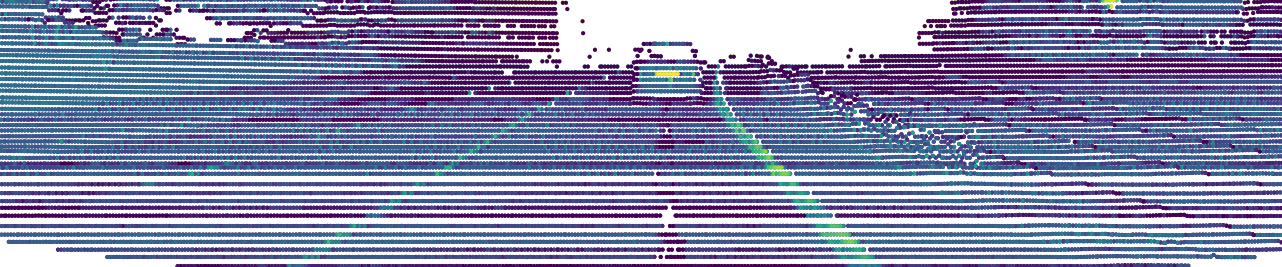}}
\caption{Different inputs for the point cloud branch in fusion-based 3D object detection. A raw point cloud (a), its voxel grids with the voxel size of [0.2m, 0.2m, 0.2m] (b), its projection on BEV (c), and its projection on RV (d). The raw point cloud data is taken the from KITTI training set.}
\label{point cloud}
\end{figure*}
\par
\vspace{4pt}\item{\textbf{BEV or RV projection}}: Another typical point cloud input for the fusion module is point cloud's BEV or RV projection. The resulting pseudo image could be thereby processed efficiently by 2D CNNs.
BEV is commonly adopted to fuse with image features because there is much less overlapping between objects on the BEV plane. 
Another popular view-based input used for fusion is RV, which is also a native representation of the rotating LiDAR sensor~\citep{Wang2018Traffic}. 
Essentially, it forms a compact 2.5D scene~\citep{DBLP:conf/cvpr/HuZHR20} instead of a sparse 3D point cloud.
Projecting the point cloud on RV preserves the full resolution of the LiDAR sensor data and avoiding the spatial loss.
However, RV suffers from the problem of the scale variation between nearby and far away objects~\citep{rangedet}. 
In the fusion pipeline, these mentioned point cloud's projections are usually first processed with 2D CNNs to get view-based features, and then pooled to the same size as the image features.
\end{description}

\subsubsection{Typical Input Combinations for LiDAR-Camera Fusion}
\label{Input Combinations}
In our survey, we find the following combinations of fusion inputs are the most popular for the LiDAR-camera fusion module: (1) point clouds' BEV/RV $+$ image feature maps, (2) voxelized point clouds $+$ image feature maps, (3) raw point clouds $+$ image feature maps and (4) raw point clouds $+$ image masks. 
In addition, exploiting more than one type of inputs for images/point clouds to form a more comprehensive fusion has become a recent trend. We also review these methods here.  Below we discuss the fusion-based methods that fall into one of these 5 fusion input categories, with a special focus on how these fusion input combinations evolve with time and technology.  
\par
\vspace{4pt}\noindent\textbf{point cloud's BEV/RV $+$ image feature maps.}
\label{dual}
Before 3D object detection became popular, 2D object detection based on images had drawn a great deal of attention~\citep{Ren2017Faster}. Therefore, as soon as LiDAR was considered for 3D object detection, several LiDAR-camera fusion algorithms were proposed to project 3D point clouds on a 2D plane, and combine the resulting 2D view of the point cloud with image feature maps. We discuss typical algorithms in this category below. 

\begin{figure}[t]
  \centering
  \includegraphics[width=0.9\linewidth]{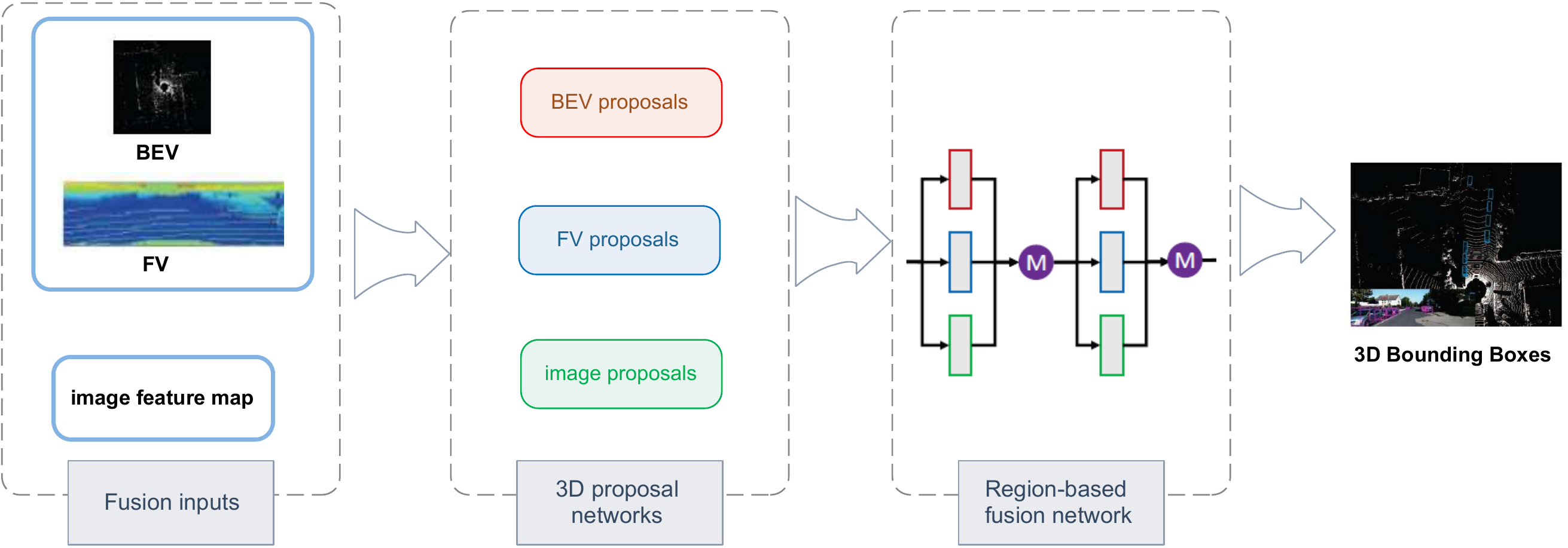}
  \caption{The MV3D pipeline that fuses point cloud's projections and image feature map~\citep{2017Multi}}
  \label{MV3D}
\end{figure}

MV3D~\citep{2017Multi} is a pioneering work in this category. As shown in Fig.~\ref{MV3D}, it takes the FV (front view) and BEV of a point cloud as input and exploits a 3D Region Proposal Network (RPN) to generate 3D proposals. Next, MV3D integrates multi-view features vectors from multi-proposals into the same length and puts them through a region-based fusion network.
AVOD~\citep{8594049} achieves better performance than MV3D, especially in the small object category by designing a more advanced RPN that employs high-resolution feature maps. It also merges features from multiple views in the RPN phase to generate more accurate positive proposals. AVOD only takes point cloud's BEV and image as input, which effectively decreases the computation cost. Based on AVOD, SCANet~\citep{DBLP:conf/icassp/LuCZZMZ19} utilizes an encoder-decoder based proposal network with a Spatial-Channel Attention (SCA) module to capture multi-scale contextual information and an Extension Spatial Upsample (ESU) module to recover the spatial information.

Nevertheless, these methods have limitations, especially when detecting small objects such as pedestrians and cyclists. To overcome these limitations, Contfuse~\citep{Contfuse} performs continuous convolutions~\citep{8578372} to extract multi-scale convolutional feature maps from point cloud's BEV and fuse them with image features. The engagement of continuous convolution captures local information from neighboring observations and leads to less geometric information loss. In addition, another downside of using point cloud's BEV or FV as fusion input lies in the inevitable 3D spatial information loss when projecting the 3D point cloud to the 2D plane. 

The point cloud's range view (RV) can avoid the mentioned information loss problem. Compared to the BEV and FV projections, a RV is a compact, and more importantly, an intrinsic representation from LiDAR. As such, a very recent trend is to combine the point cloud's RV with RGB image feature maps directly without incurring the projection loss. With the RV representation as input, FuseSeg~\citep{2020FuseSeg} establishes the point-pixel mapping and maximizes the multi-modal information.
In Tab.~\ref{view and feature}, we summarize the contributions and limitations of fusion methods in this category.
\begin{figure}[t]
  \centering
  \includegraphics[width=0.9\linewidth]{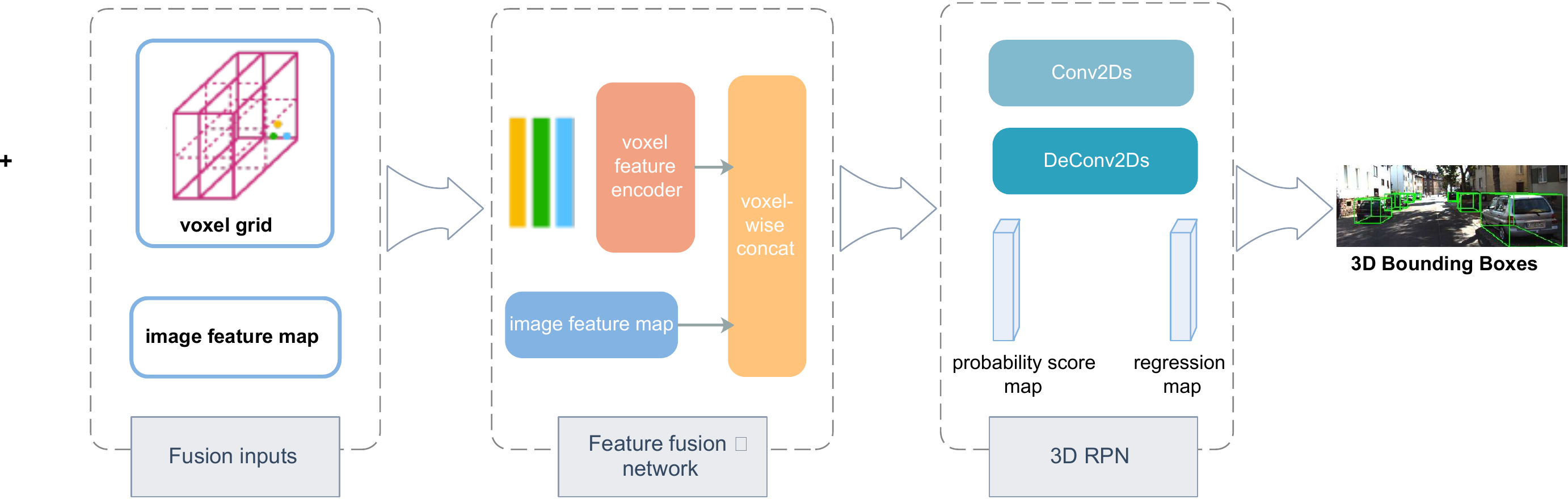}
  \caption{The MVX-Net pipeline that fuses voxelized point clouds and image feature maps~\citep{MVXNet}}
  \label{MVX-Net}
\end{figure}

\begin{table*}[ht]
\caption{Summary of methods that fuse point cloud's projections and image feature maps}
\label{view and feature}
\renewcommand\arraystretch{1.3}
\resizebox{\textwidth}{!}{
\begin{tabular}{llllll}
\hline
\textbf{Methods}                                  & \textbf{Year} & \textbf{Venue} & \textbf{\begin{tabular}[c]{@{}l@{}}image\\ backbone\end{tabular}} & \textbf{\begin{tabular}[c]{@{}l@{}}point cloud's\\  projections\end{tabular}} & \textbf{Contributions}                                                                                                                                                                                                                            \\ \hline
MV3D~\citep{2017Multi}                           & 2017          & CVPR           & VGG-16                                                            & BEV, FV                                                               & \begin{tabular}[c]{@{}l@{}}$\bullet$ Pioneer in exploiting BEV and FV LiDAR projections \\ and monocular camera frames to detect vehicles.\\ $\bullet$ Design a deep fusion architecture which allows interaction\\ between LiDAR and camera data.\end{tabular} \\ \hline
AVOD~\citep{8594049}                                 & 2018          & IROS           & VGG-16                                                            & BEV                                                                   & \begin{tabular}[c]{@{}l@{}}$\bullet$ Improve the detection of small targets via a feature \\ extractor that produces high-resolution feature maps.\end{tabular}                                                                                          \\ \hline
Contfuse~\citep{Contfuse}                             & 2018          & ECCV           & ResNet                                                            & BEV                                                                   & \begin{tabular}[c]{@{}l@{}}$\bullet$ Exploit continuous convolutions to fuse at different levels\\ of resolution.\end{tabular}                                                                                                                           \\ \hline
SCANet~\citep{DBLP:conf/icassp/LuCZZMZ19}             & 2019          & ICASSP         & VGG-16                                                            & BEV                                                                   & \begin{tabular}[c]{@{}l@{}}$\bullet$ Propose a spatial-channel attention module that is capable\\ of encoding multi-scale and global context information.\end{tabular}                                                                                  \\ \hline
FuseSeg~\citep{2020FuseSeg}                          & 2020          & WACV           & MobileNetV2                                                       & RV                                                                    & \begin{tabular}[c]{@{}l@{}}$\bullet$ Pioneer to utilize dense range views of point clouds. \\ $\bullet$ Establish point-wise correspondences between the range\\ view and image features.\end{tabular}                                                        \\ \hline
\end{tabular}
}
\end{table*}

\vspace{4pt}\noindent\textbf{voxelized point clouds $+$ image feature maps.} 
Voxelization turns irregular point clouds into regular 3D voxels. With voxelization becomes popular with point cloud processing~\cite{deng2020voxel,9157234,SECOND,8578570}, voxel grids have been commonly used as fusion input. The methods that fall in this category are summarized in Tab.~\ref{voxels and feature}.

As shown in Fig.~\ref{MVX-Net}, \citet{MVXNet} use voxels and image feature maps as input, projecting non-empty voxel features to the image plane through calibration. The image features are then concatenated to the voxel features through a designed fusion network. At the last stage, the 3D RPN processes the aggregated data and produces the 3D detection results. 
Similarly, in another work 3D-CVF~\citep{3D-CVF}, the spatial attention maps~\citep{Attention} are applied to weigh each modality depending on their contributions to the detection task. 
3D-CVF~\citep{3D-CVF} employs auto-calibrated projection to construct smooth joint LiDAR-camera features.

Nonetheless, methods in this category face the \textit{feature blurring} problem when only the center point of every voxel grid is projected onto the image feature. This results in the loss of detailed spatial information within each voxel.
Recently, to overcome this bottleneck, VPF-Net\citep{zhu2021vpfnet} cleverly aligns and aggregates the point cloud and image features at the ``virtual'' points. Particularly, with the density lying between 3D voxels and 2D pixels, the virtual points can nicely bridge the resolution gap between the two sensors and preserve more information for processing. Later, PointAugmenting~\citep{PointAugmenting} solves the blurring problem by first ``decorating'' raw points with corresponding features extracted by pre-trained 2D detection models. Then decorated points are voxelized and further processed. PointAugmenting also benefits from an occlusion-aware point filtering algorithm, which consistently pastes virtual objects into images and point clouds during training. 

\begin{table*}[ht]
\caption{Summary of methods that fuse voxelized point clouds and image feature maps}
\label{voxels and feature}
\renewcommand\arraystretch{1.35}
\resizebox{\textwidth}{!}{
\begin{tabular}{lllll}
\hline
\textbf{Methods}                             & \textbf{Year} & \textbf{Venue} & \textbf{\begin{tabular}[c]{@{}l@{}}image\\ backbone\end{tabular}} & \textbf{Contributions}                                                                                                                                                                                                                                 \\ \hline
MVX-Net~\citep{MVXNet}            & 2019          & ICRA           & VGG-16                                                            & \begin{tabular}[c]{@{}l@{}}$\bullet$ Propose two fusion schemes to fuse multi-modal information. \\
$\bullet$ \textit{PointFusion}: to aggregate  3D point is aggregated by an image \\ feature to capture a dense context; \textit{VoxelFusion}: a relatively later\\ fusion strategy where image features are appended at the voxel level.\end{tabular}     \\ \hline
3D-CVF~\citep{3D-CVF}              & 2020          & ECCV           & ResNet                                                            & \begin{tabular}[c]{@{}l@{}}$\bullet$ Combine the camera and LiDAR features using the cross-view\\ spatial feature fusion strategy.\\ $\bullet$ Employ an attention map to weigh the information from each\\ modality depending on their contributions.\end{tabular} \\ \hline
VPF-Net~\citep{zhu2021vpfnet}           & 2021          & TMM            & 2D CNNs                                                           & \begin{tabular}[c]{@{}l@{}}$\bullet$ Effectively alleviate the resolution mismatch problem in fusing\\ LiDAR and camera data.\\ $\bullet$ Explore further optimization through cut-n-paste based data\\ augmentation. \end{tabular}                                   \\ \hline
PointAugmenting~\citep{PointAugmenting}  & 2021          & CVPR           & ResNet                                                            & \begin{tabular}[c]{@{}l@{}}$\bullet$ Decorate point clouds with the corresponding CNN features.\\ $\bullet$ Design a novel cross-modal data augmentation algorithm considering\\ the modality consistency.\end{tabular}                                             \\ \hline
\end{tabular}
}
\end{table*}

\vspace{4pt}\noindent\textbf{raw point clouds $+$ image feature maps.}
As mentioned before, voxelized point clouds could cause certain degree of information loss due to voxelization process. 
Later, the advent of PointNet~\citep{8099499} makes it possible to process the raw point cloud directly without any projection or voxelization. Consequently, PointNet inspires a series of studies to combine points directly with feature maps as fusion input (summarized in Tab.~\ref{points and feature}).

\begin{figure}[t]
  \centering
  \includegraphics[width=0.9\linewidth]{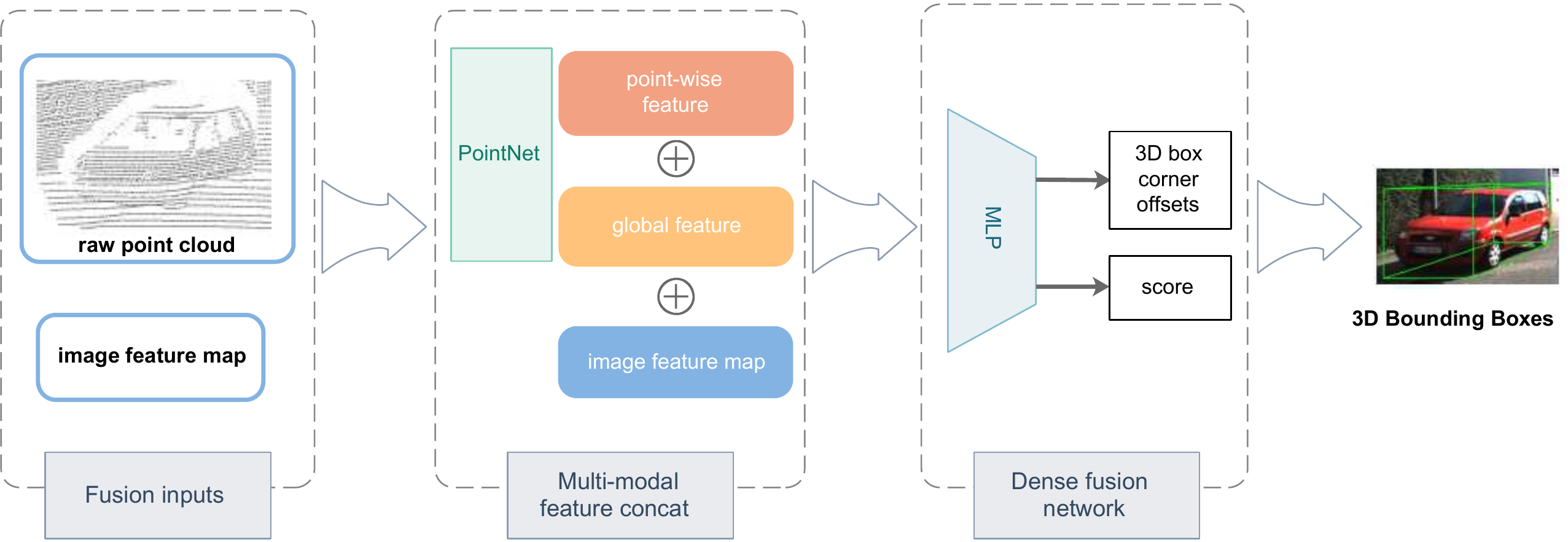}
  \caption{The PointFusion pipeline that fuses raw point clouds and image feature maps~\citep{8578131}}
  \label{pointfusion}
\end{figure}

Different from previous fusion-based methods, PointFusion~\citep{8578131} combines global image features from ResNet-50 and point cloud features from PointNet in a concatenation fashion, as demonstrated in Fig.~\ref{pointfusion}. Such concatenation operations, though simple, cannot align the multi-modal features nicely. 
Therefore, \citet{2020EPNet} propose LI fusion layer that explicitly establishes the mapping between point features and camera image features, thus providing finer and more discriminative representations. It also exploits point cloud features to estimate the importance of corresponding image features, which reduces the influence of occlusion and depth uncertainty.

\begin{figure}[t]
  \centering
  \includegraphics[width=0.9\linewidth]{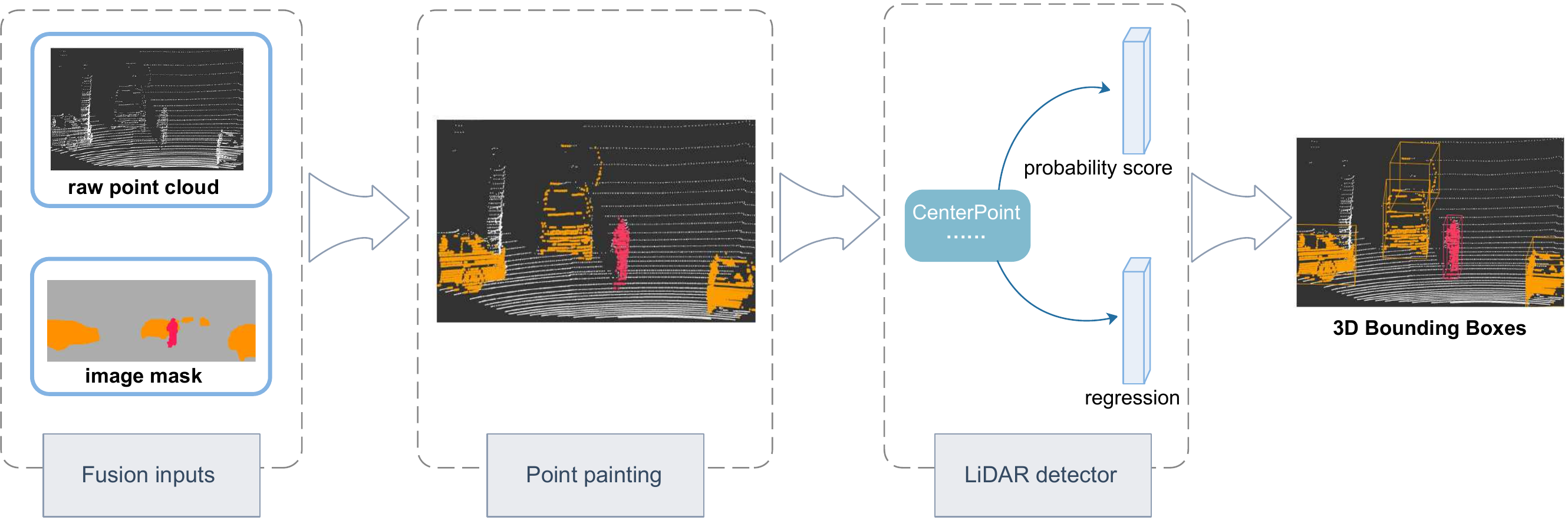}
  \caption{The PointPainting pipeline that fuses raw point clouds and image masks~\citep{9156790}}
  \label{pointpainting}
\end{figure}

\begin{table*}[ht]
\caption{Summary of methods that fuse raw point clouds and image feature maps}
\label{points and feature}
\renewcommand\arraystretch{1.3}
\resizebox{\textwidth}{!}{
\begin{tabular}{lllll}
\hline
\textbf{Methods}  & \textbf{Year} & \textbf{Venue} & \textbf{\begin{tabular}[c]{@{}l@{}}Image\\ backbone\end{tabular}}                  & \textbf{Contributions}                                                                                                                                                                                           \\ \hline
PointFusion~\citep{8578131}    & 2018          & CVPR           & ResNet                                                                             & \begin{tabular}[c]{@{}l@{}}$\bullet$ Can directly unitize ResNet and PointNet.\\ $\bullet$ Integrate the global and local features to predict \\ the bounding box.\end{tabular}                                                \\ \hline
PI-RCNN~\citep{PI-RCNN}        & 2020          & AAAI           & U-Net                                                                              & \begin{tabular}[c]{@{}l@{}}$\bullet$ Directly apply continuous convolution on raw points\\ to preclude the quantization loss.
\\ $\bullet$ Employ representation of deeper semantic features.\end{tabular}                    \\ \hline
EPNet~\citep{2020EPNet}        & 2020          & ECCV           & \begin{tabular}[c]{@{}l@{}}Four light-weighted\\ convolutional blocks\end{tabular} & \begin{tabular}[c]{@{}l@{}}$\bullet$ Enhance the point features with semantic image\\ features at a point-wise level.\\ $\bullet$ Exploit a consistency loss to encourage both\\ localization and classification.\end{tabular} \\ \hline
\end{tabular}
}
\end{table*}
\vspace{4pt}\noindent\textbf{raw point clouds $+$ image masks.}
\citet{PI-RCNN} conduct continuous convolution directly on 3D points, and meanwhile, it retrieves deeper semantic features instead of image features as image input.
The main rationales lie in the two aspects. 1) Features learned under the supervision of semantic segmentation are generally more expressive and compact when representing image. 2) It is feasible to obtain the homogeneous transformation matrix, which can build relationship between 2D masks and 3D points~\citep{9156790}.
As such, quite a few recent studies use result-level features such as segmentation masks to fuse with raw points, which is shown in Tab.~\ref{points and mask}.

In order to fuse the point cloud data and image masks, the LiDAR points are projected by a homogeneous transformation into the image plane, which establishes the 3D-2D mapping between the two.
This transformation on the KITTI dataset~\citep{geiger2013vision} is $T_{\text {camera} \leftarrow \text {LiDAR}}$, while it requires extra care for the nuScenes~\citep{nuScenes} transformation since the LiDAR and camera sensors operate at different frequencies. Let $T_{\text {car} \leftarrow \text {LiDAR}}$ be the transformation from the LiDAR sensor to the reference frame of cars, and let $T_{\text {camera} \leftarrow \text {car}}$ be the transformation from the reference frame of cars to the camera sensor. The complete matrix calculation is as below:
\begin{align}
   T_{\text {camera} \leftarrow \text {LiDAR}} = T_{\text {camera} \leftarrow \text {car}} \times T_{\text {car} \leftarrow \text {LiDAR}}.
\end{align}
Consequently, we can append the 2D mask as an additional channel of the corresponding 3D point.

In Fig.~\ref{pointpainting}, we present PointPainting~\citep{9156790} as a typical example fusion network.
It takes raw points and segmentation results as fusion input.
Next, in the fusion module (PointPaining dotted box of Fig.~\ref{pointpainting}), we first project the points onto the image, and then append segmentation scores to the raw LiDAR point. 
More importantly, PointPainting could be freely applied to both point-based and voxel-based LiDAR detectors and further improves the overall performance. 

Inspired by the successful PointPainting, CenterPointV2~\citep{yin2021center} gets almost the state-of-the-art result on nuScenes, and HorizonLiDAR3D~\citep{HorizonLiDAR3D} ranks the top on Waymo Open Dataset Challenge. 

\vspace{4pt}\noindent\textbf{Multi-Inputs for a Modality.}
\label{multi}
Meanwhile, as deep learning networks that are designed to process point clouds and images become increasingly diverse, it is also common for a single modality to adopt multiple inputs for fusion. 

MMF~\citep{8954034} is a pioneer in this category. It presents an end-to-end architecture that performs multiple tasks including 2D and 3D object detection, depth completion, \emph{etc}. Specifically, the fusion module takes the image feature map as well as the pseudo-LiDAR point clouds from the image branch, and BEV from the point cloud branch. These inputs are then fused jointly for 3D object detection.
Recently, \citet{MVAF} propose a multi-representation fusion framework that takes voxel grids, point cloud's RV projection, and image feature maps as input. They further estimate the importance of these three sources with attention modules to achieve adaptive fusion. 
\subsubsection{Discussion}
To summarize, point cloud inputs evolve from point cloud's projections, voxel grids, to raw points, which strives to minimize the information loss incurred in point cloud projection or voxelization. Meanwhile,
RGB image inputs evolve from lower-level feature maps to higher-level semantic segmentation results, which strives to exploit the richness of image data. 

Furthermore, to leverage the different perspectives from the same data stream, recent fusion networks start to take advantage of multiple inputs from the same modality. These trends are enabled by the rapidly growing computing capabilities as well as the fast development of powerful deep learning networks. These factors combined, more accurate LiDAR-camera fusion results are delivered.  
\begin{table*}[ht]
\caption{Summary of methods that fuse raw point clouds and image masks}
\label{points and mask}
\renewcommand\arraystretch{1.5}
\resizebox{\textwidth}{!}{
\begin{tabular}{llll}
\hline
\textbf{Methods}  & \textbf{Year} & \textbf{Venue} & \textbf{Contributions}                                                                                                                                                             \\ \hline
PointPainting~\citep{9156790}        & 2020          & CVPR           & \begin{tabular}[c]{@{}l@{}}$\bullet$ Pioneer in painting LiDAR point clouds with image-based semantic mask.\\ $\bullet$ Achieve fine-grained point-wise correspondence.\end{tabular}           \\ \hline
HorizonLiDAR3D~\citep{HorizonLiDAR3D}    & 2020          & Arxiv          & \begin{tabular}[c]{@{}l@{}}$\bullet$ Introduce a one-stage, anchor-free, and NMS-free detector.\\ $\bullet$ Effectively enhance the point cloud using point painting and test time augmentation. 
\end{tabular} \\\hline
CenterPointV2~\citep{yin2021center} & 2021          & CVPR           & \begin{tabular}[c]{@{}l@{}}$\bullet$ Represent, detect, and track 3D objects as points.\\ $\bullet$ The representation is compatible with off-the-shelf 3D encoders. \end{tabular}        
\\\hline
\end{tabular}
}
\end{table*}

\subsection{Fusion Granularity: how to fuse?}
\label{Fusion Granularity}
In this section, we discuss the third design choice for fusion-based algorithms. It is defined by at what granularity the two data streams are combined, which also addresses the question of ``how to fuse".

There are usually three options: RoI-level, voxel-level, and point-level (with the last one at the finest granularity). 
In general, fusion granularity is crucial to the complexity and effectiveness of the fusion framework.
Finer fusion granularity requires more computation but often leads to superior performance.
Below we discuss the three granularity levels in detail (summarized in Tab.~\ref{methods summary}).
\begin{figure}[t]
  \centering
  \includegraphics[width=0.9\linewidth]{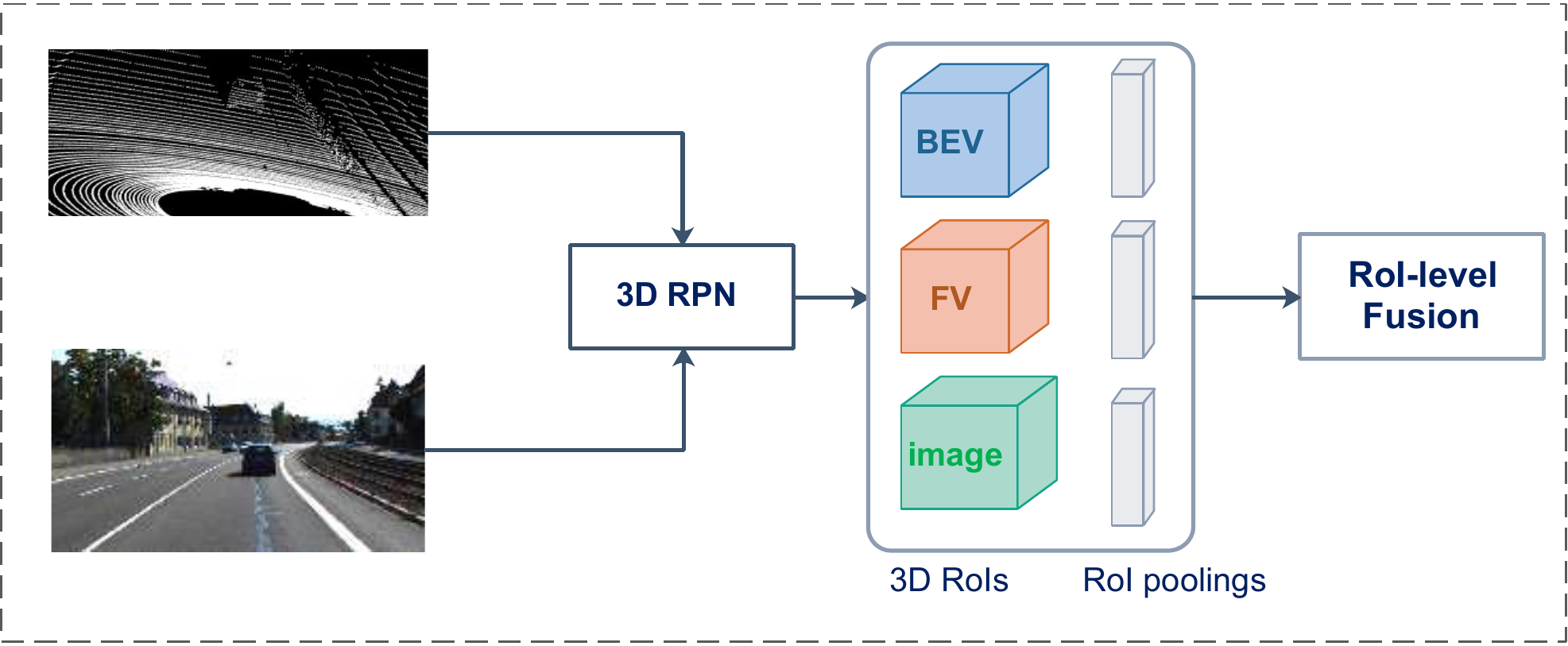}
  \caption{Illustration of RoI-level fusion. To perform fusion at RoI-level, we first obtain 3D RoIs from a shared set of 3D proposals and we employ RoI pooling~\citep{girshick2015fast} to get feature vectors of the same length.}
  \label{RoI-level}
\end{figure}
\subsubsection{RoI-level}
In essence, RoI-level fusion only fuses features at selected object regions instead of dense locations on the feature maps. Hence RoI-level fusion is normally performed at a relatively late stage (\emph{i.e.}, after the 3D region proposal generation stage).
This fusion granularity happens when applying RoI-pooling for each view to obtain feature vectors of the same length~\citep{2017Multi,8594049}, as illustrated in Fig.~\ref{RoI-level}. 
Also, it usually happens at the object proposal level in order to get 3D frustums from 2D RoIs through geometrical relationships~\citep{2019Frustum,8578131}.

As a result, RoI-level fusion limits the ability of the neural network to capture the cross-modality interactions at earlier stages.
To overcome this drawback, RoI-level fusion is often combined with other fusion granularity for further refinement of  proposals~\citep{8954034,3D-CVF}.

\subsubsection{Voxel-level}
\begin{figure}[t]
  \centering
  \includegraphics[width=0.9\linewidth]{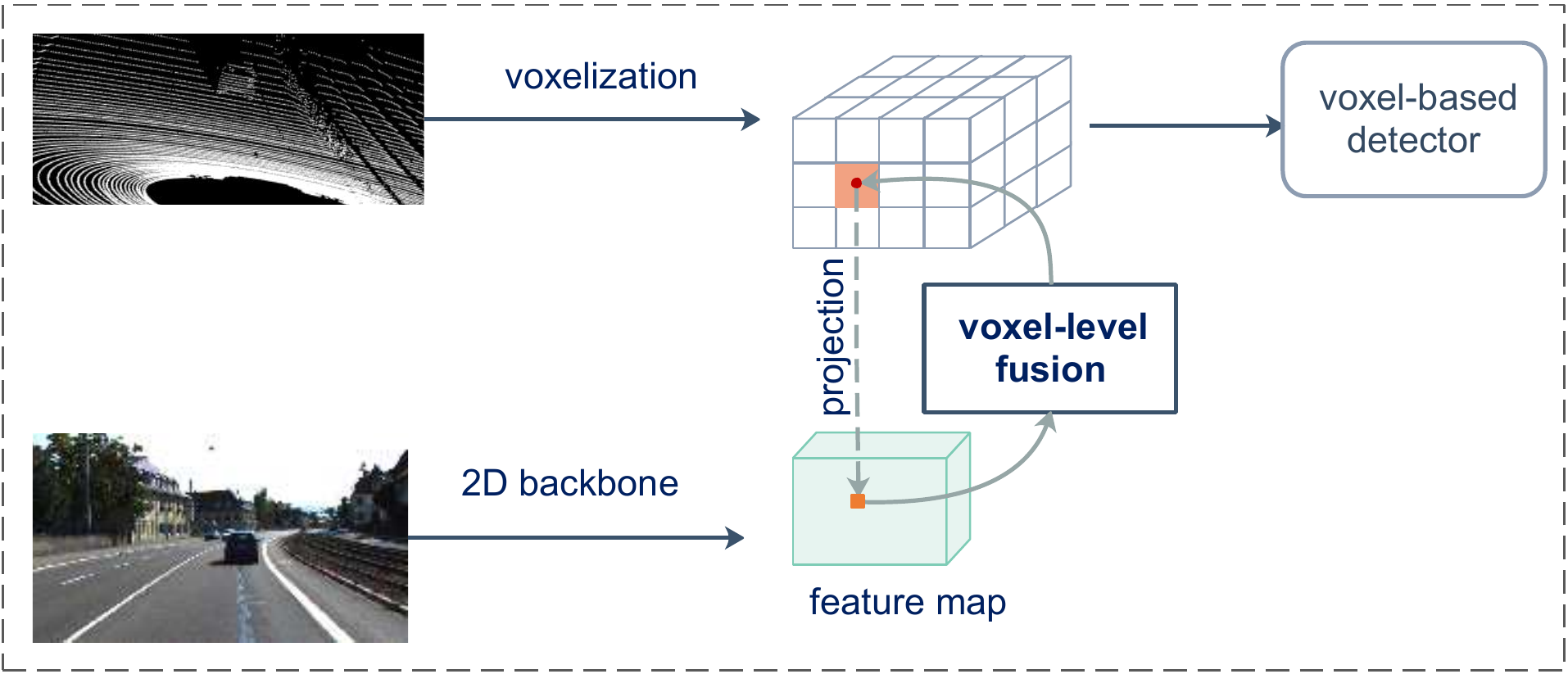}
  \caption{Illustration of voxel-level fusion. We can gain the relationship between voxel centers and image features through: 1) use camera projection matrix to obtain reference point; 2) fetch the corresponding feature in the image domain.
  }
  \label{voxel-level}
\end{figure}

Voxel-level fusion exploits a relatively earlier fusion stage compared with RoI-level. Voxelized point cloud data is usually projected onto the image plane so we can append the image feature to each voxel, which is described in Fig.~\ref{voxel-level}. 
Here, we establish a relatively approximate correspondence between the voxel features and image features.
Specifically, we project each voxel feature center to the image plane through camera projection matrix. After obtaining a reference point in the image domain, the corresponding image feature is appended to the LiDAR voxel feature branch.
Voxel-level fusion leads to a certain degree of information loss, resulting from both the spatial information loss in voxelization and the non-smooth camera feature maps.
To address this issue, one can combine neighboring image feature pixels by the interpolated projection to correct the spatial offsets, which can achieve more accurate correspondence between voxels and the image feature.~\citep{Contfuse,3D-CVF}.
Furthermore, instead of adopting a one-to-one matching pattern, we could explore cross attention mechanism that enables each voxel to perceive the whole image domain and adaptively attend corresponding 2D features.

In contrast to RoI-level fusion, this voxel-level granularity is finer and more precise.
Besides, to deal with empty voxels derived from LiDAR sparsity, voxel-level fusion could aggregate dense image information to compensate for sparse LiDAR features~\citep{MVXNet}. 
\subsubsection{Point-level}
\begin{figure}[t]
  \centering
  \includegraphics[width=0.9\linewidth]{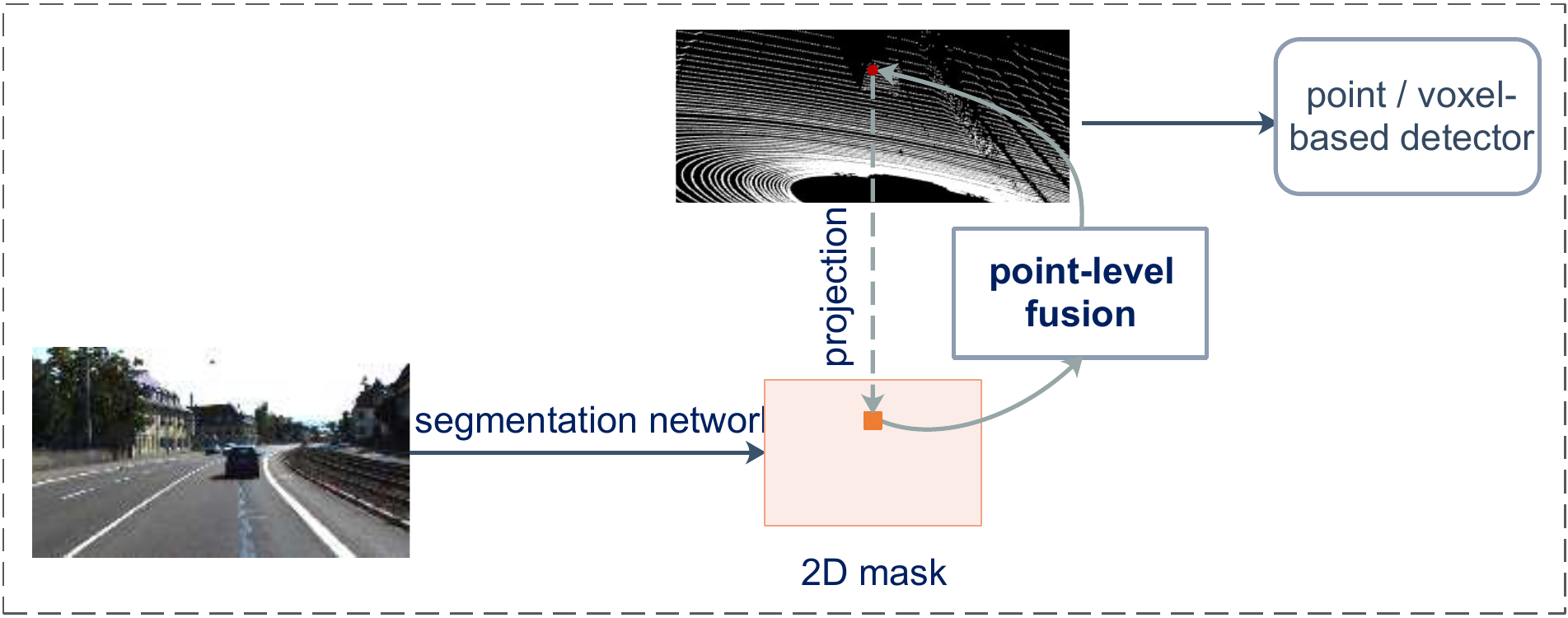}
  \caption{Illustration of point-level fusion. We first perform a 3D-2D projection between points and pixels, and then the corresponding image mask can be added as the additional  channel to decorate the 3D point cloud. For each decorated point, we flexibly select the voxel-based or point-based detector.}
  \label{point-level}
\end{figure}

Point-level fusion is usually early fusion, where every 3D point is aggregated by an image feature or mask in order to capture a dense context.
By ``lifting'' the corresponding image features or masks to the coordinates of the 3D points, point-level fusion provides an additional channel for each 3D point.
Specifically, we use the known transformation matrix~\citep{Calibration} to project 3D points to 2D image pixels and thereby establish a 3D-2D mapping.
Next, we can decorate the point or voxel features with the corresponding image masks through the mapping index.
Fig.~\ref{point-level} outlines this process.
The outstanding advantage of point-level fusion is the capability of summarizing useful information from both modalities since the image features are concatenated at a very early stage.
Compared with the above two fusion granularity levels, we can simply build corresponding relations between dense images and sparse point clouds without the blurring problem (refer to Sec.~\ref{Input Combinations})~\citep{9156790,PI-RCNN}. 
\par
Although experimental results show that point-level fusion effectively improves the overall performance~\citep{9156790,yin2021center}, there are still limitations. 
Firstly, due to the inherent occlusion problem in the image domain, 3D points which are mapped to the occluded image region may get the invalid image information~\citep{9156790}. 
Secondly, point-level fusion is less efficient in terms of memory consumption as compared to voxel-level fusion, as pointed out in~\citep{MVXNet}. 
\subsubsection{Discussion}
\begin{figure*}
  \centering
  \includegraphics[width=\linewidth]{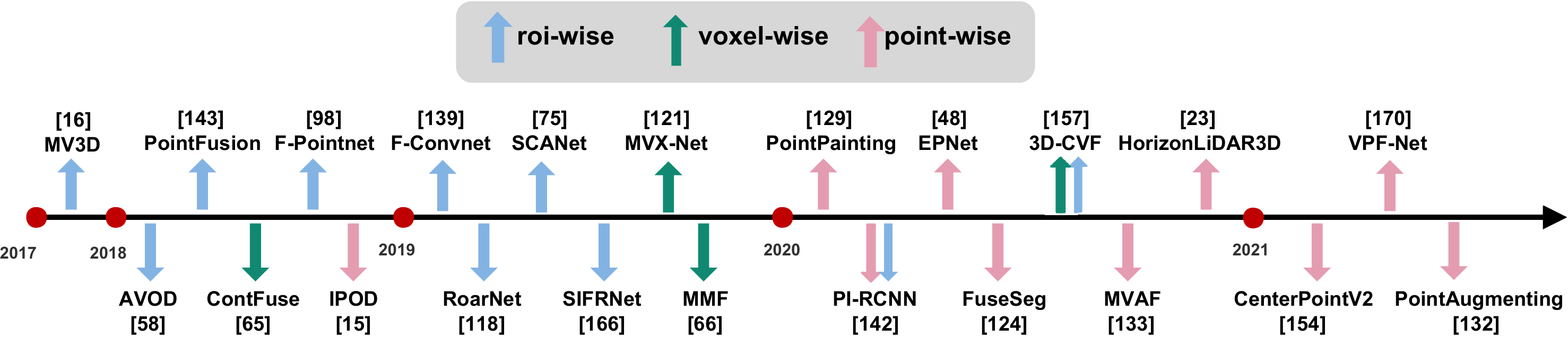}
  \caption{Timeline of the fusion-based 3D object detection methods. We use different colors to mark their fusion granularity.}
  \label{Timeline}
\end{figure*}

Fig.~\ref{Timeline} clearly shows the years in which the deep learning based multi-modal 3D detection methods appeared. We also mark the fusion granularity of each method.  
With the passage of time, we observe that the granularity was relatively coarse at first and becomes finer. 
Meanwhile, some fusion methods adopt more than one fusion granularity level for further refinement.

\subsection{LiDAR-Camera Fusion: summary and development}
\label{Summary of Camera-LiDAR Fusion Methods}
In summary, RoI-level fusion is rather limited as this fusion lacks deep feature interaction.
The later voxel-level and point fusion methods allow deep feature exchange and has its own merits.
However, some researches recently reveal that such methods are easily affected by the sensor misalignment due to the hard association between points and pixels established by calibration matrices.

Most recently, the success of BEV-based methods in BEV map segmentation encourages us to extend it to the fusion-based 3D object detection task~\cite{roddick2020predicting,yang2021projecting,zhou2022cross,liu2022petr,liu2022petrv2,huang2021bevdet,huang2022bevdet4d}.
Follow-up works~\cite{bai2022transfusion,liu2022bevfusion} have proved that fusing LiDAR features with camera features in BEV is robust against degenerated image quality and sensor misalignment. As such, a new \textbf{BEV-level} paradigm for LiDAR-camera fusion has emerged. 
Instead of collecting 2D masks or features by the 3D-2D hard association, these methods directly lift image features to the 3D world, and these lifted features can be processed to the BEV level to fuse with the LiDAR BEV feature at a certain stage of the detection pipeline.
For example, BEVfusion~\citep{liu2022bevfusion} lifts every image feature to the BEV space with an off-the-shelf depth estimator LSS~\citep{philion2020lift} in a learnable fashion, then these lifted points are processed by a separate 3D encoder to produce a BEV map, the LiDAR-camera fusion happens at the BEV level by merging the two BEV maps from both modalities.

\subsection{Fusion with Other Sensors}
\label{Fusion with Other Sensors}
So far, we have discussed LiDAR-camera fusion methods in depth. We next briefly summarize methods that involve the fusion with millimeter wave radar (which we refer to as mmWave radar in this paper for brevity) sensors. 
The employment of mmWave radar is getting popular recently due to its long ranges, low cost, and sensitivity to motions~\citep{GRIF}. 
Accordingly, we briefly discuss Radar-Camera fusion and LiDAR-Radar fusion.
\par
For Radar-Camera fusion, \citet{DBLP:conf/icra/ChadwickM019} project radar detection results to the image plane to boost the object detection accuracy for distant objects. 
Similarly, \citet{DBLP:conf/icip/NabatiQ19} use radar detection results to first generate 3D object proposals, then project them to the image plane to perform joint 2D object detection and depth estimation. 
CenterFusion~\citep{DBLP:journals/corr/abs-2011-04841} proposes to exploit both radar and camera data for 3D object detection. It first utilizes a center point detection network to detect objects by identifying their center points on the image. Next, it solves the key data association problem using a novel frustum-based method to associate radar detections with the corresponding 2D proposals. 
The above methods all directly use radar detection results without exploring features of radar points.
Instead, \citet{GRIF} propose a low-level sensor fusion 3D object detector that combines two RoIs from radar and camera feature maps by a Gated RoI Fusion (GRIF), which provides more robust vehicle detection performance.
\par
For LiDAR-Radar fusion, RadarNet~\citep{DBLP:conf/eccv/YangGLCU20} fuses radar and LiDAR data for 3D object detection. It employs an early fusion approach to learn joint representations from the two sensors and a decision fusion mechanism to exploit the radar’s radial velocity evidence. Disappointingly, RadarNet faces significant performance degradation in rare but critical adverse weather conditions. To remedy this, \citet{Qian_2021_CVPR} exploit complementary radar which is less impacted by adverse weather and becomes prevalent on vehicles. They present a two-stage deep fusion detector to enhance the overall detection results. Specifically, this method first generates 3D proposals from LiDAR and complementary radar and then fuse region-wise features between multi-modal sensor streams. 
\par
Finally, we would like to point out that it is also of use to fuse multiple sensors of the same kind. 
HorizonLiDAR3D~\citep{HorizonLiDAR3D} combines all point clouds generated by five LiDAR sensors to augment the information of the point cloud data. In this work, a simple concatenation of point clouds from all LiDAR sensors is performed. 

\section{Open Challenges and Possible Solutions}
\label{sec:open_challenge}

Sensor modalities hold different properties and capture the same scene from various perspectives, rendering it a challenging task to combine data from multiple modalities into a coherent data stream. In this section, we discuss open challenges and possible solutions for multi-modal 3D object detection, which we hope to provide helpful guidelines on how to improve the performance of the multi-sensor perception systems.
\subsection{Open Challenge I: Multi-Sensor Calibration}
\label{4.1}
\begin{figure}[t]
    \centering
    \includegraphics[width=\columnwidth]{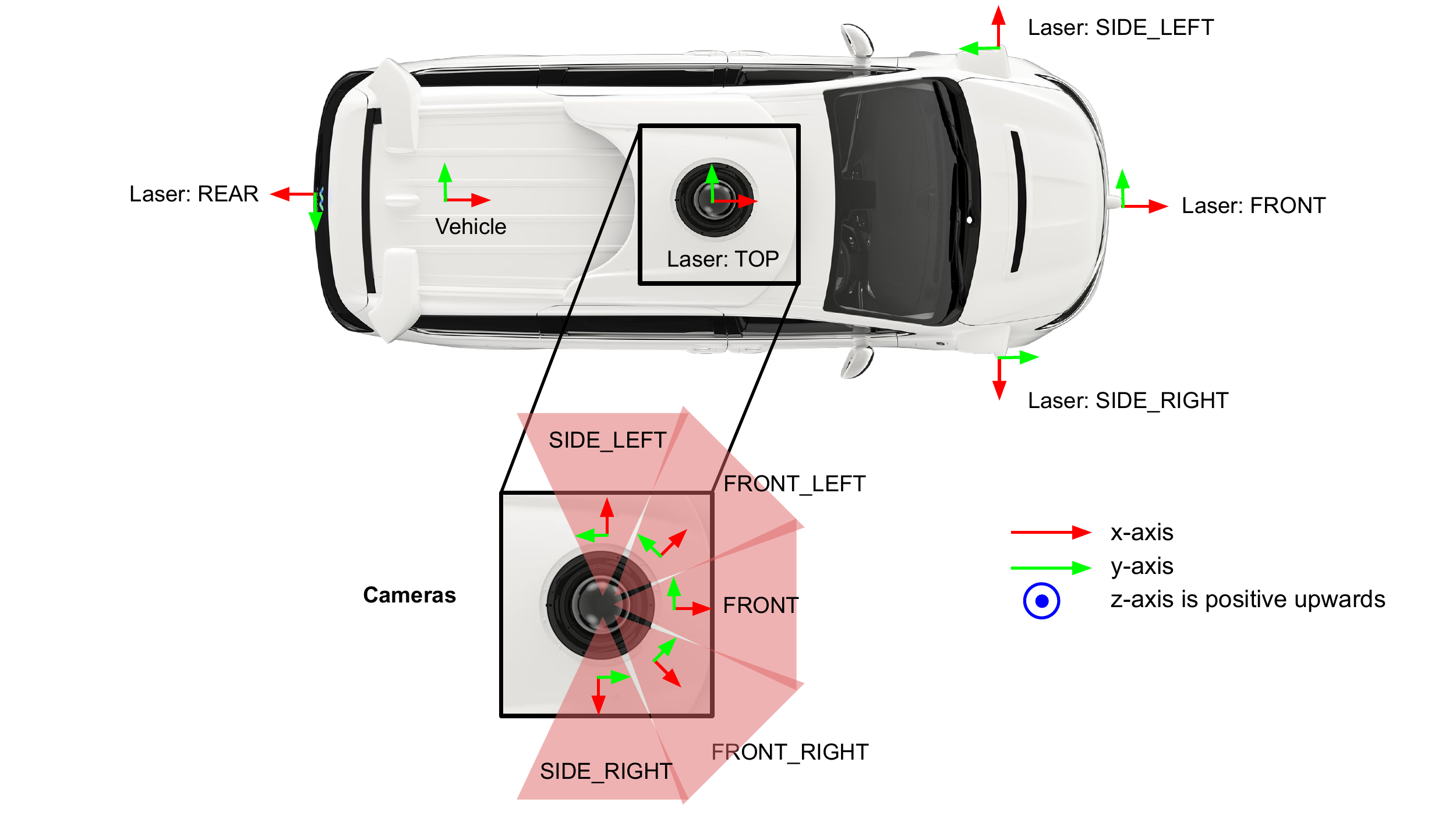}
    \caption{Cameras and LiDAR sensors deployed on the Waymo autonomous vehicle~\cite{waymo}}
    \label{fig:sensor_layout}
\end{figure}

As shown in Fig. \ref{fig:sensor_layout}, multiple sensors mounted on the autonomous vehicle are from different sensor coordinates. Fusion based methods are required the alignment of these sensor data. Here, we use LiDAR-camera fusion as example to explain this challenge. Point clouds are a set of points indicating 3D coordinates of the objects. RGB images are matrices of pixels, with each pixel's coordinate represented as $(x, y)$, where $x, y$ is the pixel's row and column index, respectively. To build the map from 3D LiDAR coordinates to the 2D image plane, we must perform calibration between the two.

Traditional calibration methods use a calibration target to derive the intrinsic and extrinsic camera parameters. This cumbersome process requires lots of manual efforts. 
A common practice is to develop a targetless, automatic calibration method that can continuously calibrate the LiDAR sensor and camera on the fly. 
Target-less calibration is currently an active topic of research in this field~\citep{auto_calib_rss,auto_calib_aaai, regnet}. 
These methods automatically calibrate among multiple sensors without human experts.  
However, inevitable bumps and jitters when driving AVs lead to the variation of the extrinsic parameters for the well calibrated LiDAR-camera system. Much worse, the error will gradually accumulate if not corrected in time, which may eventually affect the perception results.
A possible solution to prevent this problem is to integrate the LiDAR and camera in a suite\citep{apolloscape, waymo}, preventing their relative displacement to the greatest extent.

\subsection{Open Challenge II: Information Loss during Fusion}

When fusing data from multiple modalities, a certain amount of information will be lost inevitably due to projection, quantization, feature burring, \emph{etc.} When devising a multi-modal fusion network, we need to pay attention to the stage, input, and granularity of the fusion operation, in order to minimize the information loss.

The choice of fusion stage results in a different level of information loss. A later fusion stage is easy to implement, but cannot enjoy the rich information embedded in the raw data or earlier feature maps. Considering the complexity of the problem, it is very challenging, if at all possible, to pinpoint the optimal fusion stage that balances information loss and ease of implementation.
To this end, a possible solution is to consider utilizing Neural Architecture Search (NAS) technique~\citep{DARTS, 2020Searching} to find the appropriate fusion stage within a pipeline. It defines search space and then devises a search algorithm to propose near-optimal neural architectures.

The choice of fusion inputs has the greatest bearing on the amount of information loss, as a result of data projection or voxelization. For example, converting a point cloud to its BEV compresses the point cloud in the vertical direction and thus leads to the loss of height information. Converting a point cloud to its RV suffers from the problem of scale variation. Accordingly, it's important to find suitable input representations that reserve rich geometric and semantic information as much as possible. 
Moving forward, we could investigate several possible solutions. Specifically, we could exploit the attention mechanism~\citep{Attention,DBLP:conf/cvpr/0004GGH18} to enhance certain features for each modality. Or, we could employ multiple representations to retain important information. For example, we can utilize both the point cloud and the corresponding voxel grid as fusion input for the point cloud branch~\cite{deng2020voxel,DBLP:journals/tcsv/DengZZL21}.
However, existing approaches do not take full advantage of temporal fusion input, which potentially limits the performance of multi-modal 3D object detection. In the further, we believe it is of significance to learn the 4D spatio-temporal information fusion across sensor and time.

The choice of fusion granularity can also affect the amount of information loss, \emph{e.g.}, aligning the multi-modal data in a coarse granularity leads to the problem of feature blurring. 
A possible solution is to employ learnable calibration offsets to aggregate neighbor spatial information~\citep{3D-CVF}. In this way, we can maximize the effect of data fusion.

\begin{figure}[t]
    \centering
    \subfigure[the raw image]{
    \includegraphics[width=0.8\linewidth]{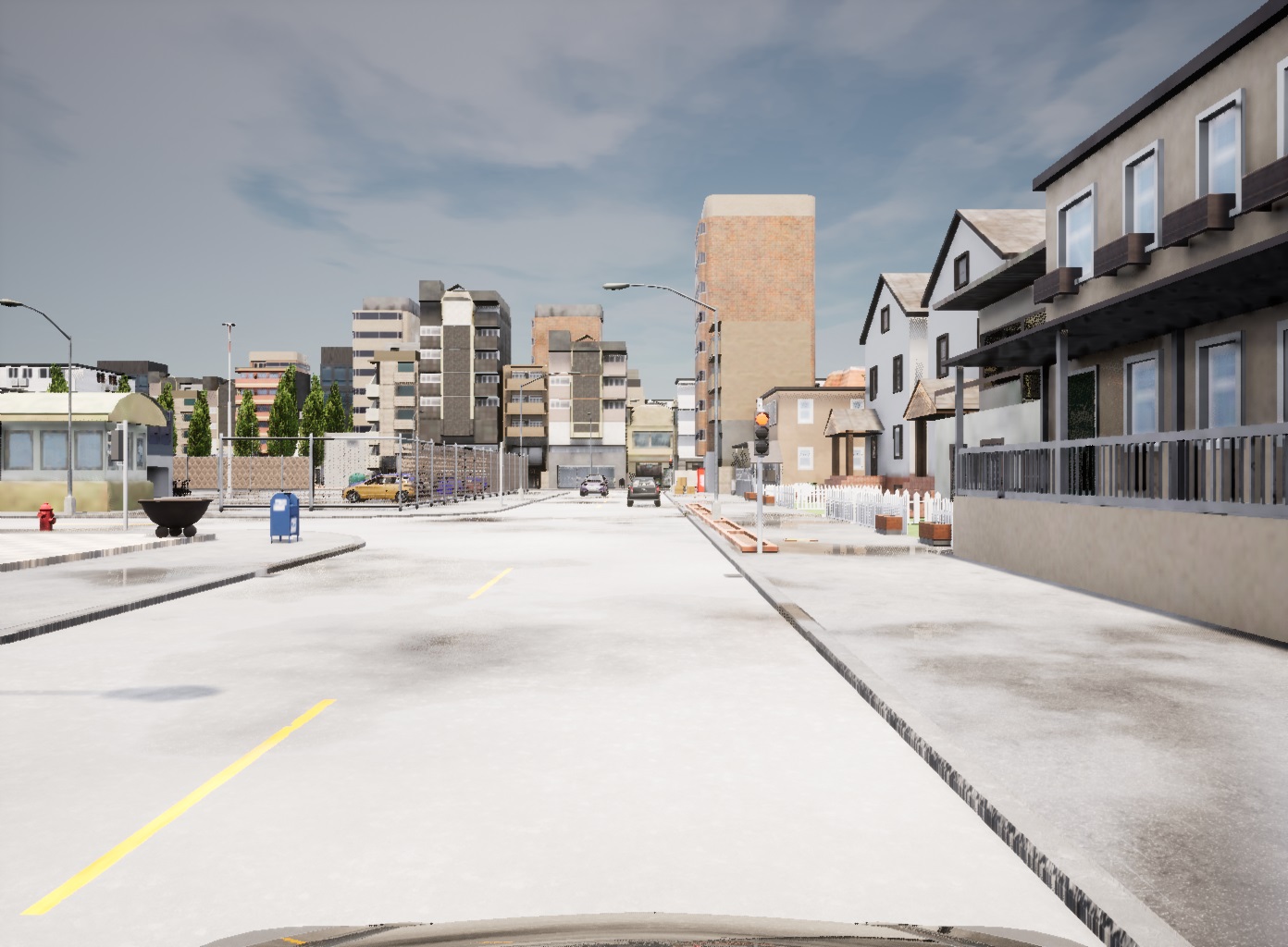}
    }
    \subfigure[the synthetic point cloud]{
    \includegraphics[width=0.8\linewidth]{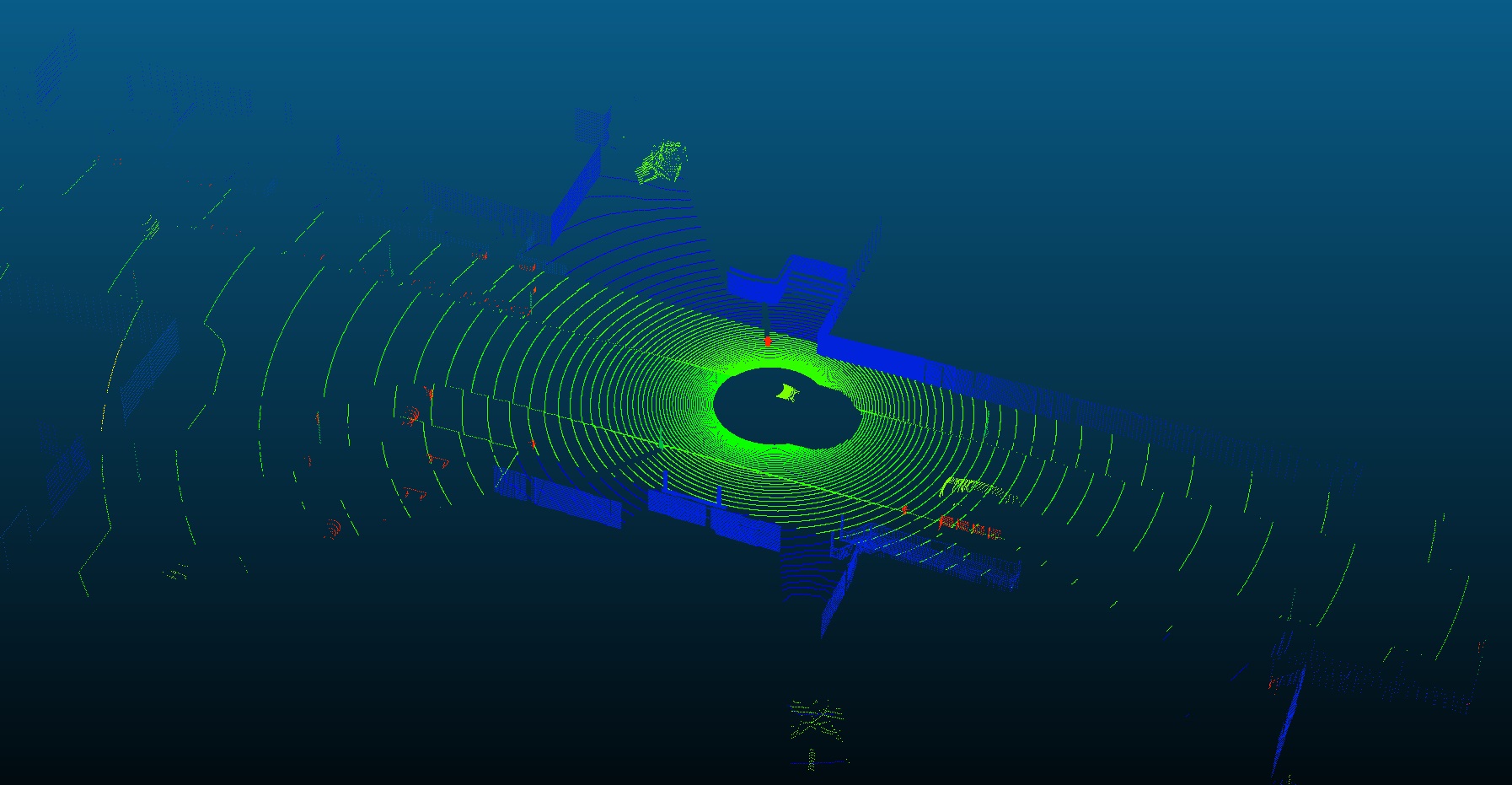}
    }
    \caption{An example RGB image (a) and the corresponding synthetic point cloud with semantic segmentation annotations (b). Both images are obtained from KITTI-CARLA synthetic dataset~\citep{deschaud2021kitti}.}
    \label{kitti_carla}
\end{figure}

\subsection{Open Challenge III: Efficient Multi-Modal Data Augmentation}
Due to the limited number of objects in the dataset, data augmentation is usually adopted to ensure efficient learning and avoid overfitting. 
Existing data augmentation techniques for each single data stream can be applied to deep fusion methods, such as 
object cut-and-paste, random flipping, scaling, rotation, and so on~\citep{SECOND,8578570}. 
However, to keep data augmentation consistent across multiple modalities, we need to build the fine-grained mapping between data elements (such as points or pixels). Unfortunately, the augmentation operations usually choose to work on randomly selected objects and are thus inconsistent across the modalities.

Recently, several methods are proposed~\citep{PointAugmenting,zhang2020multimodality} to address this problem.
\citet{zhang2020multimodality} present a new multi-modality augmentation approach by cutting point cloud and imagery patches of ground-truth objects and pasting them into different scenes in a consistent manner, which prevents misalignment between multi-modal data.
When projecting 3D points to 2D pixels, it first performs the reverse operation of translation, rotation, flip, \emph{etc.} to restore the original point cloud, then gets point-pixel mapping based on the calibration information. In the future, more efficient multi-modal augmentation techniques need to be investigated.
\subsection{Open Challenge IV: Low-Cost Multi-Modal 3D Object Detection}
\vspace{-1ex}
Monocular or stereo cameras are the most common low-cost sensors that can meet the requirements of mass production. 
However, without accurate 3D geometry information, relying on cameras alone cannot yield 3D detection results comparable to LiDAR-based methods. In fact, the state-of-the-art monocular method DD3D~\citep{park2021pseudo} achieves only 16.87\% mAP on the KITTI 3D object detection leader board; the best stereo method LIGA-Stereo~\citep{guo2021liga} can achieve 64.66\% mAP. Nevertheless, the best LiDAR-only method BtcDet~\citep{xu2021behind} has obtained 82.86\% mAP.

Moving forward, with the development of \textit{knowledge distillation}~\citep{hinton2015distilling}, one could exploit LiDAR data to distill 3D geometric information for camera-based detectors using large-scale and well-calibrated multi-modal data. Such a method can potentially achieve accurate detection as well as low system cost. 

\subsection{Open Challenge V: Shortage of Large Datasets}
Another bottleneck in multi-modal 3D detection is the availability of high-quality, publicly usable datasets annotated with ground-truth information. Currently, popular datasets in 3D detection have the following issues: small size, class imbalance, and labeling errors, as discussed in Sec.~\ref{sec:datasets}.

Unsupervised and weakly-supervised fusion networks could allow the networks to be trained on large, unlabeled or partially labeled datasets~\citep{Pseudo-labeling}.

There are also emerging works on generating synthetic datasets for RGB images and point clouds~\citep{deschaud2021kitti,carla,virtual_kitti,meta-sim,manivasagam2020lidarsim, sdr,playing_for_data,synthia, aiodrive}, which provide large-scale data with rich annotations. Fig.~\ref{kitti_carla} shows an example of the KITTI-CARLA~\citep{deschaud2021kitti} synthetic dataset. 
However, there may be a domain gap between synthetic datasets and real-world datasets.
Some recent works~\citep{hodavn2019photorealistic,denninger2020blenderproc,guizilini2021geometric,richter2022enhancing} try to utilize technologies such as photorealistic rendering, unsupervised domain adaptation, and generative adversarial networks (GANs)~\citep{goodfellow2020generative} to bridge the gap between synthetic and real-world data.
Still, how to use the models trained on the synthetic data to deal with real-world scenarios remains to be further investigated. 

\section{Conclusion}
\label{sec:conclusion}
Due to the increasing importance of 3D vision in applications such as autonomous driving, this paper reviews the recent multi-modal 3D object detection networks, especially those that fuse camera images and LiDAR point clouds. 
We first carefully compare popular sensors and discuss their advantages and disadvantages and summarize the common problems of single-modal methods. 
We then provide an in-depth summary of several popular datasets that are commonly used for autonomous driving. 
In order to provide a systematic review, we discuss the multi-modal fusion methods based upon their choices for the following three design considerations: (1) fusion stage, \emph{i.e.}, where does the fusion take place in the pipeline, (2) fusion input, \emph{i.e.}, what data inputs are used for fusion, and (3) fusion granularity, \emph{i.e.}, at what granularity level the two data streams are combined. 
Finally, we discuss open challenges and potential solutions in multi-modal 3D object detection.

\bibliographystyle{spbasic}      
{\footnotesize
\bibliography{reference}
}
\end{document}